\documentclass{article} % For LaTeX2e
\usepackage{iclr2026_conference,times}
\usepackage{amsmath,amsfonts,bm}

\def\eqref#1{equation~\ref{#1}}
\def\1{\bm{1}}

\DeclareMathAlphabet{\mathsfit}{\encodingdefault}{\sfdefault}{m}{sl}
\SetMathAlphabet{\mathsfit}{bold}{\encodingdefault}{\sfdefault}{bx}{n}

\usepackage{hyperref}
\usepackage{url}

\usepackage{amsthm}
\newtheorem{theorem}{Theorem}[section]

\newtheorem{lemma}[theorem]{Lemma}

\newtheorem{remark}[theorem]{Remark}
\usepackage{amsmath}
\usepackage{amssymb}
\usepackage{amsfonts}
\usepackage{enumitem}
\usepackage{algorithm}
\usepackage{algpseudocode}
\usepackage{kotex}
\usepackage{makecell}
\usepackage{multirow}
\usepackage{graphicx}
\usepackage{caption}
\usepackage{wrapfig}
\usepackage{booktabs}

\title{Bayesian Neural Networks for Functional ANOVA model}

\makeatletter
\def\@fnsymbol#1{%
  \ifcase#1\or
    \dagger % 1번째 footnote 기호: *
  \or \diamondsuit % 2번째: †
  \or \star % 3번째: ‡
  \or \mathsection % 4번째: §
  \or \mathparagraph % 5번째: ¶
  \or \| % 6번째: ‖
  \or **% 7번째: **
  \or \dagger\dagger % 8번째: ††
  \or \ddagger\ddagger % 9번째: ‡‡
  \fi}
\makeatother
\author{
    Seokhun Park$^{1*}$%
    , Choeun Kim$^{1*}$%
    , Jihu Lee$^1$%
    , Yunseop Shin$^1$%
    , Insung Kong$^2$%
    , Yongdai Kim$^{1\dagger}$%
    \\[0.4em]
    \textsuperscript{1}Department of Statistics, Seoul National University
    \\
    \textsuperscript{2}Department of Applied Mathematics, University of Twente
    \\
    \texttt{\footnotesize \{shrdid, kimchoeun, rieky0426, dbstjq48\}@snu.ac.kr,} \\
    \texttt{\footnotesize insung.kong@utwente.nl, ydkim0903@gmail.com}
}

\iclrfinalcopy

\begin{document}

\maketitle
\def\thefootnote{*}\footnotetext{Equal contribution.}
\def\thefootnote{$\dagger$}\footnotetext{Corresponding author.}

\begin{abstract}
With the increasing demand for interpretability in machine learning, functional ANOVA decomposition has gained renewed attention as a principled tool for breaking down high-dimensional function into low-dimensional components that reveal the contributions of different variable groups.
Recently, Tensor Product Neural Network (TPNN) has been developed and applied as basis functions in the functional ANOVA model, referred to as ANOVA-TPNN.
A disadvantage of ANOVA-TPNN, however, is that the components to be estimated must be specified in advance, which makes it difficult to incorporate higher-order TPNNs into the functional ANOVA model due to computational and memory constraints.
In this work, we propose Bayesian-TPNN, a Bayesian inference procedure for the functional ANOVA model with TPNN basis functions, enabling the detection of higher-order components with reduced computational cost compared to ANOVA-TPNN.
We develop an efficient MCMC algorithm 
and demonstrate that Bayesian-TPNN performs well by analyzing multiple benchmark datasets.
Theoretically, we prove that the posterior of Bayesian-TPNN is consistent.
\end{abstract}

\section{Introduction}
As artificial intelligence (AI) models become increasingly complex, the demand for interpretability has grown accordingly.
To address this need, various interpretable models—including both post-hoc explanations \citep{ribeiro2016should, lundberg2017unified} and inherently transparent models \citep{agarwal2021neural,koh2020concept,radenovic2022neural,park2025tensor}—have been studied.
Among various interpretable approaches, our study focuses on the functional ANOVA model, a particularly important class of interpretable models that decompose a high-dimensional function into a sum of low-dimensional functions called {\it componenets} or {\it interactions}.
Notable examples of the functional ANOVA model are the generalized additive Model \citep{hastie1986generalized}, SS-ANOVA 
\citep{SSANOVA} and MARS \citep{MARS}.
Because complex structures of a given high-dimensional model can be understood by interpreting 
low-dimensional components, the functional ANOVA models have been extensively used in interpretable AI applications \citep{lengerich2020purifying, martens2020neural, choi2024meta, herren2022statistical}.

In recent years, various neural networks have been developed to estimate components in the functional ANOVA model.
Neural Additive Models (NAM, \cite{agarwal2021neural}) estimates each component of the functional ANOVA model using deep neural networks (DNN), and Neural Basis Models (NBM, \cite{radenovic2022neural}) significantly reduce the computational burden of NAM by using basis deep neural networks (DNN). 
NODE-GAM \citep{chang2021node} can select and estimate the components in the functional ANOVA model simultaneously, and \cite{thielmann2024neural} proposes NAMLSS, which modifies NAM to estimate the predictive distribution.
\cite{park2025tensor} proposes ANOVA-TPNN, which estimates the components  under the uniqueness constraint and thus provides a stable estimate of each component.

Existing neural-network approaches to functional ANOVA model require prohibitive computation when the input dimension $p$ is large, because the number of components—and thus the required networks—grows exponentially. 
As a result, only 1–2 dimensional components are typically used, yielding suboptimal prediction when higher-order interactions matter.

In this paper, we propose a Bayesian neural network (BNN) for the functional ANOVA model which can estimate higher-order interactions (i.e., components whose input dimension is greater than 2) without requiring huge amounts of computing resources.
{\it The main idea of the proposed BNN is to infer the architecture (the architectures of neural networks for each component) as well as the parameters (the weights and biases in each neural network).
To explore higher posterior regions of the architecture, a specially designed MCMC algorithm is developed which searches the architectures in a stepwise manner (i.e., growing or pruning the current architecture) and thus huge computing resources for memorizing and processing all of the predefined neural networks for the components can be avoided.}

% Bayesian Neural Networks (BNN; \cite{mackay1992practical, neal2012bayesian, wilson2020bayesian, izmailov2021bayesian}) offer a principled Bayesian framework for training DNNs and have garnered significant attention in the fields of machine learning and artificial intelligence.
% Beyond their theoretical appeal, Bayesian Neural Networks (BNNs) are often credited with stronger generalization and more informative uncertainty quantification than their frequentist counterparts \citep{wilson2020bayesian, izmailov2021bayesian}. 
% By maintaining posterior distributions over the architectures and weights, BNNs yield calibrated predictive uncertainties that can improve the quality of decision making in terms of prediction accuracy and uncertainty quantification.
% These advantages have motivated applications across diverse domains, including recommender systems \citep{wang2015collaborative}, topic modeling \citep{gan2015scalable} and medical diagnosis \citep{filos2019systematic}, where reliable generalization and uncertainty estimates are particularly consequential.
Bayesian Neural Networks (BNN; \cite{mackay1992practical, neal2012bayesian, wilson2020bayesian, izmailov2021bayesian}) provide a principled Bayesian framework for training DNNs and have received considerable attention in machine learning and AI. 
Compared to frequentist approaches, BNN offers stronger generalization and better-calibrated uncertainty estimates \citep{wilson2020bayesian, izmailov2021bayesian}, which enhance decision making. These properties have motivated applications in areas such as recommender systems \citep{wang2015collaborative}, topic modeling \citep{gan2015scalable}, and medical diagnosis \citep{filos2019systematic}.
More recently, Bayesian neural networks (BNN) that learn their own architectures have been actively studied. 
In particular, \cite{kong2023masked} introduced a node-sparse BNN, referred to as the masked BNN (mBNN), and established its theoretical properties.
\cite{nguyen2024sequential} proposes S-RJMCMC, which explores architectures and weights by jointly sampling parameters and altering the number of nodes.

This is the first work on BNN that efficiently estimates higher-order components in the functional ANOVA model without requiring substantial computing resources.
Our main contributions can be outlined as follows.
\begin{itemize}
    \item We propose a BNN for the functional ANOVA model called Bayesian-TPNN which treats the architecture as a learnable parameter, and develop an MCMC algorithm which efficiently explores high-posterior regions of the architecture.
    
    \item For theoretical justifications of the proposed BNN, we prove the posterior consistency of the prediction model
    as well as each component.
    
    \item Through experiments on multiple real datasets, we show that the proposed BNN provides more accurate and stable estimation and uncertainty quantification than other neural networks for the functional ANOVA model.
    On various synthetic datasets, we further show that Bayesian-TPNN effectively estimates important higher-order components.
\end{itemize}

\section{Preliminaries}

\subsection{Notation}
Let $\mathbf{x} = (x_1, \dots, x_p)^\top \in \mathcal{X}$ be a $p$-dimensional input vector, where $\mathcal{X}= \mathcal{X}_{1} \times \cdots \times \mathcal{X}_{p} \subseteq [0, 1]^p$.
We write $[p] = \{1, \dots, p\}$ and its power set with cardinality $d$ as $\text{power}([p],d)$.
For any component $S \subseteq [p]$, we denote $\mathbf{x}_{S} = (x_{j}, {j \in S})^\top$ and define $\mathcal{X}_{S} = \prod_{j \in S}\mathcal{X}_{j}$.
A function defined on $\mathcal{X}_{S}$ is denoted by $f_{S}$.
For any real-valued function $f : \mathcal{X} \xrightarrow{}  \mathbb{R}$, we define the empirical $\ell_{2}$-norm as $\Vert f \Vert_{2,n}:= (\sum_{i=1}^{n}f(\mathbf{x}_{i})^{2}/ n)^{1/ 2},$ 
where $\mathbf{x}_1,\ldots,\mathbf{x}_n$ are observed input vectors.
We denote $\sigma(\cdot)$ as the sigmoid function, i.e., $\sigma(x):=1/(1+\exp(-x))$.
We denote by $\mu_{n}$ the empirical distribution of $\{\mathbf{x}_{1}, \ldots, \mathbf{x}_{n}\}$, and by $\mu_{n,j}$ the marginal distribution of $\mu_{n}$ on $\mathcal{X}_{j}$.

\subsection{Probability model for the likelihood}
We consider a nonparametric regression model in which the conditional distribution of $Y_{i}$ given $\mathbf{x}_{i}$ follows an exponential family \citep{10.1214/09-AOS762, chen2024estimating}:
\begin{align}
Y_{i}|\mathbf{x}_{i} \sim \mathbb{Q}_{f(\mathbf{x}_{i}),\eta}
\end{align}
for $i=1,...,n,$
where $f:\mathcal{X} \to \mathbb{R}$ is a regression function and $\eta$ is a nuisance parameter.
Here, we assume that $\mathbb{Q}_{f(\mathbf{x}),\eta}$ admits the density function $q_{f(\mathbf{x}),\eta}$ defined as
\begin{align}
q_{f(\mathbf{x}),\eta}(y) = \exp\bigg( { f(\mathbf{x})y - A(f(\mathbf{x})) \over \eta} + S(y,\eta) \bigg), 
\label{eq:expontial_eta}
\end{align}
where $A(\cdot)$ is the log-partition function, ensuring that the density integrates to one.
We assume that each input vector $\mathbf{x}_{i}$ has been rescaled, yielding $\mathbf{x}_{i} \in [0,1]^{p}$ for $i=1,...,n$.

\paragraph{Example 1. Gaussian regression model:}
Consider the gaussian regression $Y = f(\mathbf{x}) + \epsilon$, where $\epsilon \sim N(0,\sigma_{\epsilon}^{2})$. 
In this case, the density in (\ref{eq:expontial_eta}), corresponds to  $A(f(\mathbf{x})) := {f(\mathbf{x})^{2} / 2}$ and $S(y,\eta) := -{y^{2}/ 2\eta} - {(\log 2\pi\eta)/2}$ with
$\eta = \sigma_{\epsilon}^{2}$.

\paragraph{Example 2. Logistic regression model:}
For a binary outcome $Y \in \{0,1\}$, consider 
the logistic regression model
$Y|\mathbf{x} \sim \text{Bernoulli}(\sigma(f(\mathbf{x})))$.
In this case, there is no nuisance parameter, i.e., $\eta = 1$.
This distribution can be expressed as the exponential family with
$A(f(\mathbf{x})) := \log(1+e^{f(\mathbf{x})})$ and $S(y,\eta) := 0$.

\paragraph{Likelihood:}
Let $\mathcal{D}^{(n)}=\{(\mathbf{x}_1,y_1),\ldots,(\mathbf{x}_n,y_n)\}$
be given data which consist of $n$ pairs of observed input vectors and response variables.
For the likelihood, we assume that
$y_i$s are independent realizations of $Y_{i}|\mathbf{x}_{i} \sim \mathbb{Q}_{f(\mathbf{x}_{i}),\eta},$
where $f$ and $\eta$ are the parameters to be inferred.

\subsection{Functional ANOVA model}

For $S \subseteq [p]$, we say that $f_{S}$ satisfies the sum-to-zero condition with respect to a probability measure $\mu$ on $\mathcal{X}$ if
\begin{align}
\begin{split}
\text{For}\:\:\: S \subseteq [p],\: \: \forall  \: j \in S \mbox{ and }  \forall \: \mathbf{x}_{S\backslash\{j\}}\in \mathcal{X}_{S\backslash\{j\}},
\: \int_{\mathcal{X}_{j}}f_{S}(\mathbf{x}_{S})\mu_{j}(dx_j)=0  \label{eq:sum-to-zero_def}
\end{split}
\end{align}
holds, where $\mu_{j}$ is the marginal probability measure of $\mu$ on $\mathcal{X}_{j}$.

\begin{theorem}[Functional ANOVA Decomposition \citep{hooker2007generalized, mcbook}]
\label{thm:fANOVA_decomp}
Any real-valued function $f$ defined on $\mathbb{R}^{p}$ can be uniquely decomposed as 
\begin{align}
f(\mathbf{x}) = \sum_{S \subseteq [p]}f_{S}(\mathbf{x}_{S}), 
\label{eq:f-ANOVA}
\end{align}
almost everywhere with respect to $\Pi_{j=1}^{p}\mu_{j},$
where each component $f_{S}$ satisfies the sum-to-zero condition with respect to $\mu$.
\end{theorem}

Theorem \ref{thm:fANOVA_decomp} guarantees a unique decomposition of any real-valued multivariate function $f$ into the components satisfying the sum-to-zero condition with respect to the probability measure $\mu$.
In (\ref{eq:f-ANOVA}), we refer to $f_{S}$ as main effects when $|S|=1$, as second-order interactions when $|S| = 2$, and so on.
For brevity, we use the empirical distribution $\mu_{n}$ for $\mu$ when referring to the sum-to-zero condition.

\subsection{Tensor Product Neural Networks}
In this subsection, we review Tensor Product Neural Network (TPNN) proposed by \cite{park2025tensor} since we use it as a building block of our proposed BNN. TPNN is a specially designed neural network to satisfy the sum-to-zero condition.

For each $S \subseteq [p]$, TPNN is defined as
$f_{S}(\mathbf{x}_{S}) = \sum_{k=1}^{K_{S}}\beta_{S,k}\phi(\mathbf{x}_{S}|S,\mathfrak{B}_{S,k},\mathfrak{R}_{S,k})$ for component $f_S$, 
where $\beta_{S,k} \in \mathbb{R}$, $\mathfrak{B}_{S,k}=(b_{S,j,k}, j \in S) \in \mathbb{R}^{|S|}$, and $\mathfrak{R}_{S,k}=(\gamma_{S,j,k}, j\in S) \in (0,\infty)^{|S|}$.
Here, $\phi(\mathbf{x}_{S}|S,\mathfrak{B}_{S,k},\mathfrak{R}_{S,k})$ is defined as
\begin{small}
\begin{align}
\phi(\mathbf{x}_{S}|S,\mathfrak{B}_{S,k},\mathfrak{R}_{S,k}) := \prod_{j \in S}\bigg( 1 - \sigma\bigg({x_{j}-b_{S,j,k}\over \gamma_{S,j,k}}\bigg) + c_{j}(b_{S,j,k},\gamma_{S,j,k}) \sigma\bigg({x_{j}-b_{S,j,k}\over \gamma_{S,j,k}}\bigg) \bigg),
\label{eq:model_str_def}
\end{align} 
\end{small}
where 

\begin{equation}
\label{eq:c_j}
c_{j}(b,\gamma) := 
-\Big( 1-\int_{\mathcal{X}_{j}}\sigma\left({x_{j}-b\over \gamma}\right)\mu_{n,j}(dx_{j})\Big) \Big/\int_{\mathcal{X}_{j}}\sigma\left({x_{j}-b\over\gamma}\right)\mu_{n,j}(dx_{j}).
\end{equation}
The term $c_{j}(b,\gamma)$ is introduced to make $\phi(\mathbf{x}_{S}|S,\mathfrak{B}_{S,k},\mathfrak{R}_{S,k})$
satisfy the sum-to-zero condition.
Finally, \cite{park2025tensor} proposes ANOVA-T$^{d}$PNN, which assumes that:
\begin{align}
\label{eq:ANOVA-TPNN}
f(\mathbf{x}) = \sum_{S \subseteq [p], |S| \leq d}\sum_{k=1}^{K_{S}}\beta_{S,k}\phi(\mathbf{x}_{S}|S,\mathfrak{B}_{S,k},\mathfrak{R}_{S,k}),
\end{align}
where $d \in \mathbb{N}_{+}$ and $\{K_{S}, S \subseteq [p], |S| \leq d\}$ are hyperparameters.
Since $\phi(\cdot|S,\mathfrak{B}_{S,k},\mathfrak{R}_{S,k})$ satisfies the sum-to-zero condition for any $S \subseteq [p]$, $f_{\text{ANOVA-}\text{T}^{d}\text{PNN}}$ also satisfies the sum-to-zero condition.
Therefore, we can estimate the components uniquely by estimating the parameters in ANOVA-T$^{d}$PNN.

Here, $d$ is the maximum order of components.
Note that as the maximum order $d$ increases, the number of TPNNs in (\ref{eq:ANOVA-TPNN})
grows exponentially; therefore, in practice $d$ is set to $1$ or $2$
due to the limitation of computing resources. In addition, choosing $K_S$s is not easy.
To further illustrate these limitations, the experiments on the runtime of Bayesian-TPNN and ANOVA-T$^{2}$PNN are presented in Section \ref{sec:runtime} of Appendix.

\begin{wrapfigure}{r}{0.48\linewidth}
% \begin{figure}[t]
    \centering
    \includegraphics[width=0.9\linewidth]{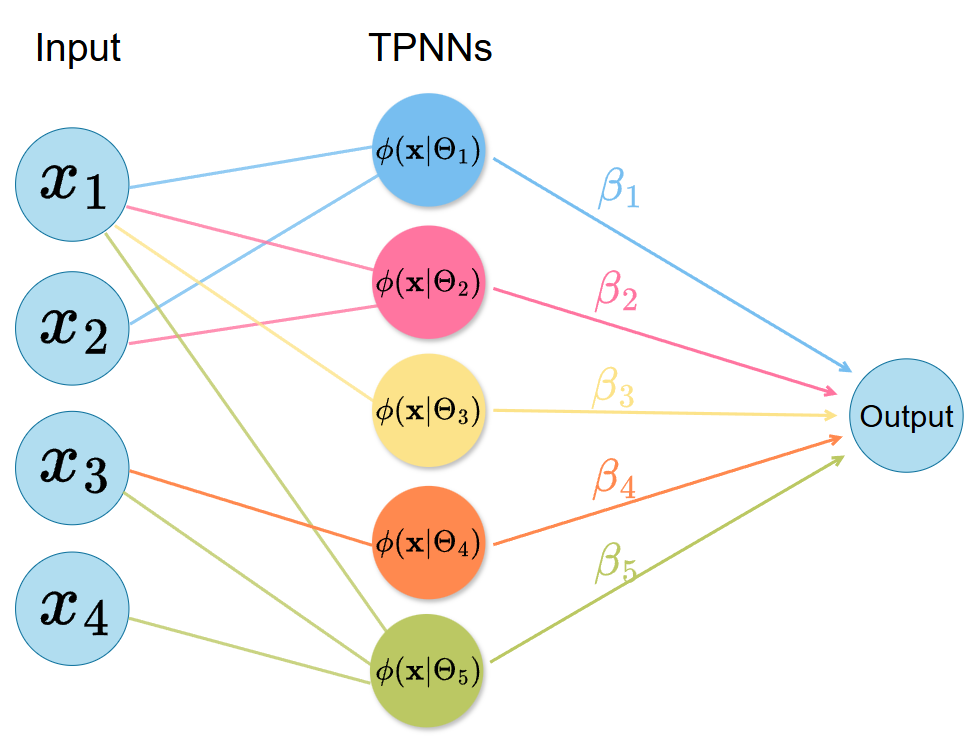}
    \caption{\footnotesize \textbf{Bayesian-TPNN with $p=4$ and $K=5$.}}
    \label{fig:archtecture_btpnn}
    \vskip -0.70cm
% \end{figure}
\end{wrapfigure}

\section{Bayesian Tensor Product Neural Networks}

In (\ref{eq:ANOVA-TPNN}), instead of fixing $S$, we treat $S$ also as learnable parameters. 
That is, we consider the following model:
\begin{align}
f(\mathbf{x}) = \sum_{k=1}^{K}\beta_{k}\phi(\mathbf{x}|\Theta_{k}), \label{eq:model}
\end{align}
where $\Theta_{k} = (S_{k},\mathbf{b}_{S_{k},k},\Gamma_{S_{k},k})$, $S_k \subseteq [p]$, and aim to learn $K$ and $(S_k, k \in [K])$ as well as  the other parameters.
Here, 
\begin{align*}
\mathbf{b}_{S_{k},k} & := (b_{j,k}, j \in S_{k}) \in [0,1]^{|S_{k}|},\\  
\Gamma_{S_{k},k} & := (\gamma_{j,k} , j \in S_{k}) \in (0,\infty)^{|S_{k}|}.
\end{align*}
for $k \in [K]$.
{\it Note that $K$ and $S_k$ are considered to be the parameters defining the architecture, but
they cannot be updated by a gradient descent algorithm since
$K$ and $S_k$s are not numeric parameters. 
Instead, we adopt a Bayesian approach in which $K$ and $S_k$s are explored via an MCMC algorithm.}
We refer to the resulting model as \emph{Bayesian Tensor Product Neural Networks (Bayesian-TPNN)}.
Bayesian-TPNN can be understood as an edge-sparse shallow neural network
when $K$ is the number of hidden nodes and $S_K$ is the set of input variables linked to the $k$-th hidden node through active edges.
See Figure \ref{fig:archtecture_btpnn} for an illustration.

\subsection{Prior}
\label{sec:prior}

The parameters in Bayesian-TPNN consist of 
$K$, $\mathcal{B}_{K} := (\beta_{1},...,\beta_{K})$, $\mathbf{S}_{K} := (S_{k}, k \in [K])$, $\mathbf{b}_{\mathbf{S}_{K},K} := (\mathbf{b}_{S_{k},k}, k \in [K])$, $\Gamma_{\mathbf{S}_{K},K} := (\Gamma_{S_{k},k}, k \in [K])$ and the nuisance parameter $\eta$ if it exists (e.g. the variance of the noise in the gaussian regression model).
The parameters can be categorized into the three groups: (1) $K$ for the node-sparsity, (2) $S_k, k=1,\ldots,K$
for the edge sparsity, and (3) all the other parameters including $(\mathbf{b}_{S_{k},k},\Gamma_{S_{k},k},k=1,...,K)$.
We use a hierarchical prior for these three groups of parameters.

\paragraph{Prior for $K$:}
We consider the following prior distribution for $K$:
\begin{align}
\pi(K=k) \propto \exp(-C_{0}k\log n), \quad \text{for} \quad k=0,...,K_{\max},
\label{eq:prior_K}
\end{align}
where $K_{\max} \in \mathbb{N}_{+}$ and $C_{0} > 0$ are hyperparameters. 
This prior is motivated by \cite{kong2023masked}.

\paragraph{Prior for $\mathbf{S}_{K}| K$:}
Conditional on $K$, we assume a prior that $S_k$s are independent and
each $S_{k}$ follows the mixture distribution: 
\begin{align}
\sum_{d=1}^{p}w_{d}\text{Uniform}\big(\text{power}([p],d)\big),
\end{align}
where $w_{d}$s are defined recursively as follows:
$w_{d} \propto \big(1-p_{\text{adding}}(d)\big)\prod_{\ell < d}p_{\text{adding}}(\ell) $
with
$p_{\text{adding}}(\ell) := \alpha_{\text{adding}}(1+\ell)^{-\gamma_{\text{adding}}}$.
Here, $p_{\text{adding}}$ is the probability of adding a variable to $S_k$, controlled by hyperparameters $\alpha_{\text{adding}}$ and $\gamma_{\text{adding}}$. This prior is inspired by Bayesian CART \citep{chipman1998bayesian}, where $S_k$ denotes split variables.

\paragraph{Prior for the numeric parameters given $K$ and $\mathbf{S}_{K}$:}
All the remaining parameters are numerical ones and hence we use standard priors for them.
\begin{itemize}
 \item Conditional on $K$, we assume a prior that $\beta_k$s are independent and follow
$\beta_{k} \sim N(0,\sigma_{\beta}^{2})$, where $\sigma_{\beta} > 0$ is a hyperparameter.

\item Conditional on $S_{k}$, we let $b_{j,k}$s and $\gamma_{j,k}$s be all independent and
$ b_{j,k} \sim \text{Uniform}(0,1) $ and $ \gamma_{j,k} \sim \text{Gamma}(a_{\gamma},b_{\gamma})$
for $j \in S_{k}$ and $k \in [K]$, where $a_{\gamma} >0$ and $b_{\gamma} >0$ are hyperparameters.

\item For the nuisance parameter in
the gaussian regression model, where  the nuisance parameter $\eta$ corresponds to $\sigma^{2},$
we set $\sigma^{2} \sim \text{IG}\big({v\over 2},{v\lambda \over 2}\big)$, where $v >0$ and $\lambda >0$ are hyperparameters and $\text{IG}(\cdot,\cdot)$ is the inverse gamma distribution. 

\end{itemize}

\subsection{MCMC Algorithm for Posterior Sampling}
\label{sec:mcmc}

We now develop an MCMC algorithm for posterior sampling of Bayesian-TPNN. 
Our overall sampling strategy is to update $K,\: \mathbf{S}_K$ and the remaining numeric parameters
iteratively using the corresponding Metropolis-Hastings (MH) algorithms, which is motivated by the MCMC algorithm of
Bayesian additive regression tree \citep{chipman2010bart}.
A novel part of our MCMC algorithm, however, is to devise a specially designed proposal distribution in the MH algorithm
such that  the proposal distribution encourages the MCMC algorithm to visit important higher-order interactions more frequently.
For this purpose, we introduce two special tools.
First, we employ a pretrained probability mass function $p_{\text{input}}(\cdot)$ on $[p],$
which represents the importance of each input variable. Further, let $p_{\text{input}}(\cdot|S)$
be the distribution $p_{\text{input}}(\cdot)$ restricted to $S \subseteq [p]$.
See Remark at the end of this subsection for the choice of $p_{\text{input}}(\cdot)$.

The second tool is a stepwise search. The stepwise search adds a new node by first copying one of existing nodes
and add an edge. By doing so, a newly added node has one more edges than the copied node and thus corresponds to
an interaction whose order is larger than the copied one by 1. 
By keeping the copied node also in the model, we
can avoid dramatic loss of accuracy.

To be more specific, let $ \theta := (K, \mathbf{S}_{K},\mathbf{b}_{\mathbf{S}_{K},K},\Gamma_{\mathbf{S}_{K},K},\mathcal{B}_{K},\eta)$ be given current parameters.
We update these parameters by sequentially updating  $K$,   $(\mathbf{S}_{K},\mathbf{b}_{\mathbf{S}_{K},K},\Gamma_{\mathbf{S}_{K},K},\mathcal{B}_{K})$ and  the nuisance parameter $\eta$. 
We now describe these 3 updates.

\paragraph{Updating $K$:} First, we devise a proposal distribution of $K^{\text{new}}$ given $K$ used in the MH algorithm.
For a given $K,$ we set $K^{\text{new}}$ as $K-1$ or $K+1$ with probability $K/K_{\max}$ and $1-K/K_{\max}$ respectively. If $K^{\text{new}}=K-1$, we remove one of $(S_{k},\mathbf{b}_{S_{k},k},\Gamma_{S_{k},k},\beta_{k}), k\in [K]$ from $\theta$ with probability $1/K$ to have $\theta^{\text{new}}$.  

For the case $K^{\text{new}}=K+1$, the crucial mission is to design an appropriate proposal of $(S_{K+1}^{\text{new}},\mathbf{b}_{S_{K+1}^{\text{new}},K+1}^{\text{new}},\Gamma_{S_{K+1}^{\text{new}},K+1}^{\text{new}},\beta_{K+1}^{\text{new}})$.
Specifically, we first generate $S_{K+1}^{\text{new}}$ and then generate $(\mathbf{b}_{S_{K+1}^{\text{new}},K+1}^{\text{new}},\Gamma_{S_{K+1}^{\text{new}},K+1}^{\text{new}},\beta_{K+1}^{\text{new}})$ conditional on $S_{K+1}^{\text{new}}$.
The proposal of $S_{K+1}^{\text{new}}$ consists of the following two alternations:

\begin{itemize}
   \item \textbf{Random}: Generate $S_{K+1}^{\text{new}}$ from the prior distribution.
   \item \textbf{Stepwise}: Propose $S_{K+1}^{\text{new}}=S_{k^*}\cup \{j_{k^*}\},$
   where $k^*\sim \text{Uniform}[K]$ and $j_{k^*} \sim p_{\text{input}}(\cdot|S_{k^{*}}^c)$.
\end{itemize}

The MH algorithm randomly selects one of $\{\textbf{Random}, \textbf{Stepwise} \}$ with probability $M/(M+K),$ and $K/(M+K)$, where $M>0$ is a hyperparameter.
This proposal combines random and stepwise search, where $S_{K+1}^{\text{new}}$ is sampled as a completely new index set from the prior with probability $M/(M+K)$, or taken as a higher-order modification of one of $S_{1},\ldots,S_{K}$ with probability $K/(M+K)$.
We employ \textbf{Stepwise} move to encourage the proposal distribution to explore higher-order interactions more frequently without losing much information in the current model (i.e. keeping all of the components in the current model).
Once $S_{K+1}^{\text{new}}$ is given, we generate $(\mathbf{b}_{S_{K+1}^{\text{new}},K+1}^{\text{new}},\Gamma_{S_{K+1}^{\text{new}},K+1}^{\text{new}}, \beta_{K+1}^{\text{new}})$ from the prior distribution.
See Section \ref{sec:K_MH} of Appendix for the acceptance probability for this proposal $\theta^{\text{new}}$ and see Section \ref{sec:impoact_K_proposal} of Appendix for experimental results demonstrating the effectiveness of the proposed MH.
%The proposal for $S_{K+1}$ is motivated by the interaction search under the heredity constraint ({\bf references}), where
%we include interactions (e.g. $f_{1,2}(x_1,x_2)$) into the model only when one of lower order interactions
%(e.g. $f_1(x_1)$ and $f_2(x_2)$) already exists in the model. Mathematically, the heredity constraint 
%reduces the search space significantly so that more efficient search would be possible, and
%the heredity constraint is known to be a reasonable assumption for practical purposes (i.e. there exists an accurate
%prediction model satisfying the heredity assumption) ({\bf reference}).
%Note that our proposal does not follow the heredity constraint with probability $M/(M+K),$ which makes the proposed MCMC algorithm be able to
%search a prediction model not satisfying the heredity constraint.
%{\bf In the numerical study, we show that this flexibility is helpful to explore high-posterior regions.}

 \paragraph{Updating $(S_{k},\mathbf{b}_{S_{k},k},\Gamma_{S_{k},k},\beta_{k})$ for $k\in [K]$:}

For a given $k,$ we consider the following three possible alterations of $S_{k}$ and $(\mathbf{b}_{S_{k},k},\Gamma_{S_{k},k})$
for the proposal of $(S_{k}^{\text{new}},\mathbf{b}_{S_{k}^{\text{new}},k}^{\text{new}},\Gamma_{S_{k}^{\text{new}},k}^{\text{new}})$:
\begin{itemize}
\item \textbf{Adding}: Adding a new variable $j^{\text{new}}$, which is selected randomly from $S_k^c$
according to the probability distribution $p_{\text{input}}(\cdot|S_k^c)$, 
 and generating $b_{j^{\text{new}}, k}$ and $\gamma_{j^{\text{new}}, k}$
 from the prior distribution.

\item \textbf{Deleting}: Uniformly at random, select an index $j$ in $S_k$ and delete it from $S_k$.

\item \textbf{Changing}: Select an index $j$ uniformly at random from $S_k$ and index $j^{\text{new}}$ from $S_k^c$
according to the probability distribution of $p_{\text{input}}(\cdot|S_{k}^{c})$
and delete $j$ from $S_k$ and add $j^{\text{new}}$ to $S_k.$ Then, 
generate $b_{j^{\text{new}}, k}$ and $\gamma_{j^{\text{new}}, k}$
 from the prior distribution.
\end{itemize}

The MH algorithm randomly selects one of \{\textbf{Adding}, \textbf{Deleting}, \textbf{Changing}\} with probability $(q_{\text{add}},q_{\text{delete}}, 
q_{\text{change}})$. This proposal distribution is a modification of one used in BART \citep{chipman1998bayesian, kapelner2016bartmachine} to grow/prune or modify a current decision tree.
See Section \ref{sec:s_algorithm} of Appendix for the acceptance probability of $(S_{k}^{\text{new}},\mathbf{b}_{S_{k}^{\text{new}},k}^{\text{new}},\Gamma_{S_{k}^{\text{new}},k}^{\text{new}})$.

Once $(S_{k},\mathbf{b}_{S_{k},k},\Gamma_{S_{k},k})$ are updated, we update 
all of the numeric parameters $(\mathbf{b}_{S_{k},k},\Gamma_{S_{k},k},\beta_{k})$ by the MH algorithm with the Langevin proposal \citep{rossky1978brownian} to accelerate the convergence of the MCMC algorithm further.
Finally, we repeat this update for $k\in [K]$ sequentially.
See Appendix \ref{sec:b-gam-mh} for details and Section I for a toy example illustrating the proposed MCMC algorithm.

 \paragraph{Updating the nuisance parameter $\eta$ :}
In the gaussian regression model, the nuisance parameter $\eta$ corresponds to the error variance $\sigma^{2}_{g}$.
Since the conditional posterior distribution of $\sigma^{2}_{g}$ is Inverse Gamma distribution, 
it is straightforward to draw $\sigma^{2}_{g}$ from $\pi(\sigma^{2}_{g}|\text{others})$. 
Details are provided in Section \ref{sec:eta_smaple} of Appendix.

% Finally, $\beta_{0}$ is updated via the MH algorithm, which is detailed in Section \ref{sec:beta0} of the Appendix.
% Algorithm \ref{alg:MCMC} summarizes the proposed MCMC algorithm of Bayeisan-TPNN

\begin{algorithm}[H]
    \caption{MCMC algorithm of Bayesian TPNN.}\label{alg:MCMC}
    \textbf{Input} $\{(\mathbf{x}_i,y_i)\}_{i=1}^n$ : data, $K$ : initial number of hidden nodes, $M_{\text{mcmc}}$ : the number of MCMC iterations,
    \begin{algorithmic}[1] %[1] enables line numbers
        \For{i : 1 to $M_{\text{mcmc}}$}
        \State Update $K$
        \For{k : 1 to $K$}
            \State Update $S_{k},\mathbf{b}_{S_{k},k},\Gamma_{S_{k},k}$
            \State Update $\mathbf{b}_{S_{k},k},\Gamma_{S_{k},k},\beta_{k}$ 
%            \State $S_k \sim \text{MH}_{S_{k}}(\beta_k,\mathbf{b}_{S_k,k},\Gamma_{S_k,k},\eta,\bm{\lambda}_k,\mathcal{D}^{(n)},\mathbf{w}) $
%            \State $(\mathbf{b}_{S_k,k},\Gamma_{S_k,k}) \sim \text{MH}_{\mathbf{b}_{S_k,k},\Gamma_{S_k,k}}( \beta_k,S_k,\lambda_k,\eta,\bm{\lambda}_k,\mathcal{D}^{(n)}) $
%            \State $\beta_{k} \sim \text{MH}_{\beta_{k}}( S_k,\mathbf{b}_{S_{k},k},\Gamma_{S_t,t},\eta,\bm{\lambda}_k,\mathcal{D}^{(n)})$
        \EndFor
%        \State $K \sim \text{MH}_{K}(\mathcal{B}_{K},\mathbf{S}_{K},\mathbf{b}_{\mathbf{S}_{K},K},\Gamma_{\mathbf{S}_{K},K})$
        %\State Update $\beta_{0}$
        \State Update $\eta$
        \EndFor
    \end{algorithmic}
\end{algorithm}

\paragraph{Predictive Inference.} 
Let $\hat{\theta}_{1},...,\hat{\theta}_{N}$ denote samples drawn from the posterior distribution.
The predictive distribution is then estimated as $\hat{p}(y|\mathbf{x}) = \sum_{i=1}^{N}p(y|\mathbf{x},\hat{\theta}_{i}) / N$.

\begin{remark}
When no prior information is available on the importance of input variables, we use a uniform distribution for $p_{\text{input}}$.
However, this noninformative choice often performs poorly when the dimension $p$ is large and higher-order interactions exist.
Our numerical studies in Section \ref{sec:p_input_exp} reveal that the choice of a good $p_{\text{input}}$ is important for exploring higher-posterior regions.
In practice, we could specify $p_{\text{input}}$ based on the importance measures of each input variable obtained by a standard method such as \cite{molnar2020interpretable}.
That is, we let $p_{\text{input}}(j) \propto \omega_j,$ where $\omega_j$ is an importance measure of the input variable $j \in [p]$.
In our numerical study, we use the global SHAP value \citep{molnar2020interpretable} based on a pretrained Deep Neural Network (DNN) for the importance measure or the feature importance using a pretrained eXtreme Gradient Boosting (XGB, \cite{chen2016xgboost}).
\end{remark}

\subsection{Posterior consistency}

For theoretical justification of Bayesian-TPNN, 
in this section, we prove the posterior consistency of Bayesian-TPNN.
To avoid unnecessary technical difficulties, we assume 
that $\phi(\mathbf{x}|\Theta_{k})$ in (\ref{eq:model}) satisfies
the sum-to-zero condition with respect to the uniform distribution. 
This can be done by using the uniform distribution instead of the empirical distribution in (\ref{eq:c_j}).

We assume that $(\mathbf{x}_{1},y_{1}),...,(\mathbf{x}_{n},y_{n})$ are realizations of 
independent copies $(\mathbf{X}_1,Y_1),\ldots, (\mathbf{X}_n,Y_n)$ of
$(\mathbf{X},Y)$ whose distribution $\mathbb{Q}_{0}$ is given as
\begin{align*}
\mathbf{X} \sim \mathbb{P}_{\mathbf{X}} \quad \text{and} \quad Y|\mathbf{X}=\mathbf{x} \sim \mathbb{Q}_{f_{0}(\mathbf{x}),1},
\end{align*}
where $f_{0}$ is the true regression function. We let $\eta=1$ for technical simplicity.
Suppose that
$f_{0}(\mathbf{x}) = \sum_{S \subseteq [p]}f_{0,S}(\mathbf{x}_{S}),$
where each $f_{0,S}$ satisfies the sum-to-zero condition with respect to the uniform distribution.
We denote $\mathbf{X}^{(n)}=\{\mathbf{X}_{1},...,\mathbf{X}_{n}\}$ and $Y^{(n)}=\{Y_{1},...,Y_{n}\}$.
Let $\pi_{\xi}(\cdot) \propto \pi(\cdot)\mathbb{I}(\Vert f \Vert_{\infty} \leq \xi)$, where $\pi(\cdot)$ is
the prior distribution of $f$ defined in Section \ref{sec:prior}.
Under regularity conditions \ref{eq:Assumption_1}, \ref{eq:Assumption_2}, \ref{eq:Assumption_5} and \ref{eq:Assumption_4} in Section \ref{app:regular_condition} of Appendix, Theorem \ref{thm:posterior_rate_component-wise} proves the posterior consistency of each component of Bayesian-TPNN.

\begin{theorem}[Posterior Consistency of Bayesian-TPNN]
\label{thm:posterior_rate_component-wise}
%Define $p^{\rm ind}_{\bold{X}}(\mathbf{x}) := \prod_{j=1}^p p_{X_j}(x_j),$ where
%$p_{X_j}$ is the density of $X_j$.
%In addition to the regularity conditions \ref{eq:Assumption_1}, \ref{eq:Assumption_2}, \ref{eq:Assumption_5} and \ref{eq:Assumption_4}, we assumes that
Assume that $0 < \inf_{\bold{x}\in \mathcal{X}} p_{\mathbf{X}}(\bold{x}) \leq \sup_{\bold{x}\in \mathcal{X}} p_{\mathbf{X}}(\bold{x}) <\infty,$
where $p_{\mathbf{X}}(\bold{x})$ is the density of $\mathbb{P}_{\mathbf{X}}.$ 
Then, there exists $\xi>0$ such that for any $\varepsilon >0,$ we have
 \begin{equation}
\pi_{\xi}\Big(f  : \Vert f_{0,S}-f_S \Vert_{2,n} > \varepsilon \Big| \mathbf{X}^{(n)},Y^{(n)} \Big) \xrightarrow{} 0
\end{equation}
for all $S\subseteq [p]$ in $\mathbb{Q}_{0}^{n}$ as $n \xrightarrow{} \infty,$
where $\pi_\xi(\cdot|\mathbf{X}^{(n)},Y^{(n)})$ is the posterior distribution of Bayesian-TPNN with the prior $\pi_\xi.$
\end{theorem}

%\begin{remark}
%If $\Vert f_{S} \Vert_{2,n}$ is small, the corresponding component can be regarded as irrelevant and removed.
%Furthermore, Theorem \ref{thm:posterior_rate_component-wise} ensures that such post-hoc component selection is consistent.
%The experimental results for this post-hoc component selection are presented in Section \ref{sec:exp_comp}.
%\end{remark}

% Theorem \ref{thm:posterior_rate_component-wise} establishes the component-wise posterior concentration rate of Bayesian-TPNN, highlighting one of its key properties. 
% This result enables effective a posteriori screening of unnecessary components.
% Specifically, for a given small positive constant $\delta$, removing $f_{S}$ if 
% $$
% \pi_{\xi}\Big(f : \Vert f_{S} \Vert_{2,n} > \epsilon_{n}\log n \Big| \mathbf{X}^{(n)},Y^{(n)}\Big) < \delta
% $$
% is guaranteed to be selection consistent by Theorem \ref{thm:posterior_rate_component-wise}.

\section{Experiments}
We present the results of the numerical experiments in this section, while further results and comprehensive details regarding the datasets, implementations of baseline models, and hyperparameter selections are provided in Sections \ref{sec:all_details_exper} to \ref{sec:unvertainty_other} of Appendix.

\subsection{Prediction performance}

\begin{table}[h]
    \caption{\footnotesize{\textbf{The averaged prediction accuracies (the standard errors) on real datasets.}}}
    \vskip -0.3cm
    \centering
    \scriptsize
    \setlength{\tabcolsep}{8pt}
    \renewcommand{\arraystretch}{1.0}
    \scalebox{0.9}{
    \begin{tabular}{c c c c c c c c c}
        \toprule
          & & \multicolumn{4}{c}{Interpretable model} & \multicolumn{3}{c}{Blackbox model}\\ \midrule
         Dataset & Measure & \makecell[c]{Bayesian\\TPNN} & \makecell[c]{ANOVA\\TPNN} & NAM & Linear & XGB & \makecell[c]{BART} & mBNN\\ \midrule
         \textsc{Abalone} \citep{abalone} & \multirow{4}{*}[-1.5em]{\makecell[c]{RMSE $\downarrow$\\(SE)} } & \makecell[c]{2.053\\(0.26)} & \makecell[c]{\textbf{2.051}\\(0.21)}& \makecell[c]{2.062\\(0.23)} & \makecell[c]{2.244\\(0.22)} & \makecell[c]{2.157\\(0.24)} & \makecell[c]{2.197\\(0.26)} &  \makecell[c]{2.081\\(0.24)}\\ 
         \textsc{Boston} \citep{harrison1978hedonic} & & \makecell[c]{\textbf{3.654}\\(0.49)} & \makecell[c]{3.671\\(0.56)} & \makecell[c]{3.832\\(0.67)} & \makecell[c]{5.892\\(0.77)} & \makecell[c]{4.130 \\ (0.56)} & \makecell[c]{4.073\\(0.67)} & \makecell[c]{4.277\\(0.51)} \\
         \textsc{Mpg} \citep{auto_mpg_9} & & \makecell[c]{\textbf{2.386}\\(0.41)} & \makecell[c]{2.623\\(0.38)} & \makecell[c]{2.755\\(0.41)} & \makecell[c]{3.748\\(0.41)} & \makecell[c]{2.531\\(0.26)} & \makecell[c]{2.699\\(0.43)} &  \makecell[c]{2.897\\(0.42)}\\
         \textsc{Servo} \citep{servo_87} & & \makecell[c]{0.351\\(0.02)} & \makecell[c]{0.594\\(0.04)} & \makecell[c]{0.802\\(0.04)} & \makecell[c]{1.117\\(0.04)} & \makecell[c]{0.314\\(0.04)} & \makecell[c]{0.342\\(0.04)} & \makecell[c]{\textbf{0.301}\\(0.04)}\\ \midrule
         \textsc{Fico} \citep{fico} & \multirow{4}{*}[-1.5em]{\makecell[c]{AUROC $\uparrow$\\(SE)}} & \makecell[c]{0.793\\(0.009)} & \makecell[c]{\textbf{0.802}\\(0.008)} & \makecell[c]{0.764\\(0.019)} & \makecell[c]{0.690\\(0.010)} & \makecell[c]{0.793\\(0.009)} & \makecell[c]{0.701\\(0.015)} & \makecell[c]{0.740\\(0.008)} \\
         \textsc{Breast} \citep{breast_cancer_wisconsin_(diagnostic)_17} &  & \makecell[c]{0.998\\(0.001)} & \makecell[c]{\textbf{0.998}\\(0.001)} & \makecell[c]{0.976\\(0.003)} & \makecell[c]{0.922\\(0.010)} & \makecell[c]{0.995\\(0.002)} & \makecell[c]{0.977\\(0.006)} & \makecell[c]{0.978\\(0.002)}\\
         \textsc{Churn} \citep{churn} &  & \makecell[c]{\textbf{0.849}\\(0.008)} & \makecell[c]{0.848\\(0.006)} & \makecell[c]{0.835\\(0.008)} & \makecell[c]{0.720\\(0.002)} & \makecell[c]{0.848\\(0.006)} & \makecell[c]{0.835\\(0.008)} & \makecell[c]{0.833\\(0.008)}\\
         \textsc{Madelon} \citep{madelon_171} & & \makecell[c]{0.854\\(0.013)} & \makecell[c]{0.587\\(0.013)} & \makecell[c]{0.644\\(0.005)} & \makecell[c]{0.548\\(0.011)} & \makecell[c]{\textbf{0.884}\\(0.006)} & \makecell[c]{0.751\\(0.011)} & \makecell[c]{0.650\\(0.018)} \\ \bottomrule
    \end{tabular}
    }
\vskip 0.3cm
    \label{tab:pred-performance}
\setlength{\tabcolsep}{8pt}
\renewcommand{\arraystretch}{1.0}
\caption{\footnotesize{\textbf{Comparison of Bayesian models in view of uncertainty quantification on real datasets.}}}
\vskip -0.3cm
\label{tab:uncertainty}

\begin{tabular}{c c c c c c c}
\toprule
 & \multicolumn{2}{c}{Bayesian-TPNN} & \multicolumn{2}{c}{BART } & \multicolumn{2}{c}{mBNN} \\ \midrule
Dataset & CRPS & NLL & CRPS & NLL & CRPS & NLL \\ \midrule
\textsc{Abalone} & \textbf{1.372} (0.19) & 2.260 (0.16) & 1.384 (0.18) & 2.261 (0.16) & 1.399 (0.16) & \textbf{2.226} (0.16) \\ 
\textsc{Boston}  & \textbf{2.202} (0.23) & 3.411 (0.37) & 2.623 (0.25) & \textbf{3.400} (0.42) & 3.144 (0.39) & 3.488 (0.26) \\ 
\textsc{Mpg}     & \textbf{1.510} (0.43) & \textbf{2.511} (0.21) & 1.553 (0.27) & 2.530 (0.20) & 2.142 (0.42) & 2.710 (0.24) \\ 
\textsc{Servo}   & 0.194 (0.01) & 0.836 (0.10) & 0.202 (0.02) & 0.849 (0.08) & \textbf{0.185} (0.02) & \textbf{0.321} (0.08) \\  \midrule
Dataset & ECE & NLL & ECE & NLL & ECE & NLL \\ \midrule
\textsc{Fico} & \textbf{0.036} (0.004) & \textbf{0.554} (0.007) & 0.054 (0.011) & 0.632 (0.012) & 0.219 (0.032) & 0.773 (0.046) \\ 
\textsc{Breast} & 0.129 (0.009) &  0.211 (0.014) & \textbf{0.118} (0.010) & \textbf{0.143} (0.032) & 0.292 (0.018) & 0.523 (0.025) \\ 
\textsc{Churn} & \textbf{0.031} (0.001) &  \textbf{0.418} (0.008) & 0.035 (0.001) & 0.430 (0.010) & 0.168 (0.037) & 0.531 (0.036) \\ 
\textsc{Madelon} & 0.076 (0.004) &  \textbf{0.478} (0.009) & \textbf{0.066} (0.004) & 0.685 (0.032) & 0.252 (0.020) & 0.840 (0.031) \\ \bottomrule
\end{tabular}

\end{table}

We compare the prediction performance of Bayesian-TPNN with baseline models including ANOVA-TPNN \citep{park2025tensor}, Neural Additive Models (NAM, \cite{agarwal2021neural}), Linear model, XGB \citep{chen2016xgboost}, Bayesian Additive Regression Trees (BART, \cite{chipman2010bart}, \cite{linero2025generalized}) and mBNN \citep{kong2023masked}.
We analyze eight real datasets and split each dataset into training and test sets with a ratio of 0.8 to 0.2.
This random split is repeated five times to obtain five prediction performance measures.

Table \ref{tab:pred-performance} reports the prediction accuracies (the Root Mean Square Error (RMSE) for regression tasks and the Area Under the ROC Curve (AUROC) for classification tasks) of the Bayes estimator of Bayesian-TPNN
along with those of its competitors,  where the best results are highlighted by \textbf{bold}.
Overall, Bayesian-TPNN achieves prediction performance comparable to that of the baseline models.
Further details of the experiments are provided in Section \ref{sec:exp_details_tabular} of Appendix.

Table \ref{tab:uncertainty} compares Bayesian-TPNN with the baseline Bayesian models in view of uncertainty quantification.
As uncertainty quantification measures, we consider Continuous Ranked Probability Score (CRPS, \cite{gneiting2007strictly}) and Negative Log-Likelihood (NLL) for regression tasks, and Expected Calibration Error (ECE, \cite{kumar2019verified}) together with NLL for classification tasks.
The results indicate that Bayesian-TPNN compares favorably with the baseline models in uncertainty quantification, which is a bit surprising since
Bayesian-TPNN is a transparent model while the other Bayesian models are black-box models.
The results of uncertainty quantification for non-Bayesian models are presented in Section \ref{sec:unvertainty_other_1} of Appendix, which are inferior to Bayesian models.

\subsection{Performance in component selection}
\label{sec:exp_comp}
\vskip -0.29cm
\begin{table}[h]
\caption{\footnotesize{\textbf{Performance of component selection on synthetic datasets.}}}
\label{tab:component-selection}
\vskip -0.3cm
\scriptsize
\centering
\begin{tabular}{c ccc ccc ccc}
\toprule
True model & \multicolumn{3}{c}{$f^{(1)}$} & \multicolumn{3}{c}{$f^{(2)}$} & \multicolumn{3}{c}{$f^{(3)}$} \\
\midrule
Order & \makecell[c]{Bayesian\\TPNN} & \makecell[c]{ANOVA\\T$^{2}$PNN} & NA$^{2}$M 
      & \makecell[c]{Bayesian\\TPNN} & \makecell[c]{ANOVA\\T$^{2}$PNN} & NA$^{2}$M
      & \makecell[c]{Bayesian\\TPNN} & \makecell[c]{ANOVA\\T$^{2}$PNN} & NA$^{2}$M \\
\midrule
1 & \makecell[c]{\textbf{1.000}\\(0.000)} & \makecell[c]{0.999\\(0.001)} & \makecell[c]{0.528\\(0.023)} 
  & \makecell[c]{0.831\\(0.008)} & \makecell[c]{\textbf{0.859}\\(0.010)} & \makecell[c]{0.417\\(0.015)} 
  & \makecell[c]{\textbf{1.000}\\(0.000)} & \makecell[c]{0.781\\(0.021)} & \makecell[c]{0.522\\(0.011)} \\ 
2 & \makecell[c]{\textbf{1.000}\\(0.000)} & \makecell[c]{0.978\\(0.007)} & \makecell[c]{0.508\\(0.024)} 
  & \makecell[c]{\textbf{0.985}\\(0.003)} & \makecell[c]{0.949\\(0.003)} & \makecell[c]{0.838\\(0.009)} 
  & \makecell[c]{\textbf{0.922}\\(0.019)} & \makecell[c]{0.704\\(0.007)} & \makecell[c]{0.542\\(0.017)} \\ 
3 & \makecell[c]{\textbf{0.740}\\(0.022)} & --- & --- 
  & \makecell[c]{\textbf{0.966}\\(0.018)} & --- & --- 
  & \makecell[c]{\textbf{0.661}\\(0.022)} & --- & --- \\ 
\bottomrule
\end{tabular}
\vskip 0.3cm
\caption{\footnotesize\textbf{Top 5 components: the important scores are normalized by their maximum.}}
\label{table:high_component_score}
\vskip -0.3cm
\scriptsize
\scalebox{0.9}{
\begin{tabular}{l cc cc cc cc cc}
\toprule
 & 
\multicolumn{2}{c}{Rank 1} & \multicolumn{2}{c}{Rank 2} & 
\multicolumn{2}{c}{Rank 3} & \multicolumn{2}{c}{Rank 4} & 
\multicolumn{2}{c}{Rank 5} \\
\midrule
Dataset & Component & Score & Component & Score & Component & Score & Component & Score & Component & Score \\
\midrule
%\textsc{Boston}  & 13 & 1.000 & 6 & 0.815 & 1 & 0.629 & 8 & 0.486 & (1,6) & 0.453 \\
\textsc{Madelon} & (49, 242, 319, 339) & 1.000 & (129, 443, 494) & 0.472 & (379, 443) & 0.374 & 106 & 0.322 & (242, 443) & 0.301 \\
\textsc{Servo}   & 1 & 1.000 & (1, 3, 4, 5) & 0.554 & 4 & 0.202 & (4, 6) & 0.193 & 8 & 0.173 \\
\bottomrule
\end{tabular}
}
\end{table}

We investigate whether Bayesian-TPNN identifies the true signal components well 
similarly to the setting in \cite{park2025tensor, tsang2017detecting}.
Synthetic datasets are generated from $Y = f^{(k)}(\mathbf{x})+\epsilon,\; k=1,2,3$, where $f^{(k)}$ is the true regression model and $\mathbf{x} \in \mathbb{R}^{50}$. Details of the experiment are described in Section \ref{sec:details_component_selec}.

We define the importance score of each component as its $\ell_2$ -norm, i.e., $\Vert f_S\Vert_{2,n}$.
A large $\Vert f_{S} \Vert_{2,n}$ implies $f_S$ is a signal.
Table \ref{tab:component-selection} reports the averages (standard errors) of AUROCs of the importance scores
obtained by Bayesian-TPNN, ANOVA-T$^{2}$PNN, and NA$^{2}$M for interaction order up to 3.
Note that  extending ANOVA-T$^{2}$PNN and NA$^{2}$M to
include the third order interactions requires additional $19,600$ neural networks, and so
we give up ANOVA-T$^{3}$PNN and NA$^{3}$M due to the limitations of our computational environment. 
Overall, Bayesian-TPNN achieves the best performance in component selection across orders and datasets, and detects higher-order interactions reasonably well.

Table \ref{table:high_component_score} presents the five most important components selected by Bayesian-TPNN on \textsc{Madelon} and \textsc{Servo} datasets.
We use these datasets as they highlight the performance gap between models with and without higher-order interactions.
Notably, Bayesian-TPNN identifies a 4th-order interaction as the most important component in the \textsc{Madelon} data, suggesting that its ability to capture higher-order interactions largely explains its superior prediction performance over ANOVA-TPNN on these datasets.
See Section \ref{sec:feature_descript_with_index} of Appendix for descriptions of the variables in \textsc{Madelon} and \textsc{Servo}.

\subsection{Interpretation of Bayesian-TPNN}
\label{sec:interpretation}

\begin{figure}[t]
\centering
    \centering
    \includegraphics[width=0.9\linewidth]{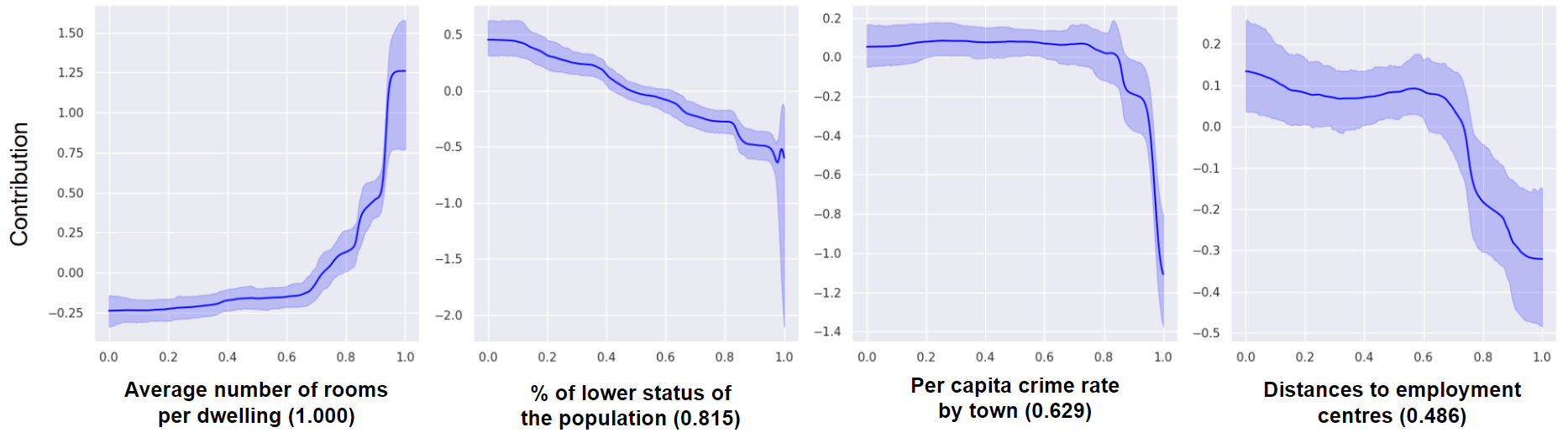}
    \vskip -0.3cm
    \caption{\footnotesize{\textbf{Plots of the functional relations of the important main effects estimated by Bayesian-TPNN on the \textsc{Boston} dataset.}} Each plot shows the Bayes estimate and 95\% credible interval of  each component. 
      Labels indicate the names of the input variables along with the normalized importance scores.}
    \label{fig:boston-main}
    \vskip -0.5cm
\end{figure}

The functional ANOVA model can provide various interpretations of the estimated prediction model through the estimated components as \cite{park2025tensor} illustrates.
In particular, by visualizing the estimated components, we can understand how each group of input variables
affects the response variable.
Figure \ref{fig:boston-main} presents the plots of the functional relations for the important main effects estimated by Bayesian-TPNN on the \textsc{Boston} dataset.
Each plot shows the Bayes estimate and the 95\% credible interval of the selected component.
The leftmost plot shows increasing trend, indicating that as the average number of rooms per dwelling increases, the price of the housing increases as well.
The second plot reveals a strictly decreasing relationship between the proportion of lower status of the population and the housing price.
The third plot indicates that housing prices decrease sharply once the crime rate exceeds a certain threshold.
The fourth plot shows that houses located farther from major employment centers are generally less expensive than those situated closer to such hubs.
More discussions about interpretation of Bayesian-TPNN are provided in Section \ref{sec:interpretable_more} of Appendix.

\subsection{Application to Concept Bottleneck Models}

Concept Bottleneck Model (CBM, \cite{koh2020concept}) is an interpretable model in which a CNN first receives an image and predicts its concepts.
These predicted concepts are then used to infer the target label, enabling explainable predictions.
To illustrate that Bayesian-TPNN can be amply combined with CBM,
we consider Independent Concept Bottleneck Models (ICBM, \cite{koh2020concept}), in which a CNN is first trained and then frozen, after which a final classifier is trained on the predicted concepts.
We compare Bayesian-TPNN with other baselines for learning the final classifier.  
In the experiment, we use \textsc{CelebA-HQ} \citep{CelebAMask-HQ} and \textsc{Catdog} \citep{jikadara2025dogcat} datasets, where we generate 5 concepts using GPT-5 \citep{openai2025gpt5}, and we obtain the concept labels for each image via CLIP \citep{radford2021learning}. 
The target labels for \textsc{CelebA-HQ} and \textsc{Catdog} are gender and cat/dog classification, respectively.
The details are provided in Section \ref{sec:details_image} of Appendix.

\begin{table}[h]
\centering
\scriptsize
\caption{\footnotesize\textbf{Prediction performance with CBM on image datasets.}}
\label{table:pred_perfor_image}
\vskip -0.3cm

% ---------- Concept = 5 ----------
\begin{tabular}{cccccc}
\toprule
 Dataset & Measure & Bayesian-TPNN & ANOVA-T$^{2}$PNN & NA$^{2}$M & Linear \\ \midrule
 \textsc{CelebA-HQ} & AUROC $\uparrow$ & \textbf{0.936} (0.002) & 0.923 (0.002) & 0.922 (0.002) & 0.893 (0.003)   \\ 
 \textsc{CatDog} & AUROC $\uparrow$ & \textbf{0.878} (0.002) & 0.853 (0.002) & 0.851 (0.002) & 0.711 (0.001) \\ 
\bottomrule
\end{tabular}
\end{table}

Table \ref{table:pred_perfor_image} presents the averages and standard errors of AUROCs when Bayesian-TPNN, ANOVA-T$^{2}$PNN, NA$^{2}$M, and Linear model
are used in the final classifier. Among them, Bayesian-TPNN attains the highest prediction performance, which can be attributed to its capability to estimate higher-order components.

% \paragraph{Fewer concepts, better prediction performance.}

\section{Conclusion}

We proposed Bayesian-TPNN, a novel Bayesian neural network for the functional ANOVA model that can detect higher-order signal components effectively
and thus achieve superior prediction performance in view of prediction accuracy and uncertainty quantification.
In addition, Bayesian-TPNN is also theoretically sound since it achieves the posterior consistency.

We used a predefined distribution $p_{\text{input}}$ for the selection probability of each input variable in the MH algorithm.
It would be interesting to update $p_{\text{input}}$ along with the other parameters. For example, it would
be possible to let $p_{\text{input}}(j)$ be proportional to the number of basis functions in the current Bayesian-TPNN model which
uses $x_j$. This would be helpful when $p$ is large. We will pursue this algorithm in the near future.

\paragraph{Reproducibility Statement.}

We have made significant efforts to ensure the reproducibility of our results.
The source code implementing our proposed model and experiments is provided in the supplementary material.
Detailed descriptions of the experimental setup, hyperparameters and datasets are provided in Section \ref{sec:all_details_exper} of Appendix. Additional ablation studies are reported in Section \ref{sec:ablat_exp} of Appendix.

\section*{Acknowledgements}
This work was supported by the National Research Foundation of Korea(NRF) grant funded by the Korea government(MSIT) (RS-2025-00556079) and by Institute of Information \& communications Technology Planning \& Evaluation (IITP) grant funded by the Korea government(MSIT) [NO.RS-2021-II211343, Artificial Intelligence Graduate School Program (Seoul National University)].

\bibliography{iclr2026_conference}
\bibliographystyle{iclr2026_conference}

\clearpage
\appendix

\begin{center}
    {\LARGE \textsc{Appendix}}
\end{center}

\section{Details of the MCMC algorithm}
\label{sec:details_mcmc}
For given data $\mathcal{D}^{(n)}$, we denote $\mathbf{x}^{(n)} = \{\mathbf{x}_{1},...,\mathbf{x}_{n}\}$.
Let $\omega_j=p_{\text{input}}(j)$.

\subsection{Sampling $K$ via MH algorithm}
\label{sec:K_MH}

\subsubsection{Case of \(K^\textup{new} = K+1\)}
\label{sec:add-K}
From current state $\theta = \left(K, \mathbf{S}_K, \mathbf{b}_{\mathbf{S}_K, K}, \Gamma_{\mathbf{S}_K, K}, \mathcal{B}_K, \eta\right)$, we propose a new state $\theta^{\text{new}}$ using one of \{\textbf{Random}, \textbf{Stepwise}\}.
Here, $\theta^\textup{new}$ is defined as
\begin{align*}
    \theta^\textup{new} &= (K+1, \mathbf{S}_{K+1}, \mathbf{b}_{\mathbf{S}_{K+1}, K+1}, \Gamma_{\mathbf{S}_{K+1},K+1}, \mathcal{B}_{K+1},\eta),
\end{align*}
where 
\begin{align*}
\mathbf{S}_{K+1} &= (\mathbf{S}_K, S^{\text{new}}_{K+1}), \\
\mathbf{b}_{\mathbf{S}_{K+1}, K+1} &=(\mathbf{b}_{\mathbf{S}_K, K}, \mathbf{b}_{S^{\text{new}}_{K+1},K+1}), \\
\Gamma_{\mathbf{S}_{K+1}, K+1} &= (\Gamma_{\mathbf{S}_K,K},\Gamma_{S^{\text{new}}_{K+1}, K+1}),\\
\mathcal{B}_{K+1} &= (\mathcal{B}_K, \beta^{\text{new}}_{K+1}).
\end{align*}

We accept the new state $\theta^{\text{new}}$ with probability
\begin{align*}
    P_{\text{accept}} = \min\left\{ 1, \prod_{i=1}^n\frac{q_{f_{\theta^\textup{new}}(\mathbf{x}_i), \eta}(y_i)}{q_{f_\theta(\mathbf{x}_i),\eta}(y_i)} \frac{\pi(\theta^\textup{new})}{\pi(\theta)}\frac{q(\theta\vert\theta^\textup{new})}{q(\theta^\textup{new}\vert\theta)}\right\},
\end{align*}
where 
\begin{align*}
f_\theta(\mathbf{x}) = \sum_{k\in[K]}\beta_k\phi(\mathbf{x}\vert S_k,\mathbf{b}_{S_k,k}, \Gamma_{S_k,k})    
\end{align*}
and
\begin{align*}
f_{\theta^\textup{new}}(\mathbf{x})= f_\theta(\mathbf{x}) + \beta_{K+1}^{\text{new}}\phi(\mathbf{x}\vert S_{K+1}^{\text{new}}, \mathbf{b}_{S_{K+1}^{\text{new}},K+1},\Gamma_{S_{K+1}^{\text{new}},K+1}).
\end{align*}

To compute the acceptance probability, we calculate the prior ratio $\pi(\theta^\textup{new})/\pi(\theta)$, and then the proposal ratio $q(\theta\vert\theta^\textup{new})/q(\theta^\textup{new}\vert\theta)$.

\paragraph{Prior Ratio.}

The prior ratio is given as

\begin{small}
\begin{align*}
    \frac{\pi(\theta^\textup{new})}{\pi(\theta)} &= \frac{\pi(K+1)\pi(\mathbf{S}_{K+1}\vert K+1)\pi(\mathbf{b}_{\mathbf{S}_{K+1}, K+1}\vert \mathbf{S}_{K+1})\pi\left(\Gamma_{\mathbf{S}_{K+1}, K+1}\vert \mathbf{S}_{K+1}\right)\pi\left(\mathcal{B}_{K+1}\vert K+1\right)}{\pi(K)\pi(\mathbf{S}_{K}\vert K)\pi(\mathbf{b}_{\mathbf{S}_{K}, K}\vert \mathbf{S}_{K})\pi\left(\Gamma_{\mathbf{S}_{K}, K}\vert \mathbf{S}_{K}\right)\pi\left(\mathcal{B}_{K}\vert K\right)}\\
    &= \frac{\pi(S_{K+1})\pi(\mathbf{b}_{S_{K+1}, K+1})\pi(\Gamma_{S_{K+1},K+1})\pi(\beta_{K+1})}{\exp(C_0\log n)}.
\end{align*}
\end{small}
\begin{align*}
\end{align*}

\paragraph{Proposal Ratio.}
For $q(\theta\vert\theta^\textup{new})$, we have
\begin{align*}
q(\theta\vert\theta^\textup{new}) &= \text{Pr}(K = K^{\text{new}}-1)\text{Pr}(\text{Choose one of $K^{\text{new}}$ TPNNs for deletion})\\
&=\frac{K^{\text{new}}}{K_\text{max}}\frac{1}{K^{\text{new}}}.
\end{align*}

For a given $\theta$, a new state \(\theta^\textup{new}\) is proposed in two ways: (1) \textbf{Random} move or (2) \textbf{Stepwise} move.

For \textbf{Random} move, we have
\begin{align}
q(\theta^{\text{new}}|\theta,\textbf{Random}) = \pi(S_{K^{\text{new}}})\pi(\mathbf{b}_{S^{\text{new}}_{K+1},K+1})\pi(\Gamma_{S^{\text{new}}_{K+1},K+1})\pi(\beta^{\text{new}}_{K+1}).
\end{align}

For \textbf{Stepwise} move, we have
\begin{align*}
q(\theta^{\text{new}}|\theta,\textbf{Stepwise}) &= 
%\text{Pr}(\text{Choose $S_{j^{\text{new}}}$ in $\mathbf{S}_{K}$ and adding one variable $k^{\text{new}}$ from $S_{j^{\text{new}}}^{c}$})\\
\text{Pr}(S_{K+1}^{\text{new}})\pi(\mathbf{b}_{S^{\text{new}}_{K+1},K+1})\pi(\Gamma_{S^{\text{new}}_{K+1},K+1})\pi(\beta^{\text{new}}_{K+1}).
\end{align*}
Here, $\text{Pr}(S_{K+1}^{\text{new}})$ is defined as 
\begin{align*}
\text{Pr}(S_{K+1}^{\text{new}}) &=  \sum_{k=1}^{K}\text{Pr}(\text{Choose}\:S_{k}\: \text{from}\:\mathbf{S}_{K})\text{Pr}(S_{K+1}^{\text{new}} = S_{k}\cup \{ j^{\text{new}}\}, j^{\text{new}} \in S_{k}^{c}) \\
&= \sum_{k=1}^{K}{1\over K}\mathbb{I}(\exists j^{\text{new}} \in S_{k}^{c} \: \text{s.t} \: S_{k}\cup \{j^{\text{new}}\} = S_{K+1}^{\text{new}})\frac{\omega_{j_{\text{new}}}}{\sum_{l\in S_k^c}\omega_l}.
\end{align*}

To sum up, we have
\begin{align*}
q(\theta^{\text{new}}|\theta) = q(\theta^{\text{new}}|\theta,\textbf{Random})\text{Pr}(\textbf{Random}) + q(\theta^{\text{new}}|\theta,\textbf{Stepwise})\text{Pr}(\textbf{Stepwise}).
\end{align*}

\subsubsection{Case of \(K^\textup{new} = K-1\)}
Since the acceptance probability of the case $K^{\text{new}}=K-1$ can be easily computed by reversing the steps in Section \ref{sec:add-K}, we omit the details here.

% For a uniformly randomly chosen $k \in [K_{\max}]$, define the proposed state $\mathbf{z}^{\text{new}}$ by setting $z_{k}^{\text{new}} = 1 - z_{k}$ and $z_{j}^{\text{new}} = z_{j}$ for all $j \neq k$.
% Finally, we accept $\mathbf{z}^{\text{new}}$ with probability $P_{\text{accept}}$ given as
% \begin{align*}
% P_{\text{accept}} = \min \Bigg\{1, {\mathcal{L}_{\mathbf{z}^{\text{new}}}(S_{k},\mathbf{b}_{S_{k},k},\Gamma_{S_{k},k},\beta_{k},\pmb{\lambda}_{k},\eta) \over \mathcal{L}_{\mathbf{z}}(S_{k},\mathbf{b}_{S_{k},k},\Gamma_{S_{k},k},\beta_{k},\pmb{\lambda}_{k},\eta)}\cdot {\exp(-C_{0}\Vert \mathbf{z}^{\text{new}}) \Vert_{1}\log n) \over \exp(-C_{0}\Vert \mathbf{z}) \Vert_{1}\log n})\cdot { {K_{\max} \choose \Vert \mathbf{z}^{\text{new}} \Vert_{1} } \over  {K_{\max} \choose \Vert \mathbf{z} \Vert_{1} } } \Bigg \}.
% \end{align*}

\subsection{Sampling $S_{k},\mathbf{b}_{k},\Gamma_{k}$ via MH algorithm}
\label{sec:s_algorithm}

Here, we consider three moves - \{\textbf{Adding, Deleting} and \textbf{Changing}\}. 
Each move is chosen with the probabilities $\text{Pr}(\textbf{Adding})=q_{\text{add}}, \text{Pr}(\textbf{Deleting})=q_{\text{delete}}, \text{Pr}(\textbf{Changing})=q_{\text{change}}$, respectively.

In \textbf{Adding} move, the proposal distribution generates $S_{k}^{\text{new}} = S_{k} \cup \{j^{\text{adding}}\}$, where $j^{\text{adding}} \in [p]\backslash S_{k}$ is chosen with a given weight vector $\pmb{\omega}:=(\omega_{1},...,\omega_{p})$. 
Note that the likelihood cannot be calculated using $S_{k}^{\text{new}}$ alone, where $S_{k}^{\text{new}}$ is the index set generated by the proposal distribution. 
To address this, we also generate $b_{j^{\text{adding}},k}$ and $\gamma_{j^{\text{adding}},k}$ from $\text{Uniform}(0,1)$ and $\text{Gamma}(a_{\gamma},b_{\gamma})$, respectively.

Furthermore, in \textbf{Deleting} move, a variable to be deleted is uniformly selected from $S_{k}$ and the new component $S_k^\textup{new} = S_k\backslash\{j^\textup{deleting}\}$ is proposed accordingly. This move also involves removing the associated numeric parameters $b_{j^{\text{deleting}},k}$ and $\gamma_{j^{\text{deleting}},k}$ from $\mathbf{b}_{S_k,k}$ and $\Gamma_{S_k,k}$, respectively.

Finally, in $\textbf{Changing}$ move, we choose an element $j^\textup{change}$ in $S_{k}$ and replace it with a randomly selected $j^{\text{new}} \in S_{k}^{c}$. 
The corresponding $b_{j^{\text{change}},k}$ and $\gamma_{j^{\text{change}},k}$ are then replaced by new values generated from $\text{Uniform}(0,1)$ and $\text{Gamma}(a_{\gamma},b_{\gamma})$, respectively. This move results in $S_k^\textup{new}= (S_k\backslash\{j^\textup{change}\}) \cup \{j^\textup{new}\}$.

Here, \textbf{Adding} and \textbf{Deleting} affect the dimensions of $\mathbf{b}_{S_k,k}$ and $\Gamma_{S_k,k}$, thus the algorithm corresponds to RJMCMC (\cite{green1995rjmcmc}) which requires Jacobian computations.
However, since we applied the identity transformation on the auxiliary variables which are generated to match the dimensions, the Jacobian is simply 1. This allows us to easily compute the acceptance probability.

\subsubsection{Transition probability for proposal distribution}
For a given weight vector $\pmb{\omega}$, the proposal distributions $q_{\pmb{\omega}}$ of $\Theta_k^\textup{new} = (S_{k}^{\text{new}}, \mathbf{b}_{S_{k}^{\text{new}},k}, \Gamma_{S_{k}^{\text{new}},k})$ are defined as:

\begin{align*}
&q_{\pmb{\omega}}(\Theta_k^\textup{new}\vert \Theta_k,\textbf{Adding}) = {\omega_{j^{\text{adding}}}\over \sum_{j \in S_{k}^{c}}\omega_{j}} \pi(b_{j^{\text{adding}},k})\pi(\gamma_{j^{\text{adding}},k}) \\
& q_{\pmb{\omega}}(\Theta_k^\textup{new}\vert, \Theta_k,\textbf{Deleting}) = {1\over \vert S_k\vert} \\
&q_{\pmb{\omega}}(\Theta_k^\textup{new}\vert \Theta_k,\textbf{Changing}) = {1\over \vert S_k\vert}{\omega_{j^{\text{new}}}\over \sum_{j \in S_{k}^{c}}\omega_{j}}  \pi(b_{j^{\text{new}},k})\pi(\gamma_{j^{\text{new}},k}).
\end{align*}

To sum up, we have
\begin{align*}
q_{\pmb{\omega}}(\Theta_{k}^{\text{new}}|\Theta_{k}) &= q_{\pmb{\omega}}(\Theta_k^\textup{new}\vert \Theta_k,\textbf{Adding})\text{Pr}(\textbf{Adding}) \\
&\:\:\:\:+q_{\pmb{\omega}}(\Theta_k^\textup{new}\vert \Theta_k,\textbf{Deleting})\text{Pr}(\textbf{Deleting})\\
&\:\:\:\:+q_{\pmb{\omega}}(\Theta_k^\textup{new}\vert \Theta_k,\textbf{Changing})\text{Pr}(\textbf{Changing}).
\end{align*}

% We adopt a Gibbs sampling scheme in which parameters are updated one at a time while keeping all other parameters fixed. For each parameter, we employ a Metropolis-Hastings (MH) algorithm to draw samples from its corresponding conditional posterior distribution.

% To perform sampling with the MH algorithm, we need to calculate the acceptance probability
% $$\alpha(\theta^\textup{new};\theta) = \min\left(1, \frac{\mathbb{Q}(\mathcal{D}\vert \theta^{\text{new}}) \pi(\theta^{\text{new}})q(\theta\vert\theta^{\text{new}})}{\mathbb{Q}(\mathcal{D}\vert \theta)\pi(\theta) q(\theta^{\text{new}}\vert\theta)}\right)$$
% where \(\mathbb{Q}(\mathcal{D}\vert\theta)\) is the likelihood, \(\pi(\theta)\) is the prior, and \(q(\cdot\vert\cdot)\) denotes the proposal distribution. Note that the computation of likelihood ratio depends on the assumed distribution of \(y\). This section provides some detailed explanations about how we sampled the parameters from the posterior distribution.

\subsubsection{Posterior Ratio}
We define $\pmb{\lambda}_k := (\lambda_{k, 1},\dots,\lambda_{k,n})$ where $\lambda_{k,i} = \sum_{j \neq k}\beta_{j}\phi(\mathbf{x}_{i}|\Theta_j)$ for $i=1,...,n$ and the likelihood \(\mathcal{L}(\Theta_k,\beta_{k},\pmb{\lambda}_{k},\eta) := \prod_{i=1}^{n}q_{\lambda_{k,i} + \beta_{k}\phi(\mathbf{x}_{i}|\Theta_k),\eta}(y_{i})\).

Then, we have
\begin{align*}
\pi(\Theta_k|\beta_{k},\pmb{\lambda}_{k},\mathcal{D}^{(n)},\eta) &\propto \pi(y_{1},...,y_{n}|\Theta_k,\beta_{k},\pmb{\lambda}_{k},\mathbf{x}^{(n)},\eta)\pi(\Theta_k) \\
&= \mathcal{L}(\Theta_k, \beta_k, \pmb{\lambda}_k, \eta) \pi(\Theta_k).
\end{align*}

\vskip 0.2cm

Thus the posterior ratio of $\Theta_k^\textup{new} = (S_k^\textup{new}, \mathbf{b}_{S_k^\textup{new},k},\Gamma_{S_k^\textup{new},k})$ to $\Theta_k=(S_k, \mathbf{b}_{S_k,k},\Gamma_{S_k,k})$ is given as

\begin{align*}
{\pi(\Theta_k^\textup{new}|\beta_{k},\pmb{\lambda}_{k},\mathcal{D}^{(n)},\eta) \over \pi(\Theta_k|\beta_{k},\pmb{\lambda}_{k},\mathcal{D}^{(n)},\eta)} = {\mathcal{L}(\Theta_k^\textup{new},\beta_{k},\pmb{\lambda}_{k},\eta) \over \mathcal{L}(\Theta_k,\beta_{k},\pmb{\lambda}_{k},\eta)} \frac{\pi(\Theta_k^\textup{new})}{\pi(\Theta_k)}.
\end{align*}

\vskip 0.3cm

\subsubsection{Acceptance probability}
\label{sec:accept_probability}
In this section, for notational simplicity, we denote the hyperparameters \(\alpha_{\text{adding}}\) and \(\gamma_{\text{adding}}\) as \(\alpha\) and \(\gamma\), respectively.

For a proposed new state $\Theta_k^\textup{new}$, we accept it with probability
\begin{align*}
    P_{\text{accept}} &= \min\left\{1, \frac{\pi(\Theta_k^\textup{new}\vert \beta_k, \pmb{\lambda}_k, \mathcal{D}^{(n)}, \eta)}{\pi(\Theta_k\vert \beta_k, \pmb{\lambda}_k, \mathcal{D}^{(n)}, \eta)}\frac{q_{\pmb{\omega}}(\Theta_k\vert \Theta_k^\textup{new})}{q_{\pmb{\omega}}(\Theta_k\vert \Theta_k^\textup{new})} \right\}\\
    &=\min\left\{1, {\mathcal{L}(\Theta_k^\textup{new},\beta_{k},\pmb{\lambda}_{k},\eta) \over \mathcal{L}(\Theta_k,\beta_{k},\pmb{\lambda}_{k},\eta)} \frac{\pi(\Theta_k^\textup{new})}{\pi(\Theta_k)}\frac{q_{\pmb{\omega}}(\Theta_k\vert \Theta_k^\textup{new})}{q_{\pmb{\omega}}(\Theta_k^\textup{new}\vert \Theta_k)}\right\}.
\end{align*}
\vskip 0.1cm

Now, we will show how the product of the prior and proposal ratios is calculated in the case of \textbf{Adding}, \textbf{Deleting}, and \textbf{Changing}.

For \textbf{Adding}, we have
\begin{align*}
&\frac{\pi(\Theta_k^\textup{new})}{\pi(\Theta_k)}\frac{q_{\pmb{\omega}}(\Theta_k\vert \Theta_k^\textup{new})}{q_{\pmb{\omega}}(\Theta_k^\textup{new}\vert \Theta_k)}\\
% &= {\pi(S_{k}^{\text{new}}) \over \pi(S_{k})} \cdot {|S_{k}|\over |S_{k}^{c}|} \\
&= \alpha_{}\vert S_k^\textup{new}\vert^{-\gamma}\frac{1-\alpha(1+\vert S_k^\textup{new}\vert)^{-\gamma}}{1-\alpha\vert S_k^\textup{new}\vert^{-\gamma}}\frac{1}{p-\vert S_k^\textup{new}\vert+1} \frac{\text{Pr}(\textbf{Deleting})}{\text{Pr}(\textbf{Adding})} \frac{\sum_{l\in S_k^c }\omega_l}{\omega_{j^\textup{adding}}}.
\end{align*}

For \textbf{Deleting}, we have
\begin{align*}
&\frac{\pi(\Theta_k^\textup{new})}{\pi(\Theta_k)}\frac{q_{\pmb{\omega}}(\Theta_k\vert \Theta_k^\textup{new})}{q_{\pmb{\omega}}(\Theta_k^\textup{new}\vert \Theta_k)}\\
&= \frac{1}{\alpha (1+\vert S_k^\textup{new}\vert)^{-\gamma}}\frac{1-\alpha(1+\vert S_k^\textup{new}\vert)^{-\gamma}}{1-\alpha(2+\vert S_k^\textup{new}\vert)^{-\gamma}} (p-\vert S_k^\textup{new}\vert) \frac{\text{Pr}(\textbf{Adding})}{\text{Pr}(\textbf{Deleting})}\frac{\omega_{j^\textup{deleting}}}{\sum_{l\in S_k^c}\omega_l}.
\end{align*}

For \textbf{Changing}, we have
\begin{align*}
\frac{\pi(\Theta_k^\textup{new})}{\pi(\Theta_k)}\frac{q_{\pmb{\omega}}(\Theta_k\vert \Theta_k^\textup{new})}{q_{\pmb{\omega}}(\Theta_k^\textup{new}\vert \Theta_k)}= \frac{\omega_{j^\textup{change}} \sum_{l\in S_k^c}\omega_l}{\omega_{j^\textup{new}}\sum_{l\in (S_k^\textup{new})^c}\omega_l}.
\end{align*}

% Finally, we accept $(S_{k}^{\text{new}},\mathbf{b}_{S_{k}^{\text{new}},k},\Gamma_{S_{k}^{\text{new}},k})$ with probability $P_{\text{accept}}$ given as
% \begin{align*}
% P_{\text{accept}} = \Bigg \{ 1, {\pi(S_{k}^{\text{new}},\mathbf{b}_{S_{k}^{\text{new}},k},\Gamma_{S_{k}^{\text{new}},k}|\beta_{k},\pmb{\lambda}_{k},\mathcal{D}^{(n)},\eta) \over \pi(S_{k},\mathbf{b}_{S_{k},k},\Gamma_{S_{k},k}|\beta_{k},\pmb{\lambda}_{k},\mathcal{D}^{(n)},\eta)} \cdot {q(S_{k}, \mathbf{b}_{S_{k},k} \Gamma_{S_{k},k} |S_{k}^{\text{new}}, \mathbf{b}_{S_{k}^{\text{new}},k} \Gamma_{S_{k}^{\text{new}},k}) \over q(S_{k}^{\text{new}}, \mathbf{b}_{S_{k}^{\text{new}},k} \Gamma_{S_{k}^{\text{new}},k} |S_{k}, \mathbf{b}_{S_{k},k} \Gamma_{S_{k},k})} \Bigg \}. 
% \end{align*}

\subsection{Sampling $\mathbf{b}_{S_{k},k}$, $\Gamma_{S_{k},k}$ and $\beta_{k}$ via MH algorithm}
\label{sec:b-gam-mh}

We use Langevin Dynamics (\cite{rossky1978brownian}) as a proposal distribution for $\mathbf{b}_{S_{k},k}$, $\Gamma_{S_{k},k}$ and $\beta_{k}$.
That is, $\mathbf{b}_{S_{k},k}^{\text{new}}$, $\Gamma_{S_{k},k}^{\text{new}}$ and $\beta_{k}^{\text{new}}$ are proposed as
\begin{align*}
(\mathbf{b}_{S_{k},k}^{\text{new}}, \Gamma_{S_{k},k}^{\text{new}}, \beta_k^\textup{new}) &= (\mathbf{b}_{S_{k},k}, \Gamma_{S_{k},k}, \beta_k) + {\epsilon^{2} \over 2}U(\mathbf{b}_{S_{k},k}, \Gamma_{S_{k},k}, \beta_k) + \epsilon \mathbb{M},
\end{align*}
where 
\begin{align*}
U(\mathbf{b}_{S_{k},k}, \Gamma_{S_{k},k}, \beta_k) = \nabla_{(\mathbf{b}_{S_k,k},\Gamma_{S_k,k}, \beta_k)}\log\pi(\mathbf{b}_{S_k,k},\Gamma_{S_k,k}, \beta_k\vert \bm{\lambda}_k,S_k,\mathcal{D}^{(n)},\eta).
\end{align*}
Here, $\mathbb{M} \sim N(0,\mathbf{I})$, where $\mathbf{I}$ is the \((2\vert S_k\vert+1)\times (2\vert S_k\vert+1)\) identity matrix and $\epsilon > 0 $ is a step size.

\vskip 0.1cm

We accept the proposal $(\mathbf{b}_{S_{k},k}^{\text{new}},\Gamma_{S_{k},k}^{\text{new}},\beta_{k}^{\text{new}})$ with a probability $P_{\text{accept}}$ given as

\begin{align*}
P_{\text{accept}} = \bigg \{ 1, {\mathcal{L}(S_{k},\mathbf{b}_{S_{k},k}^{\text{new}},\Gamma_{S_{k},k}^{\text{new}},\beta_{k}^{\text{new}},\pmb{\lambda}_{k},\eta) \over \mathcal{L}(S_{k},\mathbf{b}_{S_{k},k},\Gamma_{S_{k},k},\beta_{k},\pmb{\lambda}_{k},\eta)} {\pi(\mathbf{b}_{S_{k},k}^{\text{new}}) \over \pi(\mathbf{b}_{S_{k},k})}{\pi(\Gamma_{S_{k},k}^{\text{new}}) \over \pi(\Gamma_{S_{k},k})} {\pi(\beta_k^{\text{new}}) \over \pi(\beta_k)}  \exp\bigg(-{1\over 2}(\Vert \mathbb{M}^{\text{new}} \Vert_{2}^{2} - \Vert \mathbb{M} \Vert_{2}^{2}) \bigg) \bigg \},
\end{align*}

where \(\Vert\cdot\Vert_2\) is the Euclidean norm for a vector and

\begin{align*}
\mathbb{M}^{\text{new}} = \mathbb{M} + {\epsilon \over 2}U(\mathbf{b}_{S_{k},k}, \Gamma_{S_{k},k}, \beta_k) + {\epsilon \over 2}U(\mathbf{b}_{S_{k},k}^{\text{new}}, \Gamma_{S_{k},k}^{\text{new}},\beta_k^\textup{new}).
\end{align*}

For
$\nabla_{(\mathbf{b}_{S_k,k},\Gamma_{S_k,k}, \beta_k)}\log\pi(\mathbf{b}_{S_k,k},\Gamma_{S_k,k},\beta_k | \pmb{\lambda}_k,S_k,\mathcal{D}^{(n)},\eta)$, we will calculate 

\begin{align*}
\nabla_{\mathbf{b}_{S_k,k}}\log\pi(\mathbf{b}_{S_k,k}, | \pmb{\lambda}_k,\beta_k,S_k,\Gamma_{S_k,k}, \mathcal{D}^{(n)},\eta),
\end{align*}
\begin{align*}
\nabla_{\Gamma_{S_k,k}}\log\pi(\Gamma_{S_k,k} | \pmb{\lambda}_k,\beta_k,S_k,\mathbf{b}_{S_k,k},\mathcal{D}^{(n)},\eta),
\end{align*}
and
\begin{align*}
\nabla_{\beta_k}\log\pi(\beta_k |\pmb{\lambda}_k, S_k, \mathbf{b}_{S_k,k}, \Gamma_{S_k,k},\mathcal{D}^{(n)},\eta).
\end{align*}

\subsubsection{Calculating the Gradient of the Log-Posterior with respect to $\mathbf{b}_{S_k,k}$}\label{sec:grad-b}
Without loss of generality, let $S_{k} = \{1,...,d\}$.

Since
\begin{align*}
    \pi(\mathbf{b}_{S_k,k}\vert \pmb{\lambda}_k, \beta_k, S_k,\Gamma_{S_k,k},\mathcal{D}^{(n)}, \eta) &\propto \mathcal{L}(\pmb{\lambda}_k,\beta_k, S_k, \mathbf{b}_{S_k,k},\Gamma_{S_k,k},\eta),
\end{align*}
the $j$-th gradient is given as
\begin{align*}
    \frac{\partial}{\partial b_{j,k}}\log \pi(\mathbf{b}_{S_k,k}\vert \pmb{\lambda}_k,\beta_k,S_k,\Gamma_{S_k,k},\mathcal{D}^{(n)},\eta) &= \frac{\partial}{\partial b_{j,k}}\sum_{i=1}^{n} \log q_{f(\mathbf{x}_{i}),\eta}(y_{i})\nonumber,
\end{align*}
where 
$f(\mathbf{x}_i)=\lambda_{k,i}+\beta_k\prod_{l\in S_k}\phi(x_{i,l}\vert \{l\},b_{l,k},\gamma_{l,k})$.

In turn, we have
\begin{align*}
&\frac{\partial}{\partial b_{j,k}}\sum_{i=1}^{n} \log q_{f(\mathbf{x}_{i}),\eta}(y_{i}) \\
&= \sum_{i=1}^{n} \left(\frac{\partial\log q_{f(\mathbf{x}_{i}),\eta}(y_{i})}{\partial f(\mathbf{x}_i)}\frac{\partial f(\mathbf{x}_i)}{\partial b_{j,k}}\right)\label{eq:b_score1} \\
&=\beta_k \sum_{i=1}^{n}\left(\frac{\partial\log q_{f(\mathbf{x}_{i}),\eta}(y_{i})}{\partial f(\mathbf{x}_i)}\frac{\partial \phi(x_{i,j}|\{j\}, b_{j,k},\gamma_{j,k})}{\partial b_{j,k}}\prod_{l\neq j, l \in S_{k}}\phi(x_{i,l}|\{l\},b_{l,k},\gamma_{l,k})\right).
\end{align*}

Here,
\begin{align*}
    \phi(x_{i, j}\vert \{j\},b_{j,k},\gamma_{j,k}) &= 1-\sigma\left(\frac{x_{i,j}-b_{j,k}}{\gamma_{j,k}}\right)+c_j(b_{j,k},\gamma_{j,k})\sigma\left(\frac{x_{i,j}-b_{j,k}}{\gamma_{j,k}}\right),\\
    c_j(b_{j,k},\gamma_{j,k}) &= -\Big( 1-\tilde{c}_j(b_{j,k},\gamma_{j,k})\Big) \Big/\tilde{c}_j(b_{j,k},\gamma_{j,k}),
\end{align*}
where \(\tilde{c}_j(b,\gamma) := \int_{\mathcal{X}_{j}}\sigma\left({u-b\over \gamma}\right)\mu_{n,j}(du)\).

Then, we have
\begin{align*}
     \frac{\partial \phi(x_{i,j}\vert\{j\}, b_{j,k},\gamma_{j,k})}{\partial b_{j,k}} =&  -\frac{1}{\gamma_{j,k}}\sigma\left(\frac{x_{i,j}-b_{j,k}}{\gamma_{j,k}}\right)\int_{\mathcal{X}_j}\tilde{\sigma}\left(\frac{u-b_{j,k}}{\gamma_{j,k}}\right)\mu_{n,j}(du) \\
     &+ \frac{1}{\gamma_{j,k}\tilde{c}_j(b_{j,k},\gamma_{j,k})}\tilde{\sigma}\left(\frac{x_{i,j}-b_{j,k}}{\gamma_{j,k}}\right),
\end{align*}
where \(\tilde{\sigma}(x) := \sigma(x)(1-\sigma(x))\).

\subsubsection{Calculating the Gradient of the Log-Posterior with respect to $\Gamma_{S_k,k}$}

Without loss of generality, we let $S_{k} = \{1,...,d\}$.
Similarly to Section \ref{sec:grad-b} of Appendix, we can derive the gradient of the log posterior with respect to $\gamma_{j,k}$ as
\begin{align*}
&\frac{\partial}{\partial \gamma_{j,k}}\log \pi(\Gamma_{S_k,k}| \pmb{\lambda}_k,\beta_k,S_k,\mathbf{b}_{S_k,k},\mathcal{D}^{(n)},\eta) \\
&= \left(\frac{\partial}{\partial \gamma_{j,k}}\sum_{i=1}^{n} \log q_{f(\mathbf{x}_i),\eta}(y_i)\right) + (a_\gamma-1)\frac{1}{\gamma_{j,k}} - \frac{1}{b_\gamma}\\
\end{align*}
From \(f(\mathbf{x}_i)=\lambda_{k,i}+\beta_k\prod_{l\in S_k}\phi(x_{i,l}\vert \{l\},b_{l,k},\gamma_{l,k})\), we have
\begin{align*}
&\frac{\partial}{\partial \gamma_{j,k}}\sum_{i=1}^{n} \log q_{f(\mathbf{x}_i),\eta}(y_i) \\
&= \sum_{i=1}^{n}\left(\frac{\partial \log q_{f(\mathbf{x}_i),\eta}(y_i)}{\partial f(\mathbf{x}_i)}\frac{\partial f(\mathbf{x}_i)}{\partial \gamma_{j,k}}\right)\\
&=\beta_k\sum_{i=1}^n\left( \frac{\partial \log q_{f(\mathbf{x}_i),\eta}(y_i)}{\partial f(\mathbf{x}_i)} \frac{\partial \phi(x_{i,j}\vert\{j\},b_{j,k},\gamma_{j,k})}{\partial \gamma_{j,k}}\prod_{l\neq j,l\in S_k}\phi(x_{i, l}\vert\{l\},b_{l,k},\gamma_{l,k}) \right).
\end{align*}

Here,
\begin{align*}
    &\frac{\partial \phi(x_{i,j}\vert \{j\},b_{j,k},\gamma_{j,k})}{\partial\gamma_{j,k}} \\
    &= -\frac{\int_{\mathcal{X}_j}\frac{u-b_{j,k}}{\gamma_{j,k}^2}\tilde{\sigma}\left(\frac{u-b_{j,k}}{\gamma_{j,k}}\right)\mu_{n,j}(du)}{\tilde{c}_j(b_{j,k},\gamma_{j,k})^2}\sigma\left(\frac{x_{i,j}-b_{j,k}}{\gamma_{j,k}}\right) - (c_j(b_{j,k},\gamma_{j,k})-1)\frac{x_{i,j}-b_{j,k}}{\gamma_{j,k}^2}\tilde{\sigma}\left(\frac{x_{i,j}-b_{j,k}}{\gamma_{j,k}}\right).
\end{align*}

\subsubsection{Calculating the Gradient of the Log-Posterior with respect to \(\beta_k\)}

The gradient of the log posterior for $\beta_{k}$ is given as
\begin{align*}
\nabla_{\beta_k} \log\pi(\beta_k\vert \bm{\lambda}_k,S_k,\mathbf{b}_{S_k,k}, \Gamma_{S_k,k}, \mathcal{D}^{(n)},\eta) = \sum_{i=1}^n \frac{\partial \log q_{f(\mathbf{x}_i),\eta}(y_i)}{\partial f(\mathbf{x}_i)}\phi(\mathbf{x}_i\vert \Theta_k) -\frac{\beta_k}{\sigma_\beta^2}.
\end{align*}

\subsection{Sampling Nuisance parameter $\eta$}
\label{sec:eta_smaple}

We only consider the nuisance parameter in the gaussian regression model:
\begin{align*}
    Y_{i}\vert \mathbf{x}_{i}\sim N(\cdot\vert f(\mathbf{x}_{i}),\sigma_g^2)
\end{align*}
for $i=1,...,n$, where $\sigma^2$ is a nuisance parameter.
When the prior distribution is an inverse gamma distribution
\begin{align}
    \sigma_g^2\sim\operatorname{IG}\left(\dfrac{v}{2},\dfrac{v\lambda}{2}\right),
\end{align}
we have
\begin{align}
    \sigma_g^2\vert K, \mathcal{B}_K, \mathbf{S}_K, \mathbf{b}_{S_K,K}, \Gamma_{S_K,K},\mathcal{D}^{(n)} \sim \operatorname{IG}\left(\dfrac{v}{2},\dfrac{ {1\over n} \sum_{i=1}^{n} (y_{i}-f(\mathbf{x}_{i}))^{2}+v\lambda}{2}\right),
\end{align}
and thus $\sigma_g^2$ can be sampled from the conditional posterior easily.

% Since the general strategy for sampling nuisance parameters is analogous across generalized regression models, we restrict our detailed discussion to the gaussian regression model.

% Let $\Theta_{k} = (\beta_{k},S_{k},\mathbf{b}_{S_{k},k},\Gamma_{S_{k},k})$ for $k=1,...,K$. 
% Since
% \begin{align}
%     \sigma^2\sim\operatorname{IG}\left(\dfrac{v}{2},\dfrac{v\lambda}{2}\right),
% \end{align}
% using conjugate property, we have
% \begin{align}
%     \sigma^2 | \Theta_{tpnn},\mathbf{z},\mathcal{D}^{(n)} \sim \operatorname{IG}\left(\dfrac{v}{2},\dfrac{\sum_{i=1}^{n}(y_{i}-\sum_{k=1}^{K}\beta_{k}\phi(\mathbf{x}_{i}|\Theta_k))^{2}+v\lambda}{2}\right).
% \end{align}

\newpage

\section{Details of the experiments}
\label{sec:all_details_exper}

\subsection{Data description}

\begin{table}[h]
\centering
\scriptsize
\caption{\footnotesize \textbf{Descriptions of real datasets.}}
\vskip -0.3cm
\begin{tabular}{c c c c}
\toprule
Dataset & $n$ & $p$ & Task \\ \midrule
\textsc{Abalone} & 4,178 & 8 & Regression \\ 
\textsc{Boston} & 506 & 13 & Regression \\
\textsc{Mpg} & 398 & 7 & Regression \\ 
\textsc{Servo} & 167 & 4 & Regression \\ \midrule
\textsc{Fico} & 10,459 & 23 & Classification \\ 
\textsc{Breast} & 569 & 30 & Classification \\ 
\textsc{Churn} & 7,043 & 20 & Classification \\ 
\textsc{Madelon} & 4,400 & 500 & Classification\\  \midrule
\textsc{CelebA-HQ} & 30,000 & --- & Classification  \\
\textsc{CatDog} & 24,998 & --- &  Classification \\
\bottomrule
\end{tabular}
\end{table}

\subsection{Feature descriptions for \textsc{Madelon} and \textsc{Servo} datasets}
\label{sec:feature_descript_with_index}

%\begin{table}[h]
%\centering
%\scriptsize
%\caption{\footnotesize \textbf{Feature index and its corresponding description for \textsc{Boston} dataset.}}
%\label{table:bostn_description}
%\vskip -0.3cm
%\begin{tabular}{c c}
%\toprule
%Feature index & Feature description \\ \midrule
%1 & Per capita crime rate by town \\ 
%2 & Proportion of residential land zoned for lots over 25,000 sq.ft \\ 
%3 & Proportion of non-retail business acres per town \\ 
%4 & Charles River dummy variable (1 if tract bounds river; 0 otherwise) \\ 
%5 & Nitric oxides concentration (parts per 10 million)\\ 
%6 & Average number of rooms per dwelling \\ 
%7 & Proportion of owner-occupied units built prior to 1940\\ 
%8 & Weighted distances to five Boston employment centres \\ 
%9 & Index of accessibility to radial highways \\ 
%10 & Full-value property-tax rate per $10,000$ \\ 
%11 & Pupil-teacher ratio by town \\ 
%12 & 1000$(\text{Bk} - 0.63)^2$ where Bk is the proportion of blacks by town \\ 
%13 & \% lower status of the population \\ \bottomrule
%\end{tabular}
%\end{table}

\begin{table}[h]
\centering
\scriptsize
\caption{\footnotesize \textbf{Feature index and its corresponding description for \textsc{Servo} dataset.}}
\label{table:servo_description}
\vskip -0.3cm
\begin{tabular}{cc}
\toprule
Feature index & Feature description \\ \midrule
1 & Proportional gain setting for the servo motor.\\ 
2 &  Velocity gain setting for the servo motor. \\ 
3 & Presence of Motor type A \\
4 & Presence of Motor type B \\ 
5 & Presence of Motor type C \\ 
6 & Presence of Motor type D \\ 
7 & Presence of Motor type E \\ 
8 & Presence of Screw type A \\ 
9 & Presence of Screw type B \\ 
10 & Presence of Screw type C \\ 
11 & Presence of Screw type D \\ 
12 & Presence of Screw type E \\ \bottomrule
\end{tabular}
\end{table}

Table \ref{table:servo_description} presents the feature descriptions for \textsc{Servo} dataset \citep{ servo_87}.
\textsc{Madelon} \citep{madelon_171}, introduced in the NIPS 2003 feature selection challenge, is a synthetic binary classification dataset with 500 features, only a few of which are informative while many are redundant or irrelevant.

\subsection{Experiment details for tabular dataset}
\label{sec:exp_details_tabular}

\paragraph{Data Preprocessing.}
All of the categorical input variables are encoded using the one-hot encoding.
For continuous ones, the inverse of the empirical marginal CDF is used to transform them to their marginal ranks
for Bayesian-TPNN and ANOVA-TPNN, whereas they are transformed via the mean-variance standardization
for other baseline models.

\paragraph{Implementation of baseline models.}
For implementation of baseline models,
we proceed as follows.
\begin{itemize}
    \item ANOVA-TPNN : we use the official source code provided in \texttt{https://github.com/ParkSeokhun/ANOVA-TPNN}.
    \item NA$^{1}$M : we use the official source code provided in \texttt{https://github.com/AmrMKayid/nam} and NA$^{2}$M is implemented by extending the code of NA$^{1}$M. 
    \item Linear : We use `scikit-learn' python package \citep{scikit-learn}.
    \item XGB : We use `xgboost' python package \citep{chen2016xgboost}.
    \item BART : We use `BayesTree' R package \citep{chipman2010bart}.
    % \item GBART : We use official code in \cite{linero2025generalized}.
    \item mBNN : We use official code at \texttt{https://github.com/ggong369/mBNN}.
\end{itemize}

\paragraph{Hyperparameters.}

For each model, we perform 5-fold cross validation over the following hyperparameter candidates to select the best configuration.

\begin{itemize}
    \item Bayesian-TPNN
    \begin{itemize}
        \item We set the step size in Langevin proposal as 0.01 and $q_{\text{add}} = 0.28$,  $q_{\text{delete}} = 0.28$ and $q_{\text{change}} = 0.44$ as in \cite{kapelner2016bartmachine}.
        \item We fix $\alpha_{\text{adding}} = 0.95$ and $\gamma_{\text{adding}} = 2$, as in \citet{chipman2010bart}.
        \item \(C_0 \in \{0.001, 0.005, 0.01\}\)
        \item \(a_{\gamma} \in \{1, 2, 4\}\)
        \item \(b_{\gamma} \in \{10^{-3}, 5\cdot10^{-3}, 10^{-2}\}\)
        \item $K_{\max} \in \{100, 200, 300\}$
        \item \(\sigma_\beta^2 \in \{10^{-4}, 10^{-3}, 10^{-2}\}\)
        \item \(M \in \{1, 5\}\)
        \item As in \cite{chipman2010bart}, for $\lambda$, we reparameterize it as $q_{\lambda}$, where $q_{\lambda} = \pi(\sigma^{2} \leq \hat{\sigma}_{\text{OLS}}^{2})$ and $\hat{\sigma}_{\text{OLS}}^{2}$ denotes the residual variance from estimated Linear model.
        The candidate values for $q_{\lambda}$ are $\{0.90, 0.95, 0.99\}$.
        \item We set MCMC iterations as 1000 after 1000 burn-in iterations.
    \end{itemize}
    \item ANOVA-TPNN
    \begin{itemize}
        \item We set the hyperparameter candidates to be the same as those used in \cite{park2025tensor}.
        \item $K_{S} \in \{10,30,50,100\}$
        \item Adam optimizer with learning rate 5e-3.
        \item Batch size = 4,096
        \item Maximum order of component $\in \{1,2\}$
        \item Epoch $\in \{500, 1000, 2000\}$
    \end{itemize}
    \item NAM
    \begin{itemize}
        \item We set the architecture of the deep neural networks to three hidden layers with 64, 64, and 32 units, following \citet{agarwal2021neural}.
        \item Adam optimizer with learning rate 5e-3 and weight decay 7.483e-9.
        \item Batch size = 4,096
        \item Maximum order of component $\in \{1,2\}$
        \item Epoch $\in \{500, 1000, 2000\}$
    \end{itemize}
    \item BART
    \begin{itemize}
        \item We set the hyperparmeter candidates similar to those in \citet{chipman2010bart}.
        \item Number of trees $T \in \{50,100, 200\}$
        \item $\alpha = 0.95$ and $\beta = 2$
        \item $v \in \{1,3,5\}$
        \item $q_{\lambda} \in \{0.90, 0.95, 0.99\}$
        \item For $\sigma_{\mu}=3/(k\sqrt{T})$, $k \in \{1,2,3,5\}$.
        \item  We set MCMC iterations as 1000 after 1000 burn-in iterations.
    \end{itemize}
    % \item GBART
    % \begin{itemize}
    %     \item Number of trees $T \in \{100, 200\}$
    %     \item \(\sigma_\mu^2 \in \{10^{-4}, 10^{-3}, 10^{-2}\}\)
    %     \item We set burn in iteration as 1000 and sampling iteration as 1000.
    % \end{itemize}
    \item XGB
    \begin{itemize}
        \item We consider the hyperparameter candidates used in \cite{park2025tensor}.
    \end{itemize}
    \item mBNN
    \begin{itemize}
        \item We consider the hyperparameter candidates similarly to \cite{kong2023masked}.
        \item Architecture $\in$ \{ 2 hidden layers with 500 and 500 units, 2 hidden layers with 1000 and 1000 units \}
        \item Sparsity hyperparameter $\lambda \in \{0.01, 0.1, 0.5\}$ 
        \item  We set MCMC iterations as 1000 after 1000 burn-in iterations.
    \end{itemize}
\end{itemize}

\paragraph{Computational environments.} 
In this paper, all experiments are conducted on a machine equipped with an NVIDIA RTX 4000 GPU (24GB memory), an Intel(R) Xeon(R) Silver 4310 CPU @ 2.10GHz, and 128GB RAM.

\subsection{Experiment details for image dataset}
\label{sec:details_image}

\paragraph{CNN model.} For CNN that predicts concepts, we attach a linear head for each concept on top of the pretrained ResNet18, and train both the ResNet-18 and the linear heads jointly.

\paragraph{Concept generating.}

Following \cite{oikarinen2023label}, we use GPT-5 \citep{openai2025gpt5} to generate concept dictionaries for \textsc{CelebA-HQ} and \textsc{CatDog} dataset.
Specifically, we prompted GPT-5 as follows:
\begin{itemize}
    \item CelebAMask-HQ is a large-scale face image dataset containing 30,000 high-resolution face images selected from CelebA, following CelebA-HQ.     
    In this context, we aim to classify gender using the CelebAMask-HQ dataset. Could you list five high-level binary concepts that you consider most important for gender classification?
    \item When classifying images of cats and dogs, what are the five most important concepts to consider?
\end{itemize}
Through GPT-5, we obtained a concept dictionary 
$$
\text{\{`Facial hair', `Makeup', `Long hair', `Angular contour', `Accessories'\}}$$
for dataset \textsc{CelebA-HQ} and another dictionary 
$$
\text{\{`Pointed ear', `Short snout', `Almond eye', `Slender/flexible body', `Fine/uniform fur'\}}$$
for dataset \textsc{CatDog}.
Each concept $c$ is divided into a positive part $c_{+}$ and a negative part $c_{-}$.
For example, concept `Makeup' can be divided into `Makeup' and `No Makeup', and `Slender/flexible body' can be divided into `Slender/flexible body' and `Bulky/varied body'.
In turn, we use the pretrained CLIP encoder to convert $c_{+}$ and $c_{-}$ as well as each image into embedding vectors.
For each concept, each image is then assigned a binary  label by measuring which 
of the embeddings of $c_{+}$ and $c_{-}$ the image embedding is closer to.

\paragraph{Hyperparameters.}
For ANOVA-T$^{2}$PNN and NA$^{2}$M are trained using the Adam optimizer with a learning rate of 1e-3 and batch size of 512.
For ANOVA-T$^{2}$PNN, the numbers of basis $K_{S}$ are all equal to $K$
and $K$ is determined using grid search on \{10, 50, 100\}.
For the neural network in NA$^{2}$M, we set hidden layer with sizes (64,64,32).
We implement Linear model as the linear logistic regression using the `scikit-learn' package \citep{scikit-learn}.

\subsection{Experiment details for component selection}
\label{sec:details_component_selec}

\begin{table}[h]
\centering
\scriptsize
\caption{\textbf{Definitions of $f^{(1)}$, $f^{(2)}$ and $f^{(3)}$.}}
\label{table:def_syn}
\vskip -0.3cm
\begin{tabular}{l c}
\toprule
Function & Equation \\
\midrule
$f^{(1)}(\mathbf{x})$ & 
$\pi^{x_1x_2}\sqrt{2|x_3|}-\sin^{-1}(0.5x_4)+\log(|x_3+x_5|+1)
+\dfrac{x_9}{1+|x_{10}|}\sqrt{\dfrac{x_7}{1+|x_8|}}-x_2x_7$ \\[1.2ex]
$f^{(2)}(\mathbf{x})$ & 
$x_1x_2+2^{x_3+x_5+x_6}+2^{x_3+x_4+x_5+x_7}
+\sin(x_7\sin(x_8+x_9))+\arccos(0.9x_{10})$ \\[1.2ex]
$f^{(3)}(\mathbf{x})$ & 
$\tanh(x_1x_2+x_3x_4)\sqrt{|x_5|}+\exp(x_5+x_6)
+\log((x_6x_7x_8)^2+1)+x_9x_{10}+\dfrac{1}{1+|x_{10}|}$ \\
\bottomrule
\end{tabular}
\end{table}

\begin{table}[h]
\centering
\scriptsize
\caption{\textbf{Distributions of input features for each synthetic function.}}
\label{table:dit_syn}
\vskip -0.3cm
\begin{tabular}{l c}
\toprule
Function & Distribution \\
\midrule
$f^{(1)}(\mathbf{x})$ & $X_{1},X_{2},X_{3},X_{6},X_{7},X_{9} \sim^{iid}$ \text{U(0,1)}, $X_{4},X_{5},X_{8},X_{10} \sim^{iid} \text{U(0.6,1)}$ and $X_{11},...,X_{50} \sim^{iid} \text{U(-1,1)}$ \\[1.2ex]
$f^{(2)}(\mathbf{x})$ &  $X_{1},....,X_{50} \sim ^{iid} \text{U(-1,1)}$ \\
$f^{(3)}(\mathbf{x})$ &  $X_{1},....,X_{50} \sim ^{iid} \text{U(-1,1)}$ \\
\bottomrule
\end{tabular}
\end{table}

We generate synthetic datasets from the regression model defined as
\begin{align*}
Y = f^{(k)}(\mathbf{x}) + \epsilon,
\end{align*}
where $\epsilon \sim N(0,\sigma^{2}_{\epsilon})$ and $\mathbf{x}\in\mathbb{R}^{50}$. 
Here, $f^{(k)},k=1,2,3$ are true prediction model used in \cite{tsang2017detecting} and defined in Table \ref{table:def_syn} and the input variables are generated from the distributions in Table \ref{table:dit_syn}.
Input variables indexed 1–10 are informative, as they affect the output, whereas input variables 11–50 are non-informative.
We choose $\sigma^{2}_{\epsilon}$ such that the signal-to-noise ratio is 5.

To evaluate the ability to detect signal components, we conduct experiments in the same manner as in \cite{park2025tensor}.
That is, we use AUROC based on the pairs of $\Vert\hat{f}^{(k)}_{S}\Vert_{2,n}$ and $r_{S}^{(k)}$, computed for all subsets $S \subseteq [p]$ with $|S| = 1, 2, 3$, where $\hat{f}^{(k)}_{S}$ denotes the estimate of $f_{S}^{(k)}$ in $f^{(k)}$ and $r_{S}^{(k)} = \mathbb{I}(\Vert f_{S}^{(k)} \Vert_{2,n} > 0)$ for $k \in \{1,2,3\}$.

\newpage

\section{Ablation studies}
\label{sec:ablat_exp}

\subsection{The number of basis $K$ for various values $C_{0}$}
\label{sec:K_C0}

To evaluate the effect of $C_{0}$ in (\ref{eq:prior_K}) on the number of bases $K$, we conduct experiments with the maximum number of bases $K_{\max}$ set to 200, and 1000 iterations for both burn-in and MCMC updates.
Also, $a_{\gamma}$ and $b_{\gamma}$ are set to be 0.5 and we use \textsc{Abalone} dataset.
Figure \ref{fig:C_{0}_K} shows that $K$ decreases and RMSE increases as $C_{0}$ increases.
This result demonstrates that the hyperparameter $C_{0}$ 
effectively controls model complexity by regulating the number of bases $K$.
A small value of $C_0$ is recommended since an excessively large $C_{0}$ can be detrimental to predictive performance.

\begin{figure}[h]
    \centering
    \includegraphics[width=0.7\linewidth]{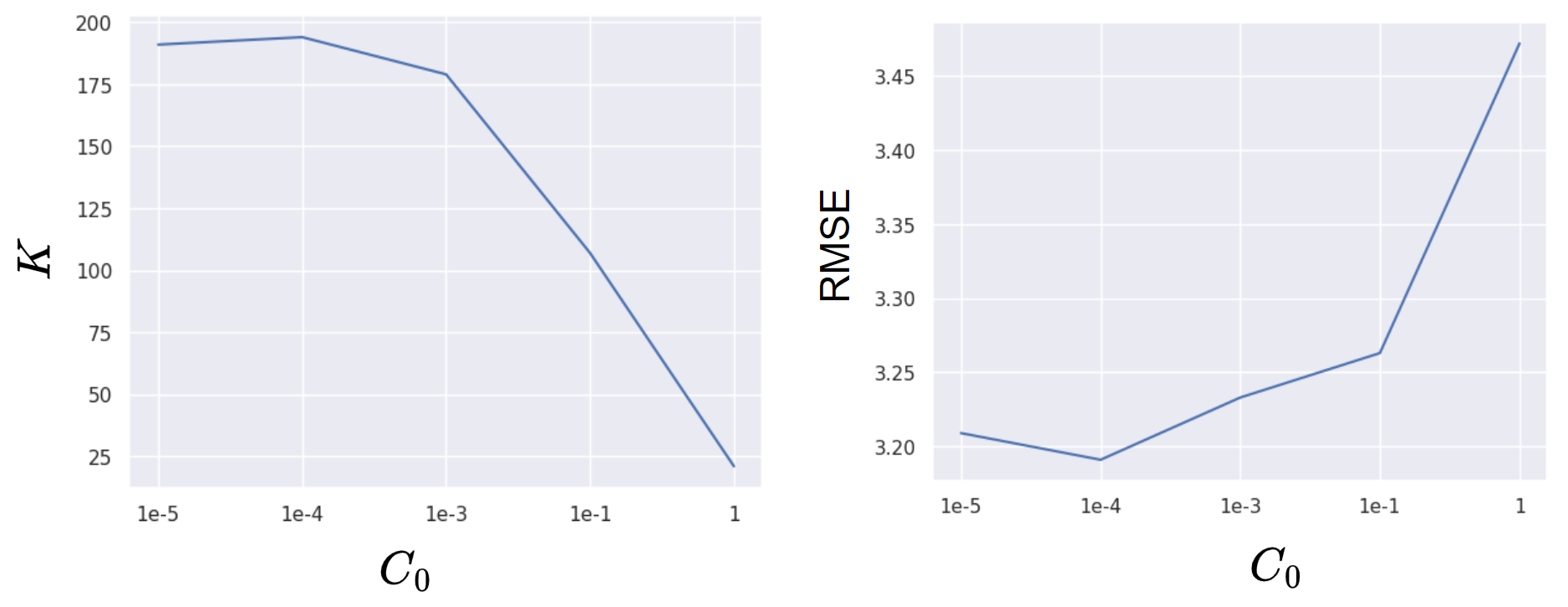}
    \vskip -0.3cm
    \caption{\footnotesize \textbf{Plots of the number of basis $K$ and RMSEs on various $C_{0}$ values.}}
    \label{fig:C_{0}_K}
\end{figure}

\subsection{Impact of the hyperparameters $a_{\gamma}$ and $b_{\gamma}$ on prediction performance}

We conduct an experiment to evaluate the effect of shape parameter $a_{\gamma}$ and scale parameter $b_{\gamma}$ on prediction performance.
Except for $a_{\gamma}$ and $b_{\gamma}$, the other hyperparameters of Bayesian-TPNN are set identical to those in Section \ref{sec:K_C0} of Appendix, and we analyze \textsc{Abalone} dataset.
We observe that prediction performance is relatively insensitive to the choice of the shape parameter $a_{\gamma}$, whereas it is somehow sensitive to the choice of the scale parameter $b_{\gamma}$.
Note that the scale of $\gamma$ controls the smoothness of each TPNN basis $\phi(\mathbf{x}|\Theta)$ and thus
the smoothness of Bayesian-TPNN model. 

\begin{table}[h]
\scriptsize
\centering
\caption{\footnotesize \textbf{Prediction performance depends on various values of $a_{\gamma}$ and $b_{\gamma}$.}}
\vskip -0.3cm
\begin{tabular}{ccccc}
\toprule
$b_{\gamma} \backslash a_{\gamma}$ &   0.5 &   1 & 2 & 3 \\ \midrule
1e-5 & 3.247 & 3.202 & 3.278 & 3.228 \\ \midrule
1e-4 & 3.224 & 3.215 & 3.184 & 3.175 \\ \midrule
0.01 & 3.211 & 3.182 & 3.184 & 3.175 \\ \midrule
0.1 &  3.213 & 3.258 & 3.282 & 3.343 \\ \bottomrule
\end{tabular}%
\end{table}

\subsection{Impact of the step size in the Langevin proposal}

We conduct an experiment to investigate the effect of the step size in the Langevin proposal for $(\mathbf{b}_{S_{k},k},\Gamma_{S_{k},k},\beta_{k})$.
Except for the step size, the other hyperparameters of Bayesian-TPNN are set identical to those in Section \ref{sec:K_C0} of Appendix, and we analyze \textsc{Abalone} dataset.
Table \ref{table:ablation_stepsize} presents the prediction performances of Bayesian-TPNN for various step sizes.
Our results show that overly large step sizes in the Langevin proposal can degrade the prediction performance due to poor acceptance and unstable exploration, whereas a moderate range yields the best performance.
Therefore, a not too large step size is recommended in practice.

\begin{table}[h]
\scriptsize
\centering
\caption{\footnotesize \textbf{Prediction performances of Bayesian-TPNN for various step sizes in the Langevin proposal .}}
\label{table:ablation_stepsize}
\vskip -0.3cm
\begin{tabular}{cccccccccc}
\toprule
Step size & 0.01 & 0.02 & 0.04 & 0.08 & 0.1 & 0.2 & 0.3 & 0.4 & 0.5\\ \midrule
RMSE &3.199 & 3.216 & 3.209 & 3.269 & 3.160 & 3.243 & 4.308 & 4.549 & 4.578\\ \bottomrule
\end{tabular}%
\end{table}

\subsection{Impact of $p_{\text{input}}$ on estimating higher-order components}

\label{sec:p_input_exp}

We conduct an experiment to evaluate the effects of using $p_{\text{input}}$ other than the uniform distribution in the MH algorithm.
We refer to the model with the uniform distribution for $p_{\text{input}}$ as Uniform Bayesian-TPNN, and the model where $p_{\text{input}}$ is determined using the feature importance from a pretrained XGB as Bayesian-TPNN.
Table \ref{table:auroc_uniform_madelon} compares prediction performances of Uniform Bayesian-TPNN (UBayesian-TPNN) and Bayesian-TPNN on \textsc{Madelon} dataset. 
To investigate why the prediction performance improvement occurs when using the nonuniform $p_{\text{input}},$
we identify the 5 most important components for each model whose results are presented in Table \ref{table:high_component_score_pure}.
UBayesian-TPNN only detects two thrid-order interactions as signals and ignores even all of the main effects.
In contrast, Bayesian-TPNN captures the fourth-order component as the most important but is also able to capture other meaningful lower-order components including two main effects effectively. 

We also analyze the synthetic datasets in Section \ref{sec:exp_comp} with UBayesian-TPNN, and the corresponding results are reported in Table \ref{table:weight_comp_selc}.
These results amply imply that $p_{\text{input}}$ plays an important role in detecting higher-order components and leading to substantial improvements in both prediction performance and component selection.

\begin{table}[h]
\caption{\footnotesize \textbf{Prediction performance on \textsc{Madelon} dataset.}}
\label{table:auroc_uniform_madelon}
\scriptsize 
\vskip -0.3cm
\centering
\footnotesize
\begin{tabular}{ccc}
\toprule
Model & UBayesian-TPNN & Bayesian-TPNN \\ \midrule
AUROC $\uparrow$ (SE) & 0.739 (0.002) & \textbf{0.854} (0.007) \\ \bottomrule
\end{tabular}
\end{table}

\begin{table}[h]
\centering
\scriptsize
\caption{\footnotesize \textbf{Top 5 components  with the important scores normalized by the maximum.}}
\label{table:high_component_score_pure}
\vskip -0.3cm
\scalebox{0.9}{
\begin{tabular}{c cc cc cc cc cc}
\toprule
Model & 
\multicolumn{2}{c}{Rank 1} & \multicolumn{2}{c}{Rank 2} & 
\multicolumn{2}{c}{Rank 3} & \multicolumn{2}{c}{Rank 4} & 
\multicolumn{2}{c}{Rank 5} \\
\midrule
 & Comp. & Score & Comp. & Score & Comp. & Score & Comp. & Score & Comp. & Score \\
\midrule
\makecell[c]{UBayesian-TPNN} & (203,289,421) & 1.000 & (30,149,212) & 0.950 & (148,176,298) & 0.006 & (75,232,442) & 0.005 & (64,373,379) & 0.004\\
\makecell[c]{Bayesian-TPNN} & (49,242,319,339) & 1.000 & (129,443,494) & 0.472 & (379,443) & 0.374 & 106 & 0.322 & (242,443) & 0.301 \\
\bottomrule
\end{tabular}
}
\end{table}

\begin{table}[h]
\centering
\scriptsize
\caption{\footnotesize{\textbf{Performance of component selection on the synthetic datasets.}}}
\label{table:weight_comp_selc}
\vskip -0.3cm
\begin{tabular}{cc cc cc c}
    \toprule
     True model & \multicolumn{2}{c}{$f^{(1)}$} & \multicolumn{2}{c}{$f^{(2)}$} & \multicolumn{2}{c}{$f^{(3)}$}  \\ \midrule
     Order & \makecell[c]{UBayesian\\TPNN} & \makecell[c]{Bayesian\\TPNN} & \makecell[c]{UBayesian\\TPNN} & \makecell[c]{Bayesian\\TPNN} & \makecell[c]{UBayesian\\TPNN} & \makecell[c]{Bayesian\\TPNN} \\ \midrule
      1 & \makecell[c]{\textbf{1.000}\\(0.000)} & \makecell[c]{\textbf{1.000}\\(0.000)} &  \makecell[c]{0.826\\(0.024)} & \makecell[c]{\textbf{0.831}\\(0.008)} &  \makecell[c]{0.824\\(0.009)} & \makecell[c]{\textbf{1.000}\\(0.000)} \\ 
      2 & \makecell[c]{0.988\\(0.010)} & \makecell[c]{\textbf{1.000}\\(0.000)}  & \makecell[c]{0.953\\(0.006)} & \makecell[c]{\textbf{0.985}\\(0.003)} & \makecell[c]{0.750\\(0.006)} & \makecell[c]{\textbf{0.922}\\(0.019)} \\ 
      3 & \makecell[c]{0.736\\(0.050)} & \makecell[c]{\textbf{0.740}\\(0.022)} & \makecell[c]{0.878\\(0.020)} & \makecell[c]{\textbf{0.966}\\(0.018)}  & \makecell[c]{0.658\\(0.011)} & \makecell[c]{\textbf{0.661}\\0.022}  \\ \bottomrule
\end{tabular}
\end{table}

% \subsection{Impact of $p_{\text{input}}$ on the convergence speed in MCMC}

% In this section, we investigate the impact of the choice of $p_{\text{input}}$ on the convergence rate of the MCMC algorithm.
% Specifically, we compare two scenarios: (1) a uniform distribution over the candidate variables, and (2) a knowledge-guided distribution, where $p_{\text{input}}$ is constructed using importance scores derived from global SHAP values of a pre-trained DNN.
% Figure \ref{fig:RMSE_trajectory} presents the RMSE trajectories of validation data across MCMC iterations on \textsc{Servo} dataset.
% The results indicate that specifying $p_{\text{input}}$ based on global SHAP values from a pre-trained DNN leads to faster MCMC convergence.

% \begin{figure}[h]
%     \centering
% \includegraphics[width=0.7\linewidth]{figure/fig_weight_uniform_convergence.png}
% \vskip -0.3cm
%     \caption{\footnotesize \textbf{Plots of RMSE trajectories across MCMC iteration.}}
%     \label{fig:RMSE_trajectory}
% \end{figure}

\subsection{Impact of stepwise search in the proposal of $K$}
\label{sec:impoact_K_proposal}

We conduct an experiment to evaluate the effectiveness of \textbf{Stepwise} move in the proposal distribution of $K$ suggested in Section \ref{sec:mcmc}.
We compare the performances of Bayesian-TPNN with and without \textbf{Stepwise} move on \textsc{Madelon} dataset.
Table \ref{table:search_exper} reports the averages and standard errors of AUROCs, ECEs, and NLLs over 5 trials and 
Table \ref{table:high_component_score_search} shows the top 5 important components. 
The results suggest that the \textbf{Stepwise} move is helpful to detect
higher-order interactions which in turn leads to improvements in both prediction performance and uncertainty quantification.

\begin{table}[h]
\centering
\scriptsize
\caption{\footnotesize \textbf{Results of performance with and without \textbf{Stepwise} move.}}
\label{table:search_exper}
\vskip -0.3cm
\begin{tabular}{ccc}
\toprule
 & With \textbf{Stepwise} move & Without \textbf{Stepwise} move \\ \midrule
AUROC $\uparrow$ (SE) & \textbf{0.854} (0.007) &  0.820 (0.002)\\ 
ECE $\downarrow$ (SE) & \textbf{0.076} (0.004)& 0.106 (0.007) \\ 
NLL $\downarrow$ (SE) & \textbf{0.479} (0.009) & 0.650 (0.005) \\ \bottomrule
\end{tabular}
\vskip 0.5cm
\caption{\textbf{Top 5 components with the important scores normalized by the maximum.}}
\label{table:high_component_score_search}
\vskip -0.3cm
\scriptsize
\scalebox{0.9}{
\begin{tabular}{c cc cc cc cc cc}
\toprule
Model & 
\multicolumn{2}{c}{Rank 1} & \multicolumn{2}{c}{Rank 2} & 
\multicolumn{2}{c}{Rank 3} & \multicolumn{2}{c}{Rank 4} & 
\multicolumn{2}{c}{Rank 5} \\
\midrule
 & Comp. & Score & Comp. & Score & Comp. & Score & Comp. & Score & Comp. & Score \\
\midrule
With \textbf{Stepwise} move & (49,242,319,339) & 1.000 & (129,443,494) & 0.472 & (379,443) & 0.374 & 106 & 0.322 & (242,443) & 0.301 \\
Without \textbf{Stepwise} move  & (129,242) & 1.000 & (29,339,379) & 0.986 & 339 & 0.622 & 337 & 0.544 & (242,443) & 0.526 \\
\bottomrule
\end{tabular}
}
\end{table}

\newpage

\section{Experiment for the Poisson regression}

\begin{figure}[h]
    \centering
    \includegraphics[width=0.9\linewidth]{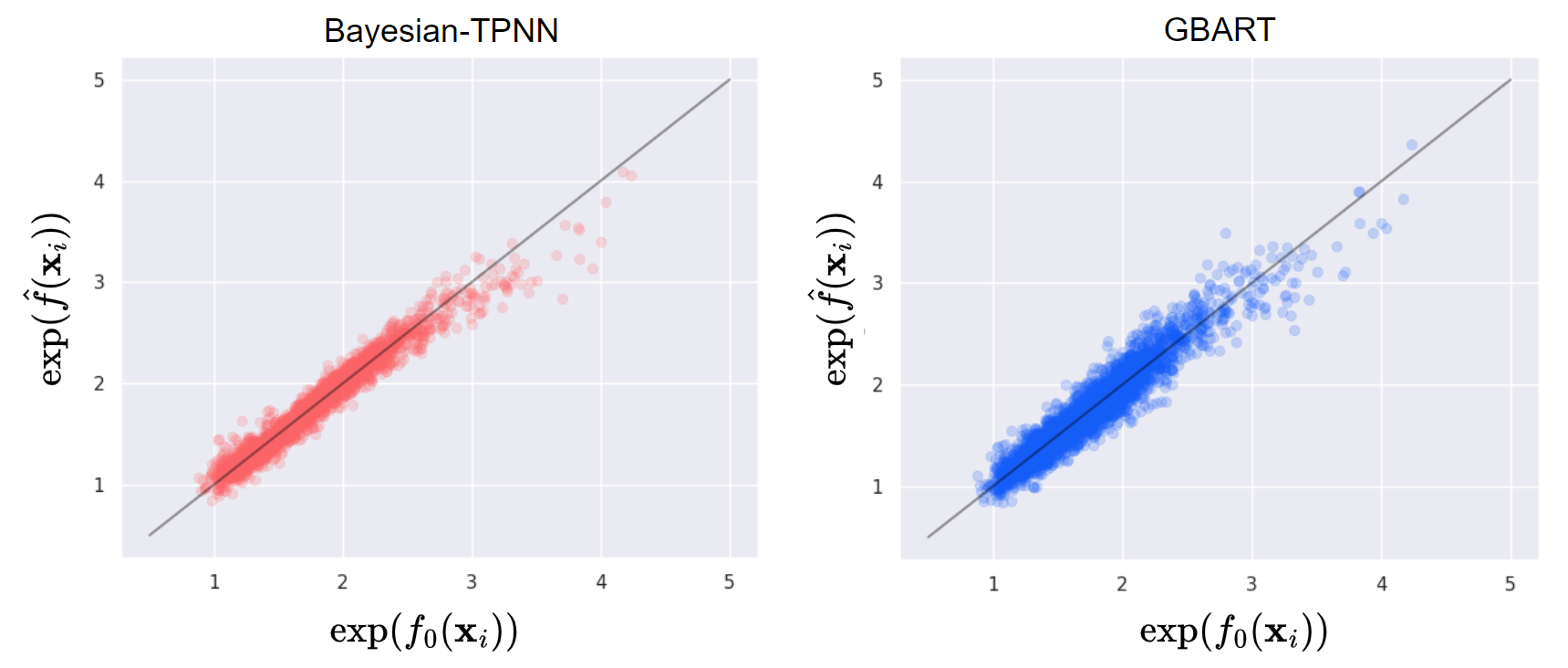}
    \caption{\textbf{Scatter Plots between the true expectations and estimated ones.}}
    \label{fig:poisson-pred}
\end{figure}

In this section, we compare the prediction performance and uncertainty quantification of Bayesian-TPNN with GBART \citep{linero2025generalized} on the Poisson regression model.
We consider the poisson regression model defined as
$$ Y_i\vert\mathbf{x}_i\sim \operatorname{Poisson}(\exp(f(\mathbf{x}_{i}))),$$
where $f$ is the regression function.
We generate a synthetic dataset of size 15,000 using the true regression function $f_{0}$ defined as
$$f_{0}(\mathbf{x})=\pi^{x_1x_2}\sqrt{2|x_3|}-\sin^{-1}(0.5x_4)+\log(|x_3+x_5|+1)
+\dfrac{x_9}{1+|x_{10}|}\sqrt{\dfrac{x_7}{1+|x_8|}}-x_2x_7,$$
where input variable $\mathbf{x}_{i} \in \mathbb{R}^{10}$ are generated from $\text{Uniform}(0,1)^{10}$ for $i=1,...,15,000$.
%We assume that $f(\mathbf{x})$ is modeled either by Bayesian-TPNN or GBART, and estimate both models.
Table \ref{tab:pois_perf} presents the RMSE  and NLL for Bayesian-TPNN and GBART, demonstrating that Bayesian-TPNN achieves superior performance to GBART even in the Poisson regression.
Here, the RMSE is calculated between $\exp(f_{0}(\mathbf{x}_{i}))$ and $\exp(\hat{f}(\mathbf{x}_{i}))$ for $i=1,..,15,000$, where $\hat{f}$ is the Bayes estimate.
Figure \ref{fig:poisson-pred} shows the scatter plot of predicted values $\exp(\hat{f}(\mathbf{x}_{i}))$ versus $\exp(f_{0}(\mathbf{x}_{i}))$ for $i=1,...,15,000.$
It implies that Bayesian-TPNN yields predictions much closer to the true values compared to GBART.

% In the possion regression model, the response \(Y\) follows a Poisson distribution, commonly used for count data. Specifically, the conditional distribution is
% $$Y\vert \mathbf{x}\sim\operatorname{Poisson}(\lambda(\mathbf{x}))$$
% and the goal is to estimate the conditional expectation \(\lambda(\mathbf{x})\). However, since \(\lambda(\mathbf{x})\) must be positive, many approaches instead model a real-valued function \(f(\mathbf{x})\) and set
% $$ Y\vert\mathbf{x}\sim \operatorname{Poisson}(\exp(f(\mathbf{x}))).$$

% For our experiment, we generate a synthetic dataset according to
% $$ Y_i\vert\mathbf{x}_i\sim \operatorname{Poisson}(\exp(f_{0}(\mathbf{x}))) $$
% where
% $$f_{0}(\mathbf{x})=\pi^{x_1x_2}\sqrt{2|x_3|}-\sin^{-1}(0.5x_4)+\log(|x_3+x_5|+1)
% +\dfrac{x_9}{1+|x_{10}|}\sqrt{\dfrac{x_7}{1+|x_8|}}-x_2x_7$$

% Each explanatory vector \(\mathbf{x}_i\in(0,1)^{10}\) is drawn from uniform distribution, and the conditional mean \(\exp(f(\mathbf{x}_i))\) is then computed. The response \(Y_i\) is sampled from the Poisson distribution identified by this value.

% We generate 15,000 observations using this procedure and split the dataset into training and test sets with an 80/20 ratio for model fitting and evaluation.

% The results are given in Table \ref{tab:pois_perf}.

\begin{table}[h]
    \caption{\footnotesize \textbf{Prediction performance and uncertainty quantification on Poisson synthetic dataset.}}
    \centering
    \begin{footnotesize}
    \begin{tabular}{ccc}
        \toprule
         & Bayesian-TPNN & GBART \\ \midrule
         RMSE \(\downarrow\) & \textbf{0.094} & 0.141 \\ 
         NLL \(\downarrow\) & \textbf{1.615} & 1.629\\  \bottomrule
%         CRPS \(\downarrow\) & \textbf{0.053} & 0.080 \\ \hline
    \end{tabular}
    \end{footnotesize}
    \label{tab:pois_perf}
\end{table}

\newpage

\section{Experiments for interpretability}
\label{sec:interpretable_more}

\subsection{Interpretability on the Image datasets}

In this section, we describe the local and global interpretations of CBM \citep{koh2020concept} with Bayesian-TPNN on  \textsc{CelebA-HQ} and \textsc{CatDog} datasets. 
Table \ref{table:desc_img} presents the description of concepts used in \textsc{CelebA-HQ} and \textsc{CatDog} datasets.

\begin{table}[h]
\centering
\footnotesize
\caption{\footnotesize \textbf{Description of image datasets.}}
\label{table:desc_img}
\vskip -0.3cm
\begin{tabular}{ccc}
\toprule
Index & \textsc{CelebA-HQ} & \textsc{CatDog} \\ \midrule 
1 & Facial hair & Pointed ear    \\ 
2 & Makeup & Short snout  \\ 
3 & Long hair & Almond eye    \\ 
4 & Angular contour & Slender/flexible body  \\ 
5 & Accessories & Fine/uniform fur  \\ \bottomrule
\end{tabular}
\vskip 0.5cm
\caption{\footnotesize \textbf{Normalized importance scores and ranks for the top 5 important components on the image datasets.}}
\label{table:top5_comp_img}
\vskip -0.3cm
\centering
\scriptsize
\caption*{\footnotesize \textsc{CelebA-HQ}}
\vskip -0.3cm
\begin{tabular}{ccccccc}
\toprule
 & Rank & 1 & 2 & 3 & 4 & 5 \\ \midrule
\makecell{Bayesian-TPNN} & \makecell{Component index \\ Score} & \makecell{2 \\ 1.000} & \makecell{4 \\ 0.665} & \makecell{(2,3) \\ 0.592} & \makecell{(2,4) \\ 0.304} & \makecell{(1,5) \\ 0.262}  \\ \midrule
\makecell{ANOVA-T$^{2}$PNN} & \makecell{Component index\\ Score} & \makecell{(2,3) \\ 1.000} & \makecell{1 \\ 0.482} & \makecell{(1,5) \\ 0.284} & \makecell{4 \\ 0.262} & \makecell{5 \\ 0.211}  \\ \midrule
\makecell{Linear} & \makecell{Component index\\ Score} & \makecell{2 \\ 1.000} & \makecell{1 \\ 0.783} & \makecell{4 \\ 0.549} & \makecell{5 \\ 0.328} & \makecell{3 \\ 0.304} \\ \bottomrule
\end{tabular}%

\vskip 0.5cm
\centering
\caption*{\footnotesize \textsc{CatDog}}
\scriptsize
\vskip -0.3cm
\begin{tabular}{ccccccc}
\toprule
 & Rank & 1 & 2 & 3 & 4 & 5 \\ \midrule
\makecell{Bayesian-TPNN} & \makecell{Component index \\ Score} & \makecell{3 \\ 1.000} & \makecell{(3,4) \\ 0.395} & \makecell{2 \\ 0.252} & \makecell{4 \\ 0.162} & \makecell{(2,3,4,5) \\ 0.086}  \\ \midrule
\makecell{ANOVA-T$^{2}$PNN} & \makecell{Component index\\ Score} & \makecell{(4,5) \\ 1.000} & \makecell{3 \\ 0.883} & \makecell{(3,5) \\ 0.882} & \makecell{4 \\ 0.716} & \makecell{(1,4) \\ 0.453}  \\ \midrule
\makecell{Linear} & \makecell{Component index\\ Score} & \makecell{5 \\ 1.000} & \makecell{1 \\ 0.698} & \makecell{3 \\ 0.352} & \makecell{2 \\ 0.023} & \makecell{4 \\ 0.021} \\ \bottomrule
\end{tabular}%
\end{table}

\paragraph{Global interpretation.}

Table \ref{table:top5_comp_img} shows the top 5 most important components along with their importance scores
(normalized by the maximum score) for Bayesian-TPNN, ANOVA-T$^{2}$PNN and Linear model.
In \textsc{CatDog} dataset, Bayesian-TPNN identifies the 4th-order component (2,3,4,5) as an important component.
It seems that complex interactions exists between the 5 concepts.

\begin{figure}[h]
    \centering
    \includegraphics[width=0.7\linewidth]{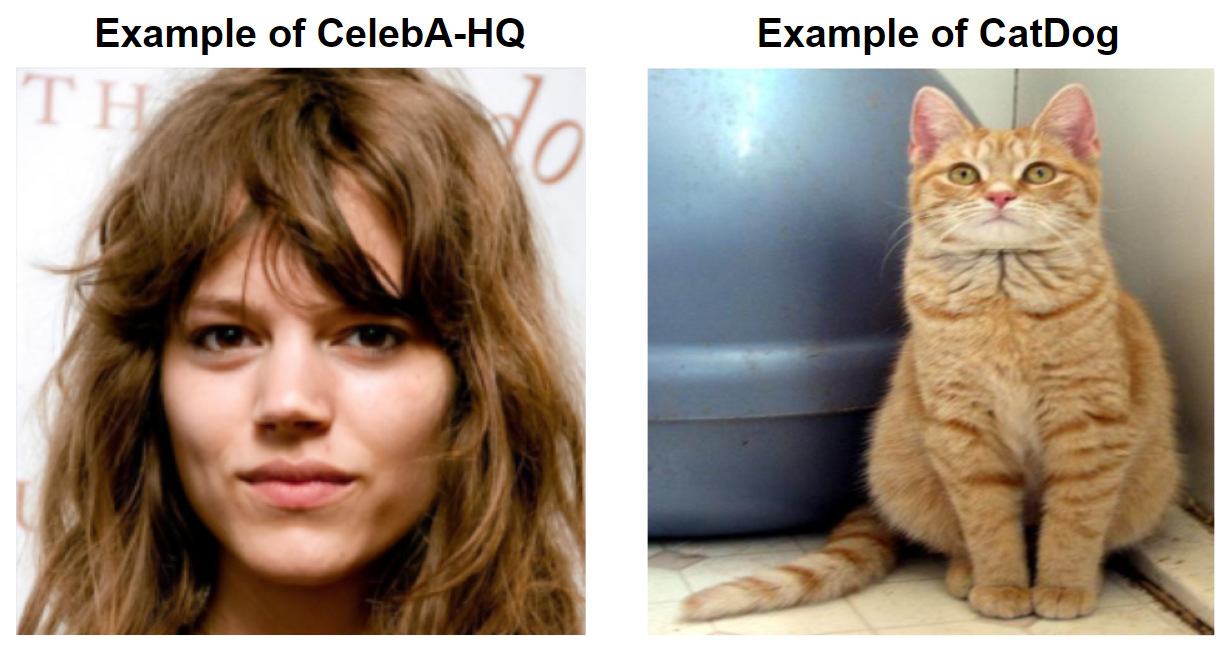}
    \caption{\footnotesize \textbf{Examples of images misclassified by Linear model.}}
    \label{fig:image_local_example}
\end{figure}

\paragraph{Local interpretation.}

Figure \ref{fig:image_local_example} presents two images where Bayesian-TPNN correctly classifies
but Linea model does not.
For the \textsc{CelebA-HQ} example image, Linear model incorrectly predicts it as male, whereas the Bayesian-TPNN correctly predicts as female.
The contributions of the important components for this image are presented in Table \ref{table:local_celeba}.
In Linear model, `Makeup' gives a positive contribution, which leads to a misclassification of the image as male.
In contrast, in Bayesian-TPNN, while the main effect of `Makeup' still provides a positive contribution, the interactions between (`Makeup', `Long hair') and (`Makeup', `Angular contour') yield negative contributions, resulting in a correct prediction as female.

For the \textsc{CatDog} example image, Linear model incorrectly predicts it as `dog', whereas Bayesian-TPNN correctly predicts as `cat'.
Table \ref{table:local_catdog} indicates that Linear model misclassifis the image as `dog' due to the positive contribution of `Almond eye'.
In contrast, although Bayesian-TPNN also assigns a positive contribution to `Almond eye', the higher-order interactions—(`Almond eye', `Slender/flexible body') and (`Short snout', `Almond eye', `Slender/flexible body', `Fine/uniform fur')—provided much stronger negative contributions, leading to the correct classification as a cat.

These two examples strongly suggest that considering higher-order interactions between concepts is necessary for the success
of CBM.

\begin{table}[h]
\centering
\footnotesize
\caption{\footnotesize \textbf{Prediction values of the 5 most important components for \textsc{CelebA-HQ} image.}}
\label{table:local_celeba}
\vskip -0.3cm
\begin{tabular}{ccccccc}
\toprule
\multirow{2}{*}{Bayesian-TPNN} & Component index & 2 & 4 & (2,3) & (2,4) & (1,5) \\  
                  & Contribution & 0.297 & 0.184 & -0.444 & -0.323 & -0.207 \\ \midrule
\multirow{2}{*}{Linear} & Component index & 1 & 2 & 3 & 4 & 5 \\ 
                  & Contribution & -0.222 & 3.746 & -1.510 & -2.665 & 1.627 \\ \bottomrule
\end{tabular}
\vskip 0.5cm
\caption{\footnotesize \textbf{Prediction values of the 5 most important components for \textsc{CatDog} image.}}
\label{table:local_catdog}
\vskip -0.3cm
\begin{tabular}{ccccccc}
\toprule
\multirow{2}{*}{Bayesian-TPNN} & Component & 3 & (3,4) & 2 & 4 & (2,3,4,5) \\ 
                  & Contribution & 0.618 & -0.767 & 0.181 & -0.778 & -0.355 \\ \midrule
\multirow{2}{*}{Linear} & Component & 1 & 2 & 3 & 4 & 5 \\ 
                  & Contribution & -4.304 & -0.630 & 9.503 & -2.463 & -4.113 \\ \bottomrule
\end{tabular}
\end{table}

\paragraph{Fewer concepts, better prediction performance.}

\begin{table}[h]
\centering
\scriptsize
\caption{\footnotesize \textbf{Prediction performance on the image datasets.}}
\label{table:number_of_concept_image}
\vskip -0.3cm
\begin{tabular}{ccc}
\toprule
 & Bayesian-TPNN with 5 concepts & Linear with 10 concepts \\ \midrule
\textsc{CelebA-HQ} & \textbf{0.936} (0.002) & 0.899 (0.001) \\ 
\textsc{CatDog} & \textbf{0.878} (0.002) & 0.869 (0.002) \\ \bottomrule
\end{tabular}%
\end{table}
 
One may argue that 5 concepts are too small for Linear model and Linear model would perform well with more concepts.
To see the validity of this argument, we compare predictive performance of Bayesian-TPNN with 5 concept and 
Linear model with 10 concepts, where
additional 5 concepts are generated through GPT-5: for \textsc{CelebA-HQ} dataset,
\begin{align*}
&\{\text{`Emphasized eyes', `Prominent lips', `Smooth skin'},\\ &\quad \text{`Pronounced cheekbones', `High contrast'}\}
\end{align*}
and for \textsc{CatDog} dataset,
\begin{align*}
&\{\text{`Long tail', `Retractable claws (hidden)', `Upright sitting or crouching posture'},\\ &\quad \text{`Small mouth / Meowing shape', `Ambush-like pose (crouched)'}\}.
\end{align*}

Table \ref{table:number_of_concept_image} presents the averages and standrad errors of AUROCs for Bayesian-TPNN with 5 concepts and Linear model with 10 concepts.
While using more concepts with Linear model improves prediction accuracy, 
 Bayesian-TPNN is still superior to Linear model even though fewer concepts
are used in Bayesian-TPNN.
This implies that capturing higher-order interactions plays a more critical role in improving prediction performance than merely increasing the number of concepts. 
Quality of concepts generated by GPT would be problematic.

\subsection{Additional results of local interpretation on the Tabular dataset}
\label{sec:local_additional}

In this section, we describe the results of local interpretation on \textsc{boston} dataset.
Specifically, we examine the contributions of the 5 most important components identified by Bayesian-TPNN
in Section \ref{sec:interpretation} at a specific input vector $\mathbf{x}.$
For a given data point 
$$
\mathbf{x} = (0.006, 18, 2.31, 0, 0.538, 6.58, 65.2, 4.09, 1, 296, 15.3 , 396.9, 4.98),
$$
the contributions of the 5 estimated components $\hat{f}_{\{13\}}, \hat{f}_{\{6\}}, \hat{f}_{\{1\}}, \hat{f}_{\{8\}}$, and $\hat{f}_{\{1,6\}}$ by Bayesian-TPNN are given as
$$
(\hat{f}_{\{13\}}(\mathbf{x}), \hat{f}_{\{6\}}(\mathbf{x}), \hat{f}_{\{1\}}(\mathbf{x}), \hat{f}_{\{8\}}(\mathbf{x}),\hat{f}_{\{1,6\}}(\mathbf{x})) = (0.575, -0.108, 0.080, -0.002, -0.001).
$$
In particular, the component $\hat{f}_{\{13\}}$ makes a substantial positive contribution to the housing price.
That is, the price of the house for a given input vector $\mathbf{x}$ is high because of the main effect $x_{13}$.

\newpage

\section{Experiment for stability of component estimation}
\label{sec:add_exp_stab}

\cite{park2025tensor} demonstrated, both theoretically and empirically, that TPNN reliably estimates the components of the functional ANOVA model.
In this section, we investigate whether Bayesian-TPNN exhibits the same stability in component estimation.
For this analysis, we randomly split the dataset into training and test datasets.
From this, we obtain estimators for the components. We repeat this procedure five times to obtain five estimators for each component. We then calculate the stability score using these estimators.
Specifically, following \cite{park2025tensor}, for predefined data $\{\mathbf{x}_{1},...,\mathbf{x}_{n}\}$, we use the stability score defined as
\begin{align*}
    \mathcal{SC}(f_S) := \frac{1}{n}\sum_{i=1}^n \frac{\sum_{j=1}^{5}(f_S^j(\mathbf{x}_i)-\bar{f}_S(\mathbf{x}_i))^2}{\sum_{j=1}^5 (f_S^j(\mathbf{x}_i))^2},
\end{align*}
where $f_{S}^{j}$ is the estimated component for $S$ obtained from the $j$-th fold and $\bar{f}_{S}(\mathbf{x}) = \sum_{j=1}^{5}f_{S}^{j}(\mathbf{x})/5$.
Finally, we use $\mathcal{SC}^{d}(f) := {1\over \sum_{k=1}^{d}{p \choose k}}\sum_{S \subseteq [p],|S|\leq d}\mathcal{SC}(f_{S})$ to compare the stability in component estimation between Bayesian-TPNN, ANOVA-TPNN and NAM.

Table \ref{table:stability_score_main} presents the results of stability scores $\mathcal{SC}^{1}(f)$ for Bayesian-TPNN, ANOVA-T$^{1}$PNN and NA$^{1}$M, where
ANOVA-T$^{1}$PNN and NA$^{1}$M estimate only the main effects.
Table \ref{table:st_second_component} shows of stability scores $\mathcal{SC}^{2}(f)$ for Bayesian-TPNN, ANOVA-T$^{2}$PNN and NA$^{2}$M, where
ANOVA-T$^{2}$PNN and NA$^{2}$M estimate up to second-order components.
These results imply that Bayesian-TPNN estimates the components more stably than ANOVA-TPNN and NAM.
Note that for \textsc{Madelon} dataset, which has an input dimension of 500, we could not train ANOVA-T$^{2}$PNN and NA$^{2}$M due to the computational environment, and thus their stability scores could not be calculated.

\begin{table}[h]
    \centering
    \scriptsize
    \caption{\footnotesize \textbf{Stability scores of Bayesian-TPNN, ANOVA-T$^{1}$PNN and NA$^{1}$M.}}
    \vskip -0.3cm
    \label{table:stability_score_main}
    \label{table:}
    \begin{tabular}{c c c c}
        \toprule
        Dataset & \makecell[c]{Bayesian\\TPNN} & \makecell[c]{ANOVA\\T$^{1}$PNN} & NA$^{1}$M \\ \midrule
        \textsc{Abalone} & \textbf{0.087} & 0.405 & 0.555 \\
        \textsc{Boston} & \textbf{0.368} & 0.425 & 0.583  \\
        \textsc{Mpg} & \textbf{0.222} & 0.411 & 0.472 \\
        \textsc{Servo} & \textbf{0.339}  & 0.651 & 0.481 \\ \midrule
        \textsc{Fico} &  \textbf{0.130} & 0.287 & 0.607  \\
        \textsc{Breast} & \textbf{0.100} & 0.286 & 0.569 \\
        \textsc{Churn} & \textbf{0.111} & 0.558 & 0.569 \\ 
        \textsc{Madelon} & \textbf{0.520} & 0.685 & 0.785 \\ \bottomrule
    \end{tabular}
\end{table}

\begin{table}[h]
    \centering
    \scriptsize
    \captionof{table}{\footnotesize{\textbf{Stability scores of Bayeisan-TPNN, ANOVA-T$^{2}$PNN and NA$^{2}$M.}}}
    \label{table:st_second_component}
    \vskip -0.3cm
    \begin{tabular}{c c c c}
        \toprule
        Dataset & \makecell[c]{Bayesian\\TPNN} & \makecell[c]{ANOVA\\T$^{2}$PNN} & NA$^{2}$M \\ \midrule
        \textsc{Abalone} & 0.400 & \textbf{0.340} & 0.770 \\
        \textsc{Boston} & 0.615 & \textbf{0.380} & 0.705 \\
        \textsc{Mpg} & \textbf{0.340} & 0.370 & 0.560 \\
        \textsc{Servo} &  \textbf{0.445} & 0.575 & 0.665\\ \midrule
        \textsc{Fico} &  \textbf{0.525} & 0.540 & 0.790 \\
        \textsc{Breast} & \textbf{0.630} & 0.675 & 0.730 \\
        \textsc{Churn} & \textbf{0.520} & 0.755 & 0.730 \\ 
        \textsc{Madelon} & \textbf{0.475} & --- & --- \\ 
        \bottomrule
    \end{tabular}
    \label{tab:stability}
\end{table}

\newpage

\section{Comparison of convergence speed and runtime in MCMC algorithm}
\label{sec:runtime}

In this section, we evaluate the convergence speed and runtime of MCMC algorithms for Bayesian-TPNN.
Specifically, we compare its convergence speed with that of mBNN, and its runtime with those of ANOVA-T$^{2}$PNN and mBNN.
In Bayesian-TPNN, we set $K_{\max}=100$. For mBNN, we use two hidden layers with 500 units each and set the number of HMC steps to 30.
For ANOVA-T$^{2}$PNN, we set $K_{S}=10$.

\begin{figure}[h]
    \centering
    \includegraphics[width=0.7\linewidth]{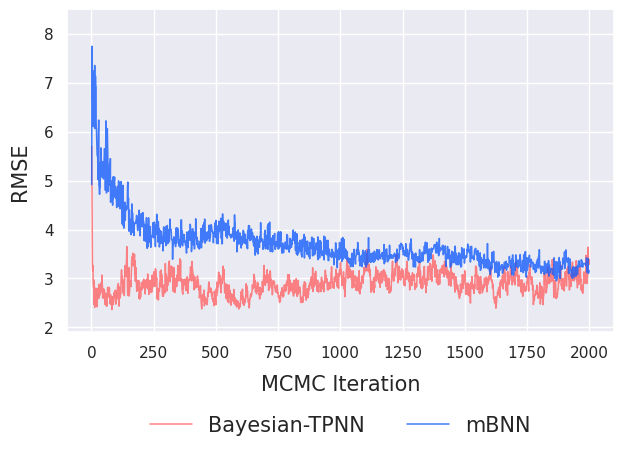}
    \vskip -0.3cm
    \caption{\footnotesize \textbf{The RMSE trajectories across MCMC iterations for Bayesian-TPNN and mBNN.}}
    \label{fig:mcmc-conv-speed-boston}
\end{figure}

Figure \ref{fig:mcmc-conv-speed-boston} shows the RMSE trajectories across MCMC iterations on \textsc{Boston} dataset for Bayesian-TPNN and mBNN.
Table \ref{table:run_time} presents the runtime comparison of Bayesian-TPNN , mBNN with 2,000 iterations and ANOVA-T$^{2}$PNN with 2,000 epochs on real datasets.
The best results are highlighted by \textbf{bold}.
In the experiments on \textsc{Fico}, \textsc{Churn}, and \textsc{Breast} datasets, the runtime difference between Bayesian-TPNN and ANOVA-T$^{2}$PNN become more pronounced. 
This is because, after data preprocessing, the input dimensions are 23, 46, and 30, respectively.
As the number of neural networks required in ANOVA-T$^{2}$PNN increases rapidly with the input dimension, the runtime increases considerably.
Note that for the \textsc{Madelon} dataset, where the input dimension is 500, training ANOVA-T$^{2}$PNN is infeasible because the number of neural networks to be trained is $125,250$.

These results imply that Bayesian-TPNN converges faster in terms of MCMC iterations compared to mBNN. Moreover, its overall runtime is shorter than both mBNN and ANOVA-T$^{2}$PNN. In particular, Bayesian-TPNN runs significantly faster than ANOVA-T$^{2}$PNN, and this advantage becomes more pronounced as the input dimension $p$ increases.

\begin{table}[h]
\centering
\scriptsize
\caption{\footnotesize \textbf{Runtime of Bayesian-TPNN, ANOVA-T$^{2}$PNN and mBNN (sec).}}
\label{table:run_time}
\vskip -0.3cm
\begin{tabular}{cccc}
\toprule
Dataset & Bayesian-TPNN & ANOVA-T$^{2}$PNN &mBNN \\ \midrule
\textsc{Abalone} & 475 & \textbf{326} & 1,273 \\ \midrule
\textsc{Boston} & \textbf{181} & 577 & 266 \\ \midrule
\textsc{Mpg} & \textbf{156} & 227 & 275 \\ \midrule
\textsc{Servo} & \textbf{159} & 400 & 242 \\ \midrule
\textsc{Fico} & \textbf{943} & 3,530 & 4,198 \\ \midrule
\textsc{Breast} & \textbf{181} & 2,363 & 310 \\ \midrule
\textsc{Churn} & \textbf{686} & 7,772 & 2,756 \\ \midrule
\textsc{Madelon} & \textbf{345} & --- & 894 \\ \bottomrule
\end{tabular}
\end{table}

\newpage

\section{Additional experiments for uncertainty quantification}
\label{sec:unvertainty_other}

\subsection{Uncertainty quantification on non-Bayesian models.}
\label{sec:unvertainty_other_1}

We report the performance of uncertainty quantification for non-Bayesian models including ANOVA-TPNN, NAM, XGB and Linear model, in Table \ref{table:results_uncetainty_other}.
% For the regression task, we estimate the error variance using the residual sum of squares and computed the negative log-likelihood (NLL).
These results indicate that Bayesian-TPNN outperforms the non-bayesian models in view of uncertainty quantification.

\begin{table}[h]
\scriptsize
\centering
\setlength{\tabcolsep}{6pt}
\caption{\footnotesize{\textbf{Uncertainty quantifications for non-bayesian models on real datasets.}}}
\label{table:results_uncetainty_other}
\vskip -0.3cm
\scalebox{0.85}{
\begin{tabular}{c c c c c c c c c}
\toprule
Dataset & \multicolumn{2}{c}{ANOVA-TPNN} & \multicolumn{2}{c}{NAM} & \multicolumn{2}{c}{XGB} & \multicolumn{2}{c}{Linear} \\ \midrule
 & CRPS & NLL & CRPS & NLL & CRPS & NLL & CRPS & NLL\\ \midrule
\textsc{Abalone} & 1.578 (0.16) & --- &  1.901 (0.27) & --- & 1.668 (0.16) &  --- & 1.638 (0.15) & --- \\ 
\textsc{Boston}  & 4.464 (0.71) & --- & 3.147 (0.35) & --- & 3.241 (0.27) & --- & 4.291 (0.44) & --- \\ 
\textsc{Mpg} & 2.478 (0.45) & --- & 3.314 (1.07) & --- & 2.343 (0.35) & --- & 2.990 (0.32) & --- \\ 
\textsc{Servo}   & 0.595 (0.02) & --- & 0.868 (0.39) & --- & 0.215 (0.03)  & --- & 0.910 (0.04) & --- \\  \midrule
 & ECE & NLL & ECE & NLL & ECE & NLL & ECE & NLL\\ \midrule
\textsc{Fico} & 0.063 (0.017) & 0.583 (0.018) & 0.198 (0.007) & 0.681 (0.012) & 0.096 (0.026) & 0.620 (0.015) & 0.055 (0.014) & 0.593 (0.017)\\ 
\textsc{Breast} & 0.100 (0.030) &  0.423 (0.071) & 0.284 (0.022) & 0.511 (0.033) & 0.063 (0.012) & 0.878 (0.172) & 0.102 (0.015) & 0.216 (0.039)\\ 
\textsc{Churn} & 0.053 (0.004) &  0.444 (0.011) & 0.318 (0.007) & 0.718 (0.008) & 0.131 (0.006) & 0.594 (0.021) & 0.078 (0.004) & 0.573 (0.002) \\ 
\textsc{Madelon} & 0.354 (0.014) &  0.752 (0.003) & 0.156 (0.009) & 0.735 (0.016) & 0.147 (0.008) & 0.703 (0.035) & 0.232 (0.011) & 0.736 (0.016) \\ \bottomrule
\end{tabular}
}
\end{table}

\subsection{Experiment for out-of-distribution detection}

Here, we conduct experiments to evaluate whether each model appropriately captures uncertainty on out-of-distribution data in binary classification.
As a measure of uncertainty for out-of-distribution data, we use the maximum predicted probability \citep{mukhoti2023deep}.
Specifically, we denote the in-distribution dataset by $\{\mathbf{x}^{\text{in}}_{1},...,\mathbf{x}^{\text{in}}_{N_{1}}\}$ and the out-of-distribution dataset by $\{\mathbf{x}^{\text{out}}_{1},...,\mathbf{x}^{\text{out}}_{N_{2}}\}$ with corresponding predictive probabilities $\hat{p}(\mathbf{x}^{\text{in}}_{i})$ for $i=1,...,N_{1}$ and $\hat{p}(\mathbf{x}^{\text{out}}_{i})$ for $i=1,...,N_{2}$. 

Let $\hat{p}_{\max}(\mathbf{x}) = \max \{\hat{p}(\mathbf{x}) , 1 - \hat{p}(\mathbf{x}) \}$.
For evaluation, we assign label 1 to the in-distribution data and label 0 to the out-of-distribution data.
Then, we compute the AUROC between the labels and the transformed scores $1+\log_{2}\hat{p}_{\max}(\mathbf{x}_{i}^{\text{in}})$ or $1+\log_{2} \hat{p}_{\max}(\mathbf{x}_{i}^{\text{out}})$.
Intuitively, predictive probabilities close to 0.5 reflect model uncertainty, and such cases can be identified as out-of-distribution.

We randomly sample a subset which size of 500 from the \textsc{Madelon} dataset, standardized it,
and use it as an out-of-distribution dataset.
For each dataset \textsc{Fico}, \textsc{Breast}, and \textsc{Churn}, we randomly split the data into training and test datasets.
In turn, we train Bayesian-TPNN and baseline models using the training dataset.
We then compute the AUROC treating the test dataset as the in-distribution dataset.
We repeat this procedure 5 times, and Table \ref{table:ood_result} presents the averages and standard errors of AUROCs for Bayesian-TPNN and baseline models on \textsc{Fico}, \textsc{Breast} and \textsc{Churn} datasets.
These results demonstrate that Bayesian-TPNN outperforms the baseline models, achieving substantially superior performance in out-of-distribution detection.

\begin{table}[h]
\centering
\scriptsize
\caption{\footnotesize \textbf{AUROC Results on in-distribution and out-of-distribution detection.}}
\label{table:ood_result}
\vskip -0.3cm
\begin{tabular}{cccc ccc c}
\toprule
Dataset & Bayesian-TPNN & ANOVA-TPNN & NAM & Linear & XGB & BART & mBNN\\ \midrule
\textsc{Fico}  & 0.606 (0.013) & 0.446 (0.020) & 0.455 (0.032) & 0.191 (0.002) & 0.605 (0.018) & \textbf{0.667} (0.004) & 0.519 (0.014)\\ 
\textsc{Breast} & \textbf{0.903} (0.015) & 0.542 (0.021) & 0.534 (0.041) & 0.112 (0.010) & 0.827 (0.022)  & 0.664 (0.023) & 0.503 (0.051) \\ 
\textsc{Churn} & \textbf{0.724} (0.006)  & 0.570 (0.040) & 0.533 (0.040) & 0.442 (0.006) & 0.420 (0.014) & 0.598 (0.009) & 0.599 (0.039)\\ \bottomrule
\end{tabular}
\end{table}

\newpage
\section{Visual illustration for Proposal}
\label{sec:evolution_explain}
\begin{figure}[h]
    \centering
    \includegraphics[width=0.4\linewidth]{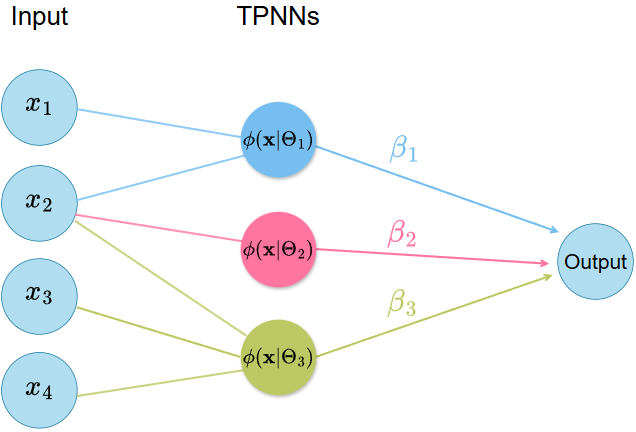}
    \caption{\textbf{Bayesian-TPNN with \(p=4, K=3\)}.}
    \label{fig:given_btpnn}
\end{figure}

In this section, we describe the visual explanation of the proposal in Section \ref{sec:mcmc}.
Given Bayesian-TPNN as in Figure \ref{fig:given_btpnn}, we explain the updating of $K$ and the updating of $S_{K}$.

% The procedure of making a new proposal associated with the model structure consists of two steps -- update \(K\) and update \(S_k\ \forall k\in[K]\).

\subsection{Updating \(K\)}
For a given \(K\), we propose \(K^\textup{new}=K-1\) or \(K^\textup{new}=K+1\).
Here, we describe only \textbf{Random} and \textbf{Stepwise} moves, corresponding to the case where $K^{\text{new}}=K+1$.

\begin{figure}[h]
    \centering
    \includegraphics[width=0.99\linewidth]{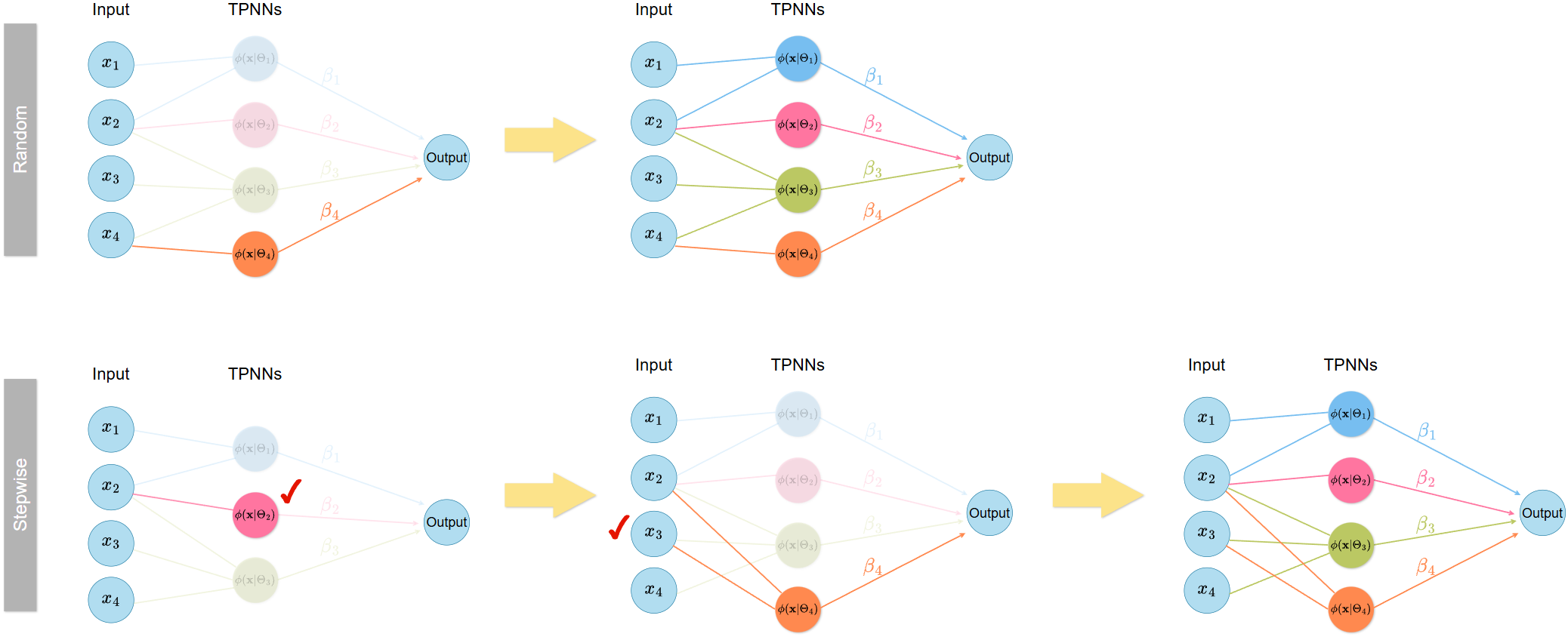}
    \caption{\textbf{Visual explanation for alternations in the proposal distribution of $K$.}}
    \label{fig:proposal-proc}
\end{figure}

In the case of \textbf{Random} move, a node is randomly generated and its edges are randomly assigned.
For \textbf{Stepwise} move, a node is first selected from the existing nodes, and then a new edge is added to create a new node.
Figure \ref{fig:proposal-proc} presents an overall illustration for these moves.

% At \textbf{Random} alternation, we generate \(S_{K^\textup{new}}\) from the prior distribution. That is, the variable set is chosen independently with the existing variables sets, \(S_1,\dots,S_K\). As can be seen in the first row of the figure \ref{fig:proposal-proc}, the orange-colored node 

\subsection{Updating \(S_k\)}
\begin{figure}[h]
    \centering
    \includegraphics[width=0.8\linewidth]{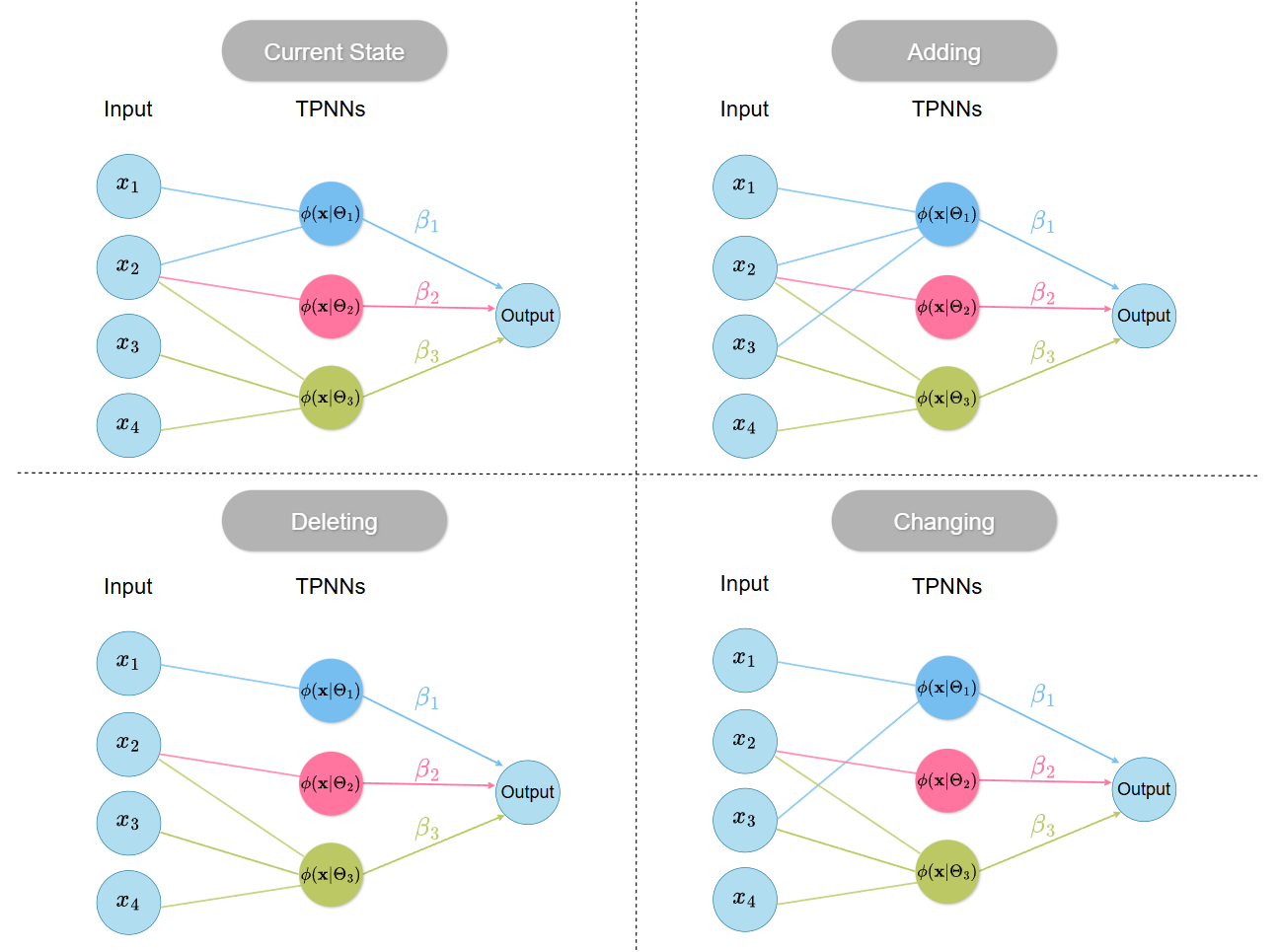}
    \caption{\textbf{Visual explanation for alternations in the proposal distribution of $S_{k}$.}}
    \label{fig:alternations_SK}
\end{figure}

Figure \ref{fig:alternations_SK} illustrates how the edges change when applying \textbf{Adding}, \textbf{Deleting}, or \textbf{Changing} moves to a given current state.

\newpage

\section{Empirical Evaluation under Minibatch Settings}
We conduct an additional experiment to empirically verify that our MCMC algorithm performs well when mini-batches are used.
When estimating Bayesian-TPNN with mini-batched data, we refer to it as MBayesian-TPNN.
Here, for $\textsc{Abalone}$ and \textsc{FICO} datasets, we set the size of mini-batch as 1,000 and 2,000, respectively. 
Table \ref{table:mbayeisan} presents the averages and standard errors of prediction performance and the uncertainty quantifications of Bayesian-TPNN and MBayesian-TPNN for 5 trials on \textsc{Abalone} and \textsc{FICO} datasets.
These results suggest that training with mini-batches does not significantly reduce prediction performance and uncertainty quantification.
In practice, these findings indicate that using mini-batches is practically acceptable, as it does not lead to meaningful degradation in performance or uncertainty estimation.

\begin{table}[h]
\scriptsize
\centering
\caption{Results of MBayesian-TPNN.}
\label{table:mbayeisan}
\begin{tabular}{ccccccc}
\toprule
 & \multicolumn{2}{c}{RMSE/AUROC} & \multicolumn{2}{c}{CRPS/ECE} & \multicolumn{2}{c}{NLL} \\ \midrule
 & \multicolumn{1}{c}{Bayesian-TPNN} & MBayesian-TPNN & \multicolumn{1}{c}{Bayesian-TPNN} & MBayesian-TPNN & \multicolumn{1}{c}{Bayesian-TPNN} & MBayesian-TPNN \\ \midrule
\textsc{Abalone} & \multicolumn{1}{c}{2.053 (0.26)} & 2.081 (0.24) & \multicolumn{1}{c}{1.372 (0.19)} & 1.391 (0.17) & \multicolumn{1}{c}{2.260 (0.16)} & 2.280 (0.18) \\ \midrule
\textsc{FICO} & \multicolumn{1}{c}{0.793 (0.009)} & {0.788 (0.005)} & \multicolumn{1}{c}{0.036 (0.004)} & {0.038 (0.003)} & \multicolumn{1}{c}{0.554 (0.007)} &  {0.564 (0.003)}\\ \bottomrule
\end{tabular}%
\end{table}

\newpage

\section{Comparison with Deep Ensemble}
In this section, we conduct additional experiment to compare Bayesian-TPNN with Deep Ensemble \citep{lakshminarayanan2017simple}.
Here, we consider candidates for each hyperparmeter of Deep Ensemble as below.
\begin{itemize}
    \item The number of MLPs : $\{5, 50, 100\}$
    \item MLP architectures : $\{(50), (100), (256,128,64), (512,256,128) \}$
    \item Learning rates : $\{1e-4,1e-3,1e-2 \}$
    \item Epochs : $\{100, 200, 500, 1000 \}$
    \item Weight for $L_{2}$ regularization : $\{1e-3,1e-2,1e-1 \}$
\end{itemize}
Table \ref{table:deepensemble} presents the averages of RMSE, AUROC, CRPS, ECE and NLLs for 5 trials on real datasets. 
These results show that the performance of Bayesian-TPNN is comparable to that of Deep Ensemble in terms of both prediction accuracy and uncertainty quantification.
\begin{table}[h]
\caption{Results of Bayesian-TPNN and Deep Ensemble.}
\label{table:deepensemble}
\resizebox{\textwidth}{!}{%
\begin{tabular}{ccccccc}
\toprule
 & \multicolumn{2}{c}{RMSE/AUROC} & \multicolumn{2}{c}{CRPS/ECE} & \multicolumn{2}{c}{NLL} \\ \midrule
 & \multicolumn{1}{c}{Bayesian-TPNN} & Deep Ensemble & \multicolumn{1}{c}{Bayesian-TPNN} & Deep Ensemble & \multicolumn{1}{c}{Bayesian-TPNN} & Deep Ensemble \\ \midrule
\textsc{Abalone} & \multicolumn{1}{c}{\textbf{2.053} (0.26)} & {2.121 (0.23)} & \multicolumn{1}{c}{\textbf{1.372} (0.19)} & {1.498 (0.17)} & \multicolumn{1}{c}{2.260 (0.16)} & {\textbf{2.036} (0.15)} \\
\textsc{Boston} & \multicolumn{1}{c}{\textbf{3.654} (0.49)} & {3.922 (0.57)} & \multicolumn{1}{c}{\textbf{2.202} (0.23)} & {2.458 (0.22)} & \multicolumn{1}{c}{\textbf{3.411} (0.37)} & {3.747 (0.40)}\\
\textsc{Mpg} & \multicolumn{1}{c}{2.386 (0.41)} & {\textbf{2.257} (0.14)} & \multicolumn{1}{c}{1.510 (0.43)} & {\textbf{1.481} (0.11)} & \multicolumn{1}{c}{\textbf{2.511} (0.21)} & {2.769 (0.47)} \\
\textsc{Servo} & \multicolumn{1}{c}{\textbf{0.351} (0.02)} & {0.398 (0.03)} & \multicolumn{1}{c}{0.194 (0.01)} & {\textbf{0.179} (0.01)} & \multicolumn{1}{c}{0.836 (0.10)} & {\textbf{0.701} (0.04)} \\
\midrule
\textsc{FICO} & \multicolumn{1}{c}{\textbf{0.793} (0.009)} & \multicolumn{1}{c}{0.773 (0.024)} & \multicolumn{1}{c}{\textbf{0.036} (0.004)} & \multicolumn{1}{c}{0.057 (0.033)} & \multicolumn{1}{c}{\textbf{0.554} (0.007)} & {0.577 (0.034)} \\
\textsc{Breast} & \multicolumn{1}{c}{\textbf{0.998} (0.001)} & \multicolumn{1}{c}{0.993 (0.003)} & \multicolumn{1}{c}{0.129 (0.009)} & \multicolumn{1}{c}{\textbf{0.075} (0.017)} & \multicolumn{1}{c}{0.211 (0.014)} & {\textbf{0.133} (0.041)} \\
\textsc{Churn} & \multicolumn{1}{c}{\textbf{0.849} (0.008)} & \multicolumn{1}{c}{0.841 (0.013)} & \multicolumn{1}{c}{\textbf{0.031} (0.001)} & \multicolumn{1}{c}{0.039 (0.002)} & \multicolumn{1}{c}{\textbf{0.418} (0.008)} & {0.424 (0.018)} \\
\textsc{Madelon} & \multicolumn{1}{c}{\textbf{0.854} (0.013)} & \multicolumn{1}{c}{0.616 (0.029)} & \multicolumn{1}{c}{\textbf{0.076} (0.004)} & \multicolumn{1}{c}{0.137 (0.061)} & \multicolumn{1}{c}{\textbf{0.478} (0.009)} & {0.719 (0.049)} \\ \bottomrule
\end{tabular}%
}
\end{table}

\newpage

\section{Applications to Genomic Dataset}
We conduct additional experiment to explore the applicability of Bayesian-TPNN to genomics dataset GSE43358 \citep{GSE43358}.
GSE43358 is a gene expression dataset with $n=57$ samples and $p=54,675$ features and we perform a classification task distinguishing between HER2-positive and non–HER2-positive cases.
Table \ref{table:prediction_perform_geo} shows that the averages and standard errors of prediction performance for Bayesian-TPNN, Linear model and XGB for 5 trials.
For Bayesian-TPNN and XGB, the hyperparameters are optimized as in the experiment for other real datasets.
Note that because the input dimension $p$ is too large, both ANOVA-TPNN and NAM could not be trained within our computational environment.
The results in Table \ref{table:prediction_perform_geo} indicate that the interpretable Bayesian-TPNN achieves prediction performance comparable to that of the black-box model XGB on GSE43358 dataset.
$\newline$
$\newline$
Table \ref{table:geo_important_comp} reports the top 10 most important components in Bayesian-TPNN with the normalized importance score.
Here, we use the importance score defined in Section 4.2, and the normalized score represents each importance value divided by the maximum importance score.
Note that one of the third order interactions is detected by Bayesian-TPNN.
The results in Table \ref{table:geo_important_comp} indicate that higher-order interactions (beyond the second order) play a crucial role, which provides a plausible explanation for the inferior prediction performance of the linear model.
Moreover, this highlights the necessity of an interpretable model such as Bayesian-TPNN, which is capable of estimating such higher-order interactions.

\begin{table}[h]
\centering
\scriptsize
\caption{Results of baseline models on GSE43358 dataset.}
\label{table:prediction_perform_geo}
\begin{tabular}{cccccc}
\toprule
Model & Bayesian-TPNN & ANOVA-TPNN & NAM & Linear & XGB \\ \midrule
AUROC & 0.949 (0.017) & -- & -- & 0.545 (0.001) & 0.953 (0.041)  \\ \bottomrule
\end{tabular}%
\end{table}

\begin{table}[h]
\centering
\scriptsize
\caption{Top 10 important components.}
\label{table:geo_important_comp}
\begin{tabular}{ccc}
\toprule
Rank & Component of GenBank accession numbers & Normalized Score \\ \midrule
1 & S69189 & 1.000 \\
2 & BF357738  & 0.924 \\
3 & (BC000129, R80390) & 0.701 \\
4 & AF307338 & 0.569 \\
5 & NM\_018297 & 0.410 \\
6 & BF061275 & 0.375 \\
7 & AF319440 & 0.365 \\
8 & (BE741754, AB037854, AK024890) & 0.334 \\
9 & AI368358 & 0.292 \\
10 & BE672684 & 0.218 \\
\bottomrule
\end{tabular}%

\end{table}

\newpage

\section{Notations and regularity conditions for the proofs}

\subsection{Additional Notations}

For two positive sequences $\{a_{n}\}$ and $\{b_{n}\}$, we write $a_{n} \lesssim b_{n}$ if there exists a constant $C > 0$ such that $a_{n} \leq Cb_{n}$ for all $n \in \mathbb{N}$.
The notation $a_{n}=o(b_{n})$ indicates that the ratio $a_{n}/b_{n}$ converges to zero as $n \xrightarrow{} \infty$. 
We denote $\mathcal{N}(\epsilon,\mathcal{F},d)$ the $\epsilon$-covering number of the function class $\mathcal{F}$ with respect to the semimetric $d$.
For a given vector $\mathbf{v}=(v_{1},...,v_{N})$, we define its $\ell_{2}$ norm as $\Vert \mathbf{v} \Vert_{2}^{2} := \sum_{i=1}^{N}v_{i}^{2}$. 
Given a real-valued function $f: \mathcal{X} \to \mathbb{R}$, we define its sup-norm as $\Vert f\Vert_{\infty} := \sup_{\mathbf{x} \in \mathcal{X}} |f(\mathbf{x})|$.
We define population $\ell_{p}$-norm with respect to a probability measure $\mu$ on $\mathcal{X}$ as $\Vert f \Vert_{p,\mu} := (\int_{\mathbf{x} \in \mathcal{X}}f(\bold{x})^{p}\mu(d\mathbf{x}))^{1/ p}$.
Let $\mathbb{P}_{\mathbf{X}}^{n} = \prod_{i=1}^{n}$, where $\mathbb{P}_{\mathbf{X}_{i}}$ is the probability distribution of $\mathbf{X}_{i}$ for $i=1,...,n$.
For two given densities $p_{1}$ and $p_{2}$, we define the Kullback-Leibler (KL) divergence as
$$
K(p_{1},p_{2}) :=\int \log(p_{1}(\bold{v}) / p_{2}(\bold{v}))p_{1}(\bold{v})d\bold{v},
$$
and let $V(p_{1},p_{2}):=\int | \log (p_{1}(\bold{v})/p_{2}(\bold{v})) - K(p_{1},p_{2})|^{2} p_{1}(\bold{v})d\bold{v}.$

\subsection{Regularity Conditions}
\label{app:regular_condition}

\begin{enumerate}[label=$(S.\arabic*)$ ]
\item For a distribution $\mathbb{P}_{\mathbf{X}}$, there exist a density $p_{\mathbf{X}}$ with respect to the Lebesgue measure on $\mathbb{R}^p$, that is bounded away from zero and infinity, i.e.,
$$
0< \inf_{\bold{x}\in \mathcal{X}}p_{\mathbf{X}}(\bold{x}) \le \sup_{\bold{x}\in \mathcal{X}}p_{\mathbf{X}}(\bold{x})<\infty.
$$
\label{eq:Assumption_1}
\item The true function $f_{0,S}$ is L-Lipschitz continuous, i.e.,
\begin{align*}
|f_{0,S}(\mathbf{x}) - f_{0,S}(\mathbf{x}')| \leq L \Vert \mathbf{x} - \mathbf{x}' \Vert_{2} 
 \end{align*}
for some positive constant $L$ and all $\mathbf{x},\mathbf{x}' \in \mathcal{X}$.
Additionally, $f_{0,S}$ is assumed to be bounded in the supremum norm by a positive constant $F$, i.e., $\Vert f_{0,S}\Vert_\infty \le F$. 
We denote the above conditions compactly as $f_{0,S}\in \text{Lip}_{L,F}.$
Moreover, we say that $f_{0} \in \text{Lip}_{0,L,F}$ if $f_{0,S} \in \text{Lip}_{L,F}$ for all $S \subseteq [p]$.
 \label{eq:Assumption_2}
 \item The log-partition function $A(\cdot)$ is differentiable with a bounded second derivative over $[-F,F]$, i.e., there exists a positive constant $C_{A}$ such that
 \begin{align*}
     1/C_{A} \leq \ddot{A}(x) \leq C_{A}
 \end{align*}
 for all $x \in [-F,F]$.
 \label{eq:Assumption_5}
 % \item Suppose there exist constants $\sigma^{2}_{\min} > 0$ and $\sigma^{2}_{\max} < \infty$ such that $\sigma_0^2\in (\sigma^2_{\min},\sigma^2_{\max}).$
 %  \label{eq:Assumption_3}
 \item $K_{\max}$ is assumed to grow at a rate $K_{\max}=O(n)$. \label{eq:Assumption_4}
\end{enumerate}

\newpage

\section{Posterior consistency of Bayesian-TPNN}

We first prove the posterior consistency of $f$ since it plays an important role
in the proof of the posterior consistency of each component $f_S.$

\subsection{Posterior consistency of $f_0$}

\begin{theorem}[Posterior Consistency of Bayesian-TPNN]
\label{thm:posterior_rate_exponetial}
We assumes that \ref{eq:Assumption_1}, \ref{eq:Assumption_2}, \ref{eq:Assumption_5} and \ref{eq:Assumption_4}.
Then, for any $\varepsilon > 0$ and $\xi\geq 2^{p}F + \varepsilon\sqrt{2\over C_{A} },$ it holds that
\begin{equation}
\label{eq:con-rate}
\pi_{\xi}\Big(f : \Vert f_{0}-f \Vert_{2,n}  > \varepsilon \Big| \mathbf{X}^{(n)},Y^{(n)}\Big) \xrightarrow{} 0
\end{equation}
in $\mathbb{Q}_0^{n}$ as $n \xrightarrow{} \infty,$
where $\mathbb{Q}_{0}^{n}$ is the probability distribution of $(\mathbf{X}^{(n)},Y^{(n)})$.
\end{theorem}

\subsection{Proof outline}

%Since we transform the input variables $\{x_{1,j}, \ldots, x_{n,j}\}$ into quantiles via the marginal empirical distribution for $j=1,...,p$, Bayesian-TPNN can be regarded as satisfying the sum-to-zero condition for data sampled from the uniform distribution.
%Accordingly, we show the posterior consistency for Bayesian-TPNN that satisfies the sum-to-zero condition with respect to the uniform distribution on $(0,1)$, and we expect that this result can be easily extended to the corresponding function class of Bayesian-TPNN which satisfies the sum-to-zero condition with data generated from $U(0,1)$.

Consider a function class $\mathcal{F} = \bigcup_{K=1}^{K_{\max}}\mathcal{F}(K)$ that satisfies the sum-to-zero condition with respect to uniform distribution on (0,1).
Here, $\mathcal{F}(K)$ is defined as
\begin{align*}
\mathcal{F}(K) = \bigg\{f : &f(\mathbf{x}) = 
\sum_{k=1}^{K}\beta_{k}\phi (\mathbf{x}|S_{k},\mathbf{b}_{S_{k},k},\Gamma_{S_{k},k}), \\ 
&\beta_{k} \in \mathbb{R}, \\
&\mathbf{b}_{S_{k},k} \in [0,1]^{|S_{k}|} , \\
&\Gamma_{S_{k},k} \in [1/n,\infty)^{|S_{k}|} \:\: \text{for} \:\: k=1,...,K \bigg\},
\end{align*}
where $$
\phi(\mathbf{x}|S_{k},\mathbf{b}_{S_{k},k},\Gamma_{S_{k}},k) = \prod_{j \in S_{k}}\bigg ( 1 - \sigma\bigg({x_{j} - b_{j,k}\over \gamma_{j,k}}\bigg) + c_{j}(b_{j,k},\gamma_{j,k})\sigma\bigg({x_{j} - b_{j,k}\over \gamma_{j,k}}\bigg) \bigg)
$$ 
and
$$
c_{j}(b_{j,k},\gamma_{j,k})  = -{1 - \int_{0}^{1}\sigma\bigg({x_{j} -b_{j,k}\over \gamma_{j,k}}\bigg)dx_{j} \over  \int_{0}^{1}\sigma\bigg({x_{j} -b_{j,k}\over \gamma_{j,k}}\bigg)dx_{j}}.
$$

For any $f \in \mathcal{F}(K)$, we denote it as $f_{K,\mathcal{B},\mathbf{b},\Gamma}$, where
\begin{align*}
\mathcal{B} =(\beta_{k} , k \in [K]), \quad \mathbf{b}=(\mathbf{b}_{S_{k},k}, k \in [K]) \quad \text{and} \quad  \Gamma = (\Gamma_{S_{k},k},k \in [K]).
\end{align*} 
Our goal is to show that
\begin{align}
\lim_{n \to \infty}\mathbb{E}_{0}^{n}[\pi_{\xi}(\Vert f - f_{0} \Vert_{2,n}  >  \varepsilon | \mathbf{X}^{(n)},Y^{(n)}) ] = 0 \label{eq:final_goal}
\end{align}
for any $\varepsilon > 0$.

We prove (\ref{eq:final_goal}) using following two steps.

\begin{enumerate}[label=$(P.\arabic*)$ ]
    \item For given data $\mathbf{x}^{(n)}$, we prove that
    \begin{align*}
\lim_{n \to \infty}\mathbb{E}_{0}^{n}[\pi_{\xi}(\Vert f - f_{0} \Vert_{2,n} > \varepsilon | \mathbf{X}^{(n)},Y^{(n)}) | \mathbf{X}^{(n)} =\mathbf{x}^{(n)}] = 0
\end{align*}
for any $\varepsilon > 0$.  \label{eq:Step_N2}
\item Finally, we show that
\begin{align*}
\lim_{n \to \infty}\mathbb{E}_{0}^{n}[\pi_{\xi}(\Vert f - f_{0} \Vert_{2,n} > \varepsilon | \mathbf{X}^{(n)},Y^{(n)})] = 0
\end{align*}
for any $\varepsilon > 0$. \label{eq:Step_N3}
\end{enumerate}

We first verify the following three conditions: there exists $\mathcal{F}^{n} \subseteq \mathcal{F}$ and positive constants $\delta, C_{1}, C_{2}$ such that
\begin{align}
& \log \mathcal{N}\left(\delta,\mathcal{F}^{n},\Vert \cdot \Vert_{\infty} \right) < nC_{1}, \label{eq:ghosal_condition1}\\
&\pi\bigg(f \in \mathcal{F} : \Vert f - f_{0} \Vert_{\infty} \leq \varepsilon\sqrt{2 \over C_{A}}\bigg) > \exp(-nC_{2}), \label{eq:ghosal_condition2}
\\
&\pi(\mathcal{F}\backslash \mathcal{F}^{n}) < \exp(-(2C_{2}+2)n).
\label{eq:ghosal_condition3}
\end{align}
After that, we will show that these three conditions imply the posterior consistency in Step \ref{eq:Step_N2} by checking the conditions in \cite{ghosal1999posterior}.

\subsection{Verifying Condition (\ref{eq:ghosal_condition1})}
\label{sec:prof_ghosal_cond1}
We consider a sieve $\mathcal{F}^{n} = \cup_{K=1}^{M_{n}}\mathcal{F}^{n}(K),$
where 
\begin{align*}
\mathcal{F}^{n}(K) = \bigg \{ f: &f(\mathbf{x}) = \sum_{k=1}^{K}\beta_{k}\phi(\mathbf{x}|S_{k},\mathbf{b}_{S_{k},k},\Gamma_{S_{k},k}) ,\\
&\beta_{k} \in [-n,n], \\ 
&\mathbf{b}_{S_{k},k} \in [0,1]^{|S_{k}|}\\
& \Gamma_{S_{k},k} \in [1/n,n]^{|S_{k}|}
\quad \text{for} \quad k=1,..,K  \bigg\},
\end{align*}
where $M_{n} = \lfloor {C_{3}n\varepsilon^{2} \over \log n }\rfloor$ and $C_{3}$ will be determined later.

Also, we consider a more general function class as :
\begin{align}
\begin{split}
\mathcal{G}^{n}(K) = \bigg \{ f : f(\mathbf{x}) &= \sum_{k=1}^{K}\beta_{k}\phi(\mathbf{x}|S_{k},\mathbf{b}_{S_{k},k},\Gamma_{S_{k},k},\mathbf{c}_{S_{k},k}) ,\\
&\beta_{k} \in [-n,n], \\ 
&\mathbf{b}_{S_{k},k} \in [0,1]^{|S_{k}|} , \\
&\Gamma_{S_{k},k}\in [1/n,n]^{|S_{k}|}, \\
&\mathbf{c}_{S_{k},k} \in [-2n,2n]^{|S_{k}|}
\quad \text{for} \quad k=1,..,K \bigg\},
\end{split} \label{eq:g_n}
\end{align} 
where the function $\phi$ is defined as
\begin{align*}
\phi(\mathbf{x}|S_{k},\mathbf{b}_{S_{k},k},\Gamma_{S_{k},k},\mathbf{c}_{S_{k},k}) = \prod_{j \in S_{k}}\bigg( 1 - \sigma\bigg({x_{j}-b_{j,k}\over \gamma_{j,k}}\bigg) + c_{j,k} \sigma\bigg({x_{j}-b_{j,k}\over \gamma_{j,k}}\bigg) \bigg).
\end{align*} 
and the vector $\mathbf{c}_{S_{k},k}$ is defined as
$\mathbf{c}_{S_{k},k} = (c_{j,k}, j \in S_{k})$. 
%Note that $\mathcal{F}^{n}(K) \subset \mathcal{G}^{n}(K)$ but functions in $\mathcal{G}^{n}(K)$
%do not need to satisfy the sum-to-zero condition.

For all $j, k$, we have
\begin{align*}
\int_{0}^{1}\sigma\bigg({x-b_{j,k}\over \gamma_{j,k}}\bigg)dx &\geq \int_{b_{j,k}}^{1} \sigma\bigg({x-b_{j,k}\over \gamma_{j,k}}\bigg) dx \\
&\geq C_{\sigma,j,k},
\end{align*}
where $C_{\sigma,j,k}$ is a positive constant and thus, we have
$|c_{j}(b_{j,k},\gamma_{j,k})| \leq C_{\sigma}, \: \forall j,k$
for some positive constant $C_{\sigma}$. Hence, for all $K \in [K_{\max}]$,
\begin{align}
\mathcal{F}^{n}(K) \subseteq \mathcal{G}^{n}(K),
\end{align}
whenever $n$ is sufficiently large. Therefore, it suffices to verify Condition~(\ref{eq:ghosal_condition1}) over
\begin{align}
\mathcal{G}^{n} = \bigcup_{K=1}^{M_{n}} \mathcal{G}^{n}(K).
\end{align}

\begin{lemma}
\label{lemma:covering}
For any integer $K$, we have
\begin{align*}
\mathcal{N}(\epsilon, \mathcal{G}^{n}(K),\Vert \cdot \Vert_{\infty}) \leq \bigg (1 + { K2^{p+4}n^{3p+1}\over \epsilon} \bigg )^{K(1+3p)}.
\end{align*}
\end{lemma}
Proof.) \\
$\newline$
First, since the maximum dimension of parameters in $\mathcal{G}^{n}(K)$ is $K(1+3p)$, we consider $K(1+3p)$-dimensional hypercube $[-2n,2n]^{K(1+3p)}$. 
Then, we have
\begin{align*}
\mathcal{N}(\epsilon_{1},[-2n,2n]^{K(1+3p)},\Vert \cdot \Vert_{1}) &\leq \bigg( \mathcal{N}(\epsilon_{1},[-2n,2n],\Vert \cdot \Vert_{1})\bigg)^{K(1+3p)} \\
&\leq \bigg ( 1+ {4n \over \epsilon_{1}} \bigg)^{K(1+3p)}.
\end{align*}

For $\mathbf{S}_{K}=(S_{k}, k \in [K])$, we define 
$\mathfrak{S} := ( \mathcal{B}_{K}, \mathbf{b}_{\mathbf{S}_{K},K},\Gamma_{\mathbf{S}_{K},K}, \mathbf{c}_{\mathbf{S}_{K},K}),$
where 
\begin{align*}
\mathcal{B}_{K} &=(\beta_{1},...,\beta_{K}), \\
\mathbf{b}_{\mathbf{S}_{K},K} &= (\mathbf{b}_{S_{k},k}, k \in [K]), \\
\Gamma_{\mathbf{S}_{K},K} &= (\Gamma_{S_{k},k},k\in[K]), \\
\mathbf{c}_{\mathbf{S}_{K},K} &= (\mathbf{c}_{S_{k},k},k \in [K]).
\end{align*}

Let $\left\{\mathfrak{S}^{1},...,\mathfrak{S}^{\mathcal{N}(\epsilon_{1},[-n,n]^{K(1+3p)},\Vert \cdot \Vert_{1})}\right\}$
 be an $\epsilon_{1}$-cover of $[-2n,2n]^{K(1+3p)},$ and for given $\mathfrak{S} \in [-2n,2n]^{K(1+3p)},$
 let $\tilde{\mathfrak{S}}$ be an element in the $\epsilon_{1}$-cover such that $\|\mathfrak{S}-\tilde{\mathfrak{S}}\|_{1} \le \epsilon_{1}.$

Note that for any $f_{\Theta} \in \mathcal{G}^{n}(K)$, we have
\begin{align*}
f_{\mathfrak{S}}(\mathbf{x}) = \sum_{k=1}^{K}\beta_{k}\prod_{j \in S_{k}}\phi(x_{j}|\{j\},b_{j,k},\gamma_{j,k},c_{j,k}),
\end{align*}
where
$$
\phi(x_{j}|\{j\},b_{j,k},\gamma_{j,k},c_{j,k}) = 1 - \sigma\bigg({x_{j}-b_{j,k}\over \gamma_{j,k}}\bigg)  + c_{j,k}\sigma\bigg( {x_{j}-b_{j,k}\over \gamma_{j,k}} \bigg)
$$
with $|c_{j,k}| \leq 2n $.
Then, for any $f_{\mathfrak{S}} \in \mathcal{G}^{n}(K)$, we have
\begin{align}
&\sup_{\mathbf{x}}\bigg|f_{\mathfrak{S}}(\mathbf{x}) - f_{\Tilde{\mathfrak{S}}}(\mathbf{x}) \bigg | \nonumber\\
&\leq \sup_{\mathbf{x}} \sum_{k=1}^{K}\bigg|\beta_{k}\prod_{j \in S_{k}}\phi(x_{j}|\{j\},b_{j,k},\gamma_{j,k},c_{j,k}) - \Tilde{\beta}_{k}\prod_{j \in S_{k}}\phi(x_{j}|\{j\},\Tilde{b}_{j,k},\Tilde{\gamma}_{j,k},\Tilde{c}_{j,k}) \bigg | \nonumber\\
\begin{split}
&\leq \sup_{\mathbf{x}} \sum_{k=1}^{K}\bigg( \bigg|\beta_{k}\prod_{j \in S_{k}}\phi(x_{j}|\{j\},b_{j,k},\gamma_{j,k},c_{j,k}) - \Tilde{\beta}_{k}\prod_{j \in S_{k}}\phi(x_{j}|\{j\},b_{j,k},\gamma_{j,k},c_{j,k})\bigg | \\
&\quad\quad\quad\quad\:+ \bigg|\Tilde{\beta}_{k}\prod_{j \in S_{k}}\phi(x_{j}|\{j\},b_{j,k},\gamma_{j,k},c_{j,k}) - \Tilde{\beta}_{k}\prod_{j \in S_{k}}\phi(x_{j}|\{j\},\Tilde{b}_{j,k},\Tilde{\gamma}_{j,k},\Tilde{c}_{j,k}) \bigg | \bigg ). \label{eq:cov_num_eq1}
\end{split}
\end{align}
\paragraph{Upper bound of first term in (\ref{eq:cov_num_eq1}).} Since
\begin{align*}
\bigg |\prod_{j \in S_{k}}\phi(x_{j}|\{j\},b_{j,k},\gamma_{j,k},c_{j,k}) \bigg|&= \bigg|\prod_{j \in S_{k}}\bigg( 1 - \sigma\bigg({x_{j}-b_{j,k} \over \gamma_{j,k}}\bigg) + c_{j,k}\sigma\bigg({x_{j}-b_{j,k}\over \gamma_{j,k}}\bigg) \bigg) \bigg| \\
&\leq \prod_{j \in S_{k}}\bigg(\bigg| 1 - \sigma\bigg({x_{j}-b_{j,k} \over \gamma_{j,k}}\bigg) + c_{j,k}\sigma\bigg({x_{j}-b_{j,k}\over \gamma_{j,k}}\bigg) \bigg|\bigg) \\
&\leq \prod_{j \in S_{k}}(1+2n) \\
&\leq (1+2n)^{p},
\end{align*}
we have
\begin{align*}
&\sup_{\mathbf{x}} \sum_{k=1}^{K} \bigg|\beta_{k}\prod_{j \in S_{k}}\phi(x_{j}|\{j\},b_{j,k},\gamma_{j,k},c_{j,k}) - \Tilde{\beta}_{k}\prod_{j \in S_{k}}\phi(x_{j}|\{j\},b_{j,k},\gamma_{j,k},c_{j,k})\bigg| \\
&\leq \sup_{\mathbf{x}}\sum_{k=1}^{K}(1+2n)^{|S_{k}|}|\beta_{k}-\Tilde{\beta}_{k}| \\
&\leq (1+2n)^{p}\epsilon_{1}.
\end{align*}

\paragraph{Upper bound of second term in (\ref{eq:cov_num_eq1}).}
Using direct calculation and triangle inequality, we have
\begin{align*}
&\bigg|\tilde{\beta}_{k}\prod_{j \in S_{k}}\bigg(\phi(x_{j}|\{j\},b_{j,k},\gamma_{j,k},c_{j,k}) - \phi(x_{j}|\{j\},\Tilde{b}_{j,k},\Tilde{\gamma}_{j,k},\Tilde{c}_{j,k})\bigg)\bigg |  \\
&= \bigg|\Tilde{\beta}_{k}\prod_{j \in S_{k}}\bigg( \sigma\bigg({x_{j}-\Tilde{b}_{j,k}\over \Tilde{\gamma}_{j,k}}\bigg)- \sigma\bigg({x_{j}-b_{j,k}\over \gamma_{j,k}}\bigg) + c_{j,k}\sigma\bigg({x_{j}-b_{j,k}\over \gamma_{j,k}}\bigg) -  \Tilde{c}_{j,k}\sigma\bigg({x_{j}-\Tilde{b}_{j,k}\over \Tilde{\gamma}_{j,k}}\bigg) \bigg )\bigg | \\
&= |\Tilde{\beta}_{k}|\prod_{j \in S_{k}}\bigg | \sigma\bigg({x_{j}-\Tilde{b}_{j,k}\over \Tilde{\gamma}_{j,k}}\bigg)- \sigma\bigg({x_{j}-b_{j,k}\over \gamma_{j,k}}\bigg) + c_{j,k}\sigma\bigg({x_{j}-b_{j,k}\over \gamma_{j,k}}\bigg) -  \Tilde{c}_{j,k}\sigma\bigg({x_{j}-\Tilde{b}_{j,k}\over \Tilde{\gamma}_{j,k}}\bigg) \bigg | \\
&\leq n\prod_{j \in S_{k}}\bigg( \bigg | \sigma\bigg({x_{j}-\Tilde{b}_{j,k}\over \Tilde{\gamma}_{j,k}}\bigg)- \sigma\bigg({x_{j}-b_{j,k}\over \gamma_{j,k}}\bigg)\bigg | + \bigg | c_{j,k}\sigma\bigg({x_{j}-b_{j,k}\over \gamma_{j,k}}\bigg) -  \Tilde{c}_{j,k}\sigma\bigg({x_{j}-\Tilde{b}_{j,k}\over \Tilde{\gamma}_{j,k}}\bigg) \bigg | \bigg ).
\end{align*}

Since $\sigma(\cdot)$ is Lipschitz function, we have
\begin{align*}
&\bigg | \sigma\bigg({x_{j}-\Tilde{b}_{j,k}\over \Tilde{\gamma}_{j,k}}\bigg)- \sigma\bigg({x_{j}-b_{j,k}\over \gamma_{j,k}}\bigg)\bigg |\\
 &\leq \bigg | {x_{j} - \Tilde{b}_{j,k}\over \Tilde{\gamma}_{j,k}} - {x_{j} - b_{j,k}\over \gamma_{j,k}} \bigg | \\
 &\leq  \bigg( \bigg | {x_{j} - \Tilde{b}_{j,k}\over \Tilde{\gamma}_{j,k}} - {x_{j} - b_{j,k}\over \Tilde{\gamma}_{j,k}} \bigg | + \bigg | {x_{j} - b_{j,k}\over \Tilde{\gamma}_{j,k}} - {x_{j} - b_{j,k}\over \gamma_{j,k}} \bigg | \bigg) \\
 &\leq 2n^{2}\bigg( |\Tilde{b}_{j,k} - b_{j,k}|+ |\Tilde{\gamma}_{j,k} - \gamma_{j,k}| \bigg ). 
\end{align*}

Similarly, we have
\begin{align*}
&\bigg | c_{j,k}\sigma\bigg({x_{j}-b_{j,k}\over \gamma_{j,k}}\bigg) -  \Tilde{c}_{j,k}\sigma\bigg({x_{j}-\Tilde{b}_{j,k}\over \Tilde{\gamma}_{j,k}}\bigg) \bigg | \\
&\leq \bigg | c_{j,k}\sigma\bigg({x_{j}-b_{j,k}\over \gamma_{j,k}}\bigg) - \Tilde{c}_{j,k}\sigma\bigg({x_{j}-b_{j,k}\over \gamma_{j,k}}\bigg) \bigg | +\bigg| \Tilde{c}_{j,k}\sigma\bigg({x_{j}-b_{j,k}\over \gamma_{j,k}}\bigg) -\Tilde{c}_{j,k}\sigma\bigg({x_{j}-\Tilde{b}_{j,k}\over \Tilde{\gamma}_{j,k}}\bigg) \bigg | \\
&\leq 4n^{3}\bigg( |c_{j,k} - \Tilde{c}_{j,k}| + |\Tilde{b}_{j,k} - b_{j,k}|+ |\Tilde{\gamma}_{j,k} - \gamma_{j,k}| \bigg ).
\end{align*}

To sum up, the upper bound of (\ref{eq:cov_num_eq1}) is 
\begin{align*}
\sup_{\mathbf{x}}|f_{\mathfrak{S}}(\mathbf{x}) - f_{\Tilde{\mathfrak{S}}}(\mathbf{x})| &\leq K\bigg((1+2n)^{p}\epsilon_{1} +  2^{p+3}n^{3p+1}\epsilon_{1}^{p} \bigg)\\
&\leq K(2n)^{3p+1}\epsilon_{1}.
\end{align*}

Let $\epsilon =  K(2n)^{3p+1}\epsilon_{1}$. 
Then, we conclude that
\begin{align*}
\mathcal{N}(\epsilon, \mathcal{G}^{n}(K),\Vert \cdot \Vert_{\infty}) \leq \bigg (1 + { 2K(2n)^{3p+2} \over \epsilon} \bigg )^{K(1+3p)}.
\end{align*}
\qed

Using Lemma \ref{lemma:covering}, we have
\begin{align*}
\mathcal{N}(\delta,\mathcal{F}^{n}, \Vert \cdot \Vert_{\infty}) &\leq \sum_{K=1}^{M_{n}} \bigg (1 + { 2K(2n)^{3p+2} \over \delta} \bigg )^{K(1+3p)} \\
&\leq M_{n}\bigg (1 + { 2M_{n}(2n)^{3p+2} \over \delta} \bigg )^{M_{n}(1+3p)}.
\end{align*}
Let $\delta = \varepsilon / 8$.
Finally, we choose $C_{3}$ such that
\begin{align*}
\log \mathcal{N}(\delta,\mathcal{F}^{n}, \Vert \cdot \Vert_{\infty}) &\leq \log M_{n} + M_{n}(1+3p)\log \bigg( 1 + {2M_{n}(2n)^{3p+2} \over \delta } \bigg) \\
&< n\varepsilon^{2} / 10.
\end{align*}
Condition (\ref{eq:ghosal_condition1}) is satisfied by letting $C_{1} = {\varepsilon^{2} / 10}$.
\qed

\subsection{Verifying Condition (\ref{eq:ghosal_condition2})}
For $S \subseteq [p]$, using Theorem 3.3 in \cite{park2025tensor},
there exist TPNNs such that
\begin{align}
\Big\Vert f_{0,S} - f_{k_{S},\hat{\mathcal{B}}_{S,k_{n,S}},\hat{\mathbf{b}}_{S,k_{n,S}},\hat{\Gamma}_{S,k_{n,S}}} \Big\Vert_{\infty} \leq {C_{S} \over k_{n,S}^{1/|S|} +1}
\end{align}
for some positive constant $C_{S}$.
Here, $\hat{\beta}_{S,k}s$ are uniformly bounded, i.e., $|\hat{\beta}_{S,k}| \leq c_{S}$ for some positive constant $c_{S}$ and $\hat{\gamma}_{j,k} = 1/k_{n,S}^{3}$ for all $j,k$ as specified in Theorem 3.3 of \cite{park2025tensor}.

Let $k_{n,S}$ such that
\begin{align}
{C_{S} \over k_{n,S}^{1/|S|} +1} \leq \varepsilon\sqrt{2} / (\sqrt{C_{A}}\cdot3\cdot 2^{p}). \label{eq:num_tpnn}
\end{align}

Let $k_{n} = \sum_{S\subseteq[p]}k_{n,S}$ and $f_{k_{n},\hat{\mathcal{B}}_{k_{n}},\hat{\mathbf{b}}_{k_{n}},\hat{\Gamma}_{k_{n}}} = \sum_{S  \subseteq [p]}f_{k_{n,S},\hat{\mathcal{B}}_{S,k_{n,S}},\hat{\mathbf{b}}_{S,k_{n,S}},\hat{\Gamma}_{S,k_{n,S}}}.$
For notational simplicity, we write $\hat{\mathcal{B}}_{k_{n}}, \hat{\mathbf{b}}_{k_{n}}$ and $\hat{\Gamma}_{k_{n}}$ simply as $\hat{\mathcal{B}}$, $\hat{\mathbf{b}}$ and $\hat{\Gamma}$, respectively. 
Since
\begin{align}
&\Vert f_{0} - f_{k_{n},\mathcal{B},\mathbf{b},\Gamma} \Vert_{\infty} \nonumber\\
&\leq \Vert f_{0} - f_{k_{n},\hat{\mathcal{B}},\hat{\mathbf{b}},\hat{\Gamma}}  \Vert_{\infty} + \Vert f_{k_{n},\hat{\mathcal{B}},\hat{\mathbf{b}},\hat{\Gamma}} - f_{k_{n},\mathcal{B},\hat{\mathbf{b}},\hat{\Gamma}} \Vert_{\infty} + \Vert f_{k_{n},\mathcal{B},\hat{\mathbf{b}},\hat{\Gamma}} - f_{k_{n},\mathcal{B},\mathbf{b},\Gamma} \Vert_{\infty} ,
\end{align}
we have 
\begin{align}
&\pi\bigg(f \in \mathcal{F} : \Vert f - f_{0} \Vert_{\infty} \leq {\varepsilon \over 3}{\sqrt{2\over C_{A}}} \bigg) \nonumber\\
&\geq \pi(K=k_{n})\bigg(\prod_{S' \subseteq [p]}\pi(S=S')\bigg) \label{eq:two_probasbility_1}\\
&\quad \times \pi\bigg(\bigg\{ \Vert f_{k_{n},\hat{\mathcal{B}},\hat{\mathbf{b}},\hat{\Gamma}} - f_{k_{n},\mathcal{B},\hat{\mathbf{b}},\hat{\Gamma}} \Vert_{\infty} \leq {\varepsilon \over 3}{\sqrt{2 \over C_{A}}} \bigg\} \bigcap \bigg\{ \Vert f_{k_{n},\mathcal{B},\hat{\mathbf{b}},\hat{\Gamma}} - f_{k_{n},\mathcal{B},\mathbf{b},\Gamma}\Vert_{\infty} \leq {\varepsilon \over 3}{\sqrt{2 \over C_{A}}}\bigg\}\bigg) \label{eq:two_probasbility_2}.
\end{align}

Therefore, it remains to derive the lower bounds for (\ref{eq:two_probasbility_1}) and (\ref{eq:two_probasbility_2}).

\paragraph{Lower bound of (\ref{eq:two_probasbility_1}).} We have 
\begin{align*}
\pi(K=k_{n})\bigg(\prod_{S' \subseteq [p]}\pi(S=S')\bigg) &= \bigg(\prod_{S' \subseteq [p]}\pi(S=S')\bigg){\exp(-C_{0}k_{n}\log n) \over \sum_{k=0}^{K_{\max}}\exp(-C_{0}k\log n)} \\
&> \exp(-\mathfrak{d}_{1}n)
\end{align*}
for some positive constant $\mathfrak{d}_{1}$.

\paragraph{Lower bound of (\ref{eq:two_probasbility_2}).}
For any $\mathcal{B}=(\beta_{k}, k \in [k_{n}])\in \mathbb{R}^{k}$, we have
\begin{align}
\Vert f_{k_{n},\mathcal{B},\hat{\mathbf{b}},\hat{\Gamma}} - f_{k_{n},\hat{\mathcal{B}},\hat{\mathbf{b}},\hat{\Gamma}} \Vert_{\infty} &\leq \sup_{\mathbf{x}}\sum_{k=1}^{k_{n}}\bigg| \beta_{k}\prod_{j \in S_{k}}\phi(x_{j}|\{j\},\hat{b}_{j,k},\hat{\gamma}_{j,k}) - \hat{\beta}_{k}\prod_{j \in S_{k}}\phi(x_{j}|\{j\},\hat{b}_{j,k},\hat{\gamma}_{j,k}) \bigg | \nonumber\\
&\leq \sup_{\mathbf{x}} \sum_{k=1}^{k_{n}}\bigg|(\beta_{k} - \hat{\beta}_{k})\prod_{j \in S_{k}}\phi(x_{j}|\{j\},\hat{b}_{j,k},\hat{\gamma}_{j,k})\bigg| \nonumber\\
&\leq \sum_{k=1}^{k_{n}}\bigg|(\beta_{k} - \hat{\beta}_{k})(1+C_{\sigma})^{p} \bigg| \label{eq:basis_bdd}\\
&\leq (1+C_{\sigma})^{p}\Vert \mathcal{B} - \hat{\mathcal{B}} \Vert_{1} \nonumber\\
&\leq (1+C_{\sigma})^{p}\sqrt{k_{n}}\Vert \mathcal{B} - \hat{\mathcal{B}}\Vert_{2}. \nonumber
\end{align}

That is, we have
\begin{align*}
\bigg \{ \Vert f_{k_{n},\hat{\mathcal{B}},\hat{\mathbf{b}},\hat{\Gamma}} - f_{k_{n},\mathcal{B},\hat{\mathbf{b}},\hat{\Gamma}} \Vert_{\infty} \leq {\varepsilon \over 3} \sqrt{2  \over C_{A}} \bigg \} \supseteq \bigg\{ \Vert \mathcal{B}-\hat{\mathcal{B}} \Vert_{2} \leq ((1+C_{\sigma})^{p}\sqrt{k_{n}})^{-1}{\varepsilon \over 3} \sqrt{2  \over C_{A}} \bigg \}.
\end{align*}

Furthermore, direct calculation yields
\begin{align*}
\Vert f_{k_{n},\mathcal{B},\hat{\mathbf{b}},\hat{\Gamma}} - f_{k_{n},\mathcal{B},\mathbf{b},\Gamma} \Vert_{\infty} &= \sup_{\mathbf{x}}\sum_{k=1}^{k_{n}}|\beta_{k}|\bigg | \prod_{j \in S_{k}}\Big(\phi(x_{j}|\{j\},\hat{b}_{j,k},\hat{\gamma}_{j,k}) - \phi(x_{j}|\{j\},b_{j,k},\gamma_{j,k})  \Big) \bigg | \\
&\leq (1+C_{\sigma})\sup_{\mathbf{x}}\sum_{k=1}^{k_{n}}|\beta_{k}|\bigg| \prod_{j \in S_{k}}\bigg({x_{j}-\hat{b}_{j,k} \over \hat{\gamma}_{j,k}} - {x_{j}-b_{j,k} \over \gamma_{j,k}}\bigg) \bigg |\\
&= (1+C_{\sigma})\sup_{\mathbf{x}}\sum_{k=1}^{k_{n}}|\beta_{k}|\bigg| \prod_{j\in S_{k}}\bigg( {b_{j,k}-\hat{b}_{j,k}\over \hat{\gamma}_{j,k}} + (x_{j}-b_{j,k}){\gamma_{j,k}-\hat{\gamma}_{j,k}\over \gamma_{j,k}\hat{\gamma}_{j,k}}  \bigg) \bigg |\\
&\leq (1+C_{\sigma})\sup_{\mathbf{x}}\sum_{k=1}^{k_{n}}|\beta_{k}|\prod_{j\in S_{k}}\bigg(\bigg| {b_{j,k}-\hat{b}_{j,k} \over \hat{\gamma}_{j,k}}\bigg| + 2\bigg|{\gamma_{j,k}-\hat{\gamma}_{j,k}\over \gamma_{j,k}\hat{\gamma}_{j,k}} \bigg |\bigg).
\end{align*}
Let $C_{n,j,k} = {|\hat{\gamma}_{j,k}|\over 2}\bigg({\varepsilon \over 3\xi(1+C_{\sigma}) k_{n}}\sqrt{2 \over  C_{A}}\bigg)^{1/|S_{k}|}$.
If $|\gamma_{j,k} - \hat{\gamma}_{j,k}| \leq \epsilon_{1}$, we have
\begin{align*}
\bigg|{\gamma_{j,k} - \hat{\gamma}_{j,k} \over \hat{\gamma}_{j,k}\gamma_{j,k}}\bigg| \leq {\epsilon_{1} \over \hat{\gamma}_{j,k}(\hat{\gamma}_{j,k}-\epsilon_{1})} \leq {1\over 4}\bigg({\varepsilon \over 3\xi k_{n}}\sqrt{2 \over  C_{A}}\bigg)^{1/|S_{k}|},
\end{align*}
where $\epsilon_{1} = {C_{n,j,k}|\hat{\gamma}_{j,k}| \over 2 + C_{n,j,k}}$.
Therefore, if 
\begin{align*}
&|\beta_{k}| \leq \xi, \\
&|b_{j,k}-\hat{b}_{j,k}| \leq 2C_{n,j,k} , \\
&|\gamma_{j,k}-\hat{\gamma}_{j,k}|\leq {C_{n,j,k_{n}}|\hat{\gamma}_{j,k}| \over 2 + C_{n,j,k}}
\end{align*}
hold, we have
\begin{align}
\Vert f_{k_{n},\mathcal{B},\hat{\mathbf{b}},\hat{\Gamma}} - f_{k_{n},\mathcal{B},\mathbf{b},\Gamma} \Vert_{\infty} \leq {\varepsilon \over 3}\sqrt{2 \over C_{A}}.
\end{align}
That is, we have
\begin{align*}
&\bigg\{ \Vert f_{k_{n},\hat{\mathcal{B}},\hat{\mathbf{b}},\hat{\Gamma}} - f_{k_{n},\mathcal{B},\hat{\mathbf{b}},\hat{\Gamma}} \Vert_{\infty} \leq {\varepsilon \over 3}\sqrt{2 \over C_{A}}\bigg\} \bigcap \bigg\{ \Vert f_{k_{n},\mathcal{B},\hat{\mathbf{b}},\hat{\Gamma}} - f_{k_{n},\mathcal{B},\mathbf{b},\Gamma}\Vert_{\infty} \leq {\varepsilon \over 3}\sqrt{2 \over C_{A}}\bigg\} \\
&\supseteq \bigg \{ \Vert \mathcal{B}-\hat{\mathcal{B}} \Vert_{2} \leq ((1+C_{\sigma})^{p}\sqrt{k_{n}})^{-1} {\varepsilon \over 3}\sqrt{2 \over C_{A}},\\
&\quad \quad \:\: |\beta_{j}| \leq \xi, \\
&\quad \quad \:\: |b_{j,k}-\hat{b}_{j,k}| \leq 2C_{n,j,k} ,\\ &\quad\quad \:\: |\gamma_{j,k}-\hat{\gamma}_{j,k}| \leq {C_{n,j,k}|\hat{\gamma}_{j,k}|\over 2+ C_{n,j,k}}, \quad \forall j \in S_{k}, \forall k \in [k_{n}] \bigg \}.
\end{align*}

It implies that
\begin{align}
&\pi\bigg(\bigg\{ \Vert f_{k_{n},\hat{\mathcal{B}},\hat{\mathbf{b}},\hat{\Gamma}} - f_{k_{n},\mathcal{B},\hat{\mathbf{b}},\hat{\Gamma}} \Vert_{\infty} \leq {\varepsilon \over 3}\sqrt{2 \over C_{A}} \bigg\} \bigcap \bigg\{ \Vert f_{k_{n},\mathcal{B},\hat{\mathbf{b}},\hat{\Gamma}} - f_{k_{n},\mathcal{B},\mathbf{b},\Gamma}\Vert_{\infty} \leq {\varepsilon \over 3}\sqrt{2 \over C_{A}} \bigg\}\bigg) \nonumber\\
&\geq \pi(\Vert \mathcal{B}-\hat{\mathcal{B}} \Vert_{2} \leq ((1+C_{\sigma})^{p}\sqrt{k_{n}})^{-1}\varepsilon\sqrt{2 } / (3\sqrt{ C_{A}}), |\beta_{k}| \leq \xi , \:\: \forall k \in [k_{n}]) \label{eq:coeff_bound}\\
&\quad\times \pi(|b_{j,k}-\hat{b}_{j,k}|\leq 2C_{n,j,k},\:\forall j \in S_{k},\:\forall k \in [k_{n}]) \label{eq:location_bound}\\
&\quad \times \pi\bigg(|\gamma_{j,k}-\hat{\gamma}_{j,k}|\leq {C_{n,j,k} \over 1 + C_{n,j,k}}|\hat{\gamma}_{j,k}|,\:\forall j \in S_{k},\:\forall k \in [k_{n}] \bigg). \label{eq:LB_prior}
\end{align}

Now, we will show that these three probabilities sufficiently large.

\paragraph{Lower bound of (\ref{eq:coeff_bound}).}
Since
\begin{align}
&\bigg\{ \Vert \mathcal{B}-\hat{\mathcal{B}} \Vert_{2} \leq ((1+C_{\sigma})^{p}\sqrt{k_{n}})^{-1}\varepsilon \sqrt{2 } / (3\sqrt{ C_{A}}), |\beta_{k}| \leq \xi , \:\: \forall k \in [k_{n}] \bigg \} \nonumber\\
&\supseteq \bigg \{ |\beta_{k}-\hat{\beta}_{k} | \leq ((1+C_{\sigma})^{p}k_{n})^{-1}\varepsilon \sqrt{2 } / (3\sqrt{ C_{A}}) , |\beta_{k}| \leq \xi , \forall k \in [k_{n}] \bigg\} \nonumber\\
&\supseteq \bigg \{ |\beta_{k}-\hat{\beta}_{k} | \leq ((1+C_{\sigma})^{p}k_{n})^{-1}\varepsilon \sqrt{2 } / (3\sqrt{C_{A}}),  \forall k \in [k_{n}] \bigg\} \label{eq:LB_prior_each}
\end{align}
for sufficiently large $n$, it suffices to get the lower bound of $\pi (|\beta_{k}-\hat{\beta}_{k} | \leq ((1+C_{\sigma})^{p}k_{n})^{-1}\varepsilon\sqrt{2 }/(3\sqrt{ C_{A}}))$
for $k \in [k_{n}]$.

For $k \in [k_{n}]$, we let 
$$
I_{k}=\big[\hat{\beta}_{k} \pm ((1+ C_{\sigma})^{p}k_{n})^{-1}\varepsilon\sqrt{2}/(3\sqrt{ C_{A}})]$$
and we have
\begin{align}
&\pi (|\beta_{k}-\hat{\beta}_{k} | \leq ((1+C_{\sigma})^{p}k_{n})^{-1}\varepsilon\sqrt{2 }/(3\sqrt{ C_{A}})) \nonumber\\
&= \int_{I_{k}}{1\over \sqrt{2\pi}\sigma_{\beta}}\exp\bigg(-{\beta_{k}^{2}\over 2\sigma_{\beta}^{2}}\bigg)d\beta_{k} \nonumber\\
&\geq |I_{k}|{1\over \sqrt{2\pi}\sigma_{\beta}}\exp\bigg(-{(\max_{S}c_{S}+((1+C_{\sigma})^{p}k_{n})^{-1}\varepsilon\sqrt{2 }/(3\sqrt{ C_{A}}) )^{2}\over 2\sigma_{\beta}^{2}}\bigg) \label{eq:normal_bound}\\
&> \exp(-\mathfrak{d}_{1}n) \nonumber
\end{align}
for some positive constant $\mathfrak{d}_{1}$,
where (\ref{eq:normal_bound}) is derived from $|\hat{\beta}_{k}| \leq \max_{S}c_{S}$.

\paragraph{Lower bound of (\ref{eq:location_bound}).}
Since 
\begin{align*}
\pi\big(|b_{j,k}-\hat{b}_{j,k}|\leq 2C_{n,j,k} \big) = 4C_{n,j,k}
\end{align*}
for all $j \in S_{k},\:k \in [k_{n}]$, we have
\begin{align*}
\pi\big(|b_{j,k}-\hat{b}_{j,k}|\leq 2C_{n,j,k},\:\forall j \in S_{k},\:\forall k \in [k_{n}]\big) &= \prod_{k \in [k_{n}], j \in S_{k}}4C_{n,j,k}\\
& > \exp(-\mathfrak{d}_{2}n)
\end{align*}
for some positive constant $\mathfrak{d}_{2}$.

\paragraph{Lower bound of (\ref{eq:LB_prior}).}
Using direct calculation, we have
\begin{align*}
\pi\bigg(|\gamma_{j,k}-\hat{\gamma}_{j,k}|\leq {C_{n,j,k} \over 2+ C_{n,j,k}}\hat{\gamma}_{j,k} \bigg ) 
&\geq \bigg( {2C_{n,j,k}\hat{\gamma}_{j,k}\over 2+C_{n,j,k}} \bigg )\min_{x \in [L_{n},U_{n}]}pdf_{\gamma}(x) \\
&= \bigg( {2C_{n,j,k}\hat{\gamma}_{j,k}\over 2+C_{n,j,k}} \bigg ) {b_{\gamma}^{a_{\gamma}}\over \Gamma(a_{\gamma})}\min_{x \in [L_{n},U_{n}]}x^{a_{\gamma}-1}\exp(-b_{\gamma}x),
\end{align*}
where $L_{n} = \hat{\gamma}_{j,k} - {C_{n,j,k}\hat{\gamma}_{j,k}\over 2 + C_{n,j,k}}$ and $U_{n} = \hat{\gamma}_{j,k} + {C_{n,j,k}\hat{\gamma}_{j,k}\over 2 + C_{n,j,k}}$.

Note that $1/ k_{n}^{3} \leq \hat{\gamma}_{i,j} \leq 1$.
For $a_{\gamma} > 1$, we have
\begin{align*}
\min_{x\in[L_{n},U_{n}]}x^{a_{\gamma}-1} &\geq L_{n}^{a_{\gamma}-1}\\
&= \bigg( {2\hat{\gamma}_{j,k} \over 2 + C_{n,j,k} } \bigg)^{a_{r}-1} \\
&> \exp(-\mathfrak{d}_{3}n)
\end{align*}
for some positive constant $\mathfrak{d}_{3}$ and for $a_{\gamma} < 1$, we have
\begin{align*}
\min_{x \in [L_{n},U_{n}]}x^{a_{\gamma}-1} &\geq U_{n}^{a_{\gamma}-1} \\
&= \bigg( \hat{\gamma}_{j,k} \bigg)^{1-a_{\gamma}} \\
&> \exp(-\mathfrak{d}_{4}n)
\end{align*}
for some positive constant $\mathfrak{d}_{4}$.
Furthermore, we have
\begin{align*}
\min_{x \in [L_{n},U_{n}]}\exp(-b_{\gamma}x) &\geq \exp(-b_{\gamma}U_{n}) \\
&\geq \exp(-2b_{\gamma}\hat{\gamma}_{i,j}) \\
&> \exp(-2\mathfrak{d}_{5}n)
\end{align*}
and
\begin{align*}
{2C_{n,j,k}\hat{\gamma}_{j,k}\over 2+C_{n,j,k}} 
&>  \exp(-2\mathfrak{d}_{7}n)
\end{align*}
for some positive constants $\mathfrak{d}_{6}$ and $\mathfrak{d}_{7}$.
Finally, the proof is completed by letting $C_{2} = \sum_{i=1}^{7}\mathfrak{d}_{i}$.

\qed

\subsection{Verifying Condition (\ref{eq:ghosal_condition3})}
\label{sec:prof_ghosal_cond3}

We will verify Condition (\ref{eq:ghosal_condition3}) with the constant $C_{3}$.

We let
\begin{align*}
Z_{1} &= \big\{K > M_{n} \big\}, \\
Z_{2} &= \big\{ \{ K \leq M_{n} \} \cap \{ \exists k \in [K] \: \text{such that} \: |\beta_{k}| > n \} \big\}, \\
Z_{3} &= \big\{ \{ K \leq M_{n} \} \cap \{\exists k \in [K] \: \text{such that} \: \Gamma_{S_{k},k} \in (n,\infty)^{|S_{k}|} \} \big\}.
\end{align*}
Since
\begin{align*}
\pi(\mathcal{F}\backslash \mathcal{F}^{n}) = \pi(Z_{1} \cup Z_{2} \cup Z_{3}),
\end{align*}
the upper bound of $\pi(\mathcal{F}\backslash \mathcal{F}^{n})$ is
\begin{align}
&\pi(\mathcal{F}\backslash \mathcal{F}^{n})\nonumber \\
&\leq \pi(K > M_{n}) \label{eq:condition3_upper1}\\
&\quad+ \pi(K \leq M_{n})\pi(\exists k \in [K] \: \text{such that} \: |\beta_{k}| > n  | K \leq M_{n}) \label{eq:condition3_upper2}\\
&\quad+ \pi(K \leq M_{n})\pi(\exists k \in [K] \: \text{such that} \: \Gamma_{S_{k},k} \in (n,\infty)^{|S_{k}|}  | K \leq M_{n}). \label{eq:condition3_upper3}
\end{align}

\paragraph{Upper bound of (\ref{eq:condition3_upper1}).}
For $M_{n} = \lfloor {C_{3}n\varepsilon^{2} \over \log n } \rfloor$, we have
\begin{align*}
\pi(K > M_{n}) &= {\sum_{k=M_{n}+1}^{K_{\max}} \exp(-kC_{0}\log n ) \over \sum_{k=0}^{K_{\max}}\exp(-kC_{0}\log n ) } \\
&\leq \exp(-M_{n}C_{0}\log n).
\end{align*}

Since $C_{3} > {C_{2} +2 \over C_{0}\log n}$ for sufficiently large $n$, we have
\begin{align*}
\pi(K > M_{n})\exp((C_{2}+2)n) \xrightarrow{} 0 \:\: \text{as} \:\: n \xrightarrow{} \infty.
\end{align*}

\paragraph{Upper bound of (\ref{eq:condition3_upper2}).}
We have
\begin{align*}
\pi(\exists k \in [K] \: \text{such that} \: |\beta_{k}| > n  | K \leq M_{n}) &\leq M_{n}\pi(|\beta_{1}|>n) \\
& \leq 2M_{n}\exp\bigg(-{n^{2}\over 2\sigma_{\beta}^{2}}\bigg),
\end{align*}
where $\sigma_{\beta}^{2}$ is a constant.
That is, we conclude that
\begin{align*}
\pi(\exists k \in [K] \: \text{such that} \: |\beta_{k}| > n  | K \leq M_{n})\exp((C_{2}+2)n) \xrightarrow{} 0 \:\: \text{as} \:\: n\xrightarrow{} \infty.
\end{align*}

\paragraph{Upper bound of (\ref{eq:condition3_upper1}).}
For any $j,k$, using Markov inequality, we have
\begin{align*}
\pi(\gamma_{j,k} > n) &\leq \mathbb{E}\bigg[\exp\bigg({b_{\gamma}\gamma_{j,k}\over 2}\bigg)\bigg]\exp\bigg(-{b_{\gamma}n\over 2}\bigg) \\
&= \bigg( {1\over 2}\bigg )^{-a_{\gamma}}\exp\bigg(-{b_{\gamma}n\over 2}\bigg).
\end{align*}
Since
\begin{align*}
\pi(\exists k \in [K] \: \text{such that} \: \Gamma_{S_{k},k} \in (n,\infty)^{|S_{k}|}  | K \leq M_{n}) &\leq M_{n}\pi(\gamma_{1,1} > n),
\end{align*}
we have
\begin{align*}
\pi(\exists k \in [K] \: \text{such that} \: \Gamma_{S_{k},k} \in (n,\infty)^{|S_{k}|}  | K \leq M_{n})\exp((C_{2}+2)n) \xrightarrow{} 0 \quad \text{as} \quad n \xrightarrow{} \infty,
\end{align*}
where $a_{\gamma}$ and $b_{\gamma}$ are positive constants.

\qed

\subsection{Verification of the Conditions in \cite{ghosal1999posterior}}
\label{sec:condition_3_verify}
For given data $\mathbf{x}^{(n)}$, let $q_{f,i}$ be the probability density of $\mathbb{Q}_{f(\mathbf{x}_{i})}$ for $i=1,...,n$.
From Theorem 2 of \cite{ghosal1999posterior}, it suffices to verify that for every $f_0\in \text{Lip}_{0,L,F}$, there exists a sieve $\mathcal{F}^{n}_{\xi}$, constants $\delta < \varepsilon /4,C_{5},C_{6} > 0$ and $C_{1} < \varepsilon^{2} /8$ such that the following three conditions hold with respect to the $\Vert \cdot \Vert_{2,n}.$
\begin{align}
& \log \mathcal{N}\left(\delta,\mathcal{F}^{n}_\xi ,\Vert \cdot \Vert_{2,n} \right) < nC_{1}, \label{eq:general_cond1}\\
&\pi_{\xi} \bigg( f \in \mathcal{F}_{\xi} : {1\over n }\sum_{i=1}^{n}K(q_{f_{0},i},q_{f,i}) \leq \varepsilon^{2} \bigg) > \exp(-nC_{5}), \label{eq:general_cond2}
\\
&\pi_{\xi}\big(\mathcal{F}_\xi\backslash \mathcal{F}^{n}_\xi\big) < \exp(-nC_{6}) \label{eq:general_cond3}.
\end{align}

To complete the proof of Theorem \ref{thm:posterior_rate_exponetial}, we will verify that the three conditions (\ref{eq:general_cond1}), (\ref{eq:general_cond2}), and (\ref{eq:general_cond3}) for given data $\mathbf{x}^{(n)}$.

\paragraph{Verifying Condition (\ref{eq:general_cond1}).}
$\newline$

Condition (\ref{eq:general_cond1}) holds under Condition (\ref{eq:ghosal_condition1}).

\paragraph{Verifying Condition (\ref{eq:general_cond2}).}
$\newline$

By using a direct calculation, for $i=1,...,n$, we have
\begin{align}
K(q_{f_{0},i},q_{f,i}) &= \int \Big( (f_{0}(\mathbf{x}_{i}) - f(\mathbf{x}_{i}))y - A(f_{0}(\mathbf{x}_{i})) + A(f(\mathbf{x}_{i})) \Big)q_{f_{0},i}(y)dy \\
&= \bigg( (f_{0}(\mathbf{x}_{i}) - f(\mathbf{x}_{i}))\mathbb{E}[Y_{i}] - A(f_{0}(\mathbf{x}_{i})) + A(f(\mathbf{x}_{i})) \bigg) \\
&= \bigg((f_{0}(\mathbf{x}_{i}) - f(\mathbf{x}_{i}))\dot{A}(f_{0}(\mathbf{x}_{i})) - A(f_{0}(\mathbf{x}_{i})) + A(f(\mathbf{x}_{i})) \bigg).
\end{align}

Using Talyor expansion, we have
\begin{align*}
K(q_{f_{0},i},q_{f,i}) = {1\over 2}\ddot{A}(\Tilde{x})(f_{0}(\mathbf{x}_{i}) - f(\mathbf{x}_{i}))^{2},
\end{align*}
where $\Tilde{x} \in [-F,F]$.
That is, we have
\begin{align*}
{1\over n}\sum_{i=1}^{n}K(q_{f_{0},i},q_{f,i}) \leq {C_{A}\over 2}\Vert f_{0}-f \Vert_{2,n}^{2}.
\end{align*}

When $\xi \geq 2^{P}F+\varepsilon \sqrt{2 \over C_{A}}$, we have
\begin{align*}
&\pi_\xi\bigg(f \in \mathcal{F}_{\xi} : \Vert f - f_{0} \Vert_{2,n} \leq \varepsilon \sqrt{2 \over C_{A}} \bigg)\geq \pi\bigg(f \in \mathcal{F}_{\xi} : \Vert f - f_{0} \Vert_{2,n} \leq  \varepsilon \sqrt{2 \over C_{A}} \bigg).
\end{align*}

Therefore, the proof is done by Condition (\ref{eq:ghosal_condition2}).

\qed

\paragraph{Verifying Condition (\ref{eq:general_cond3}).}
$\newline$

Since
\begin{align*}
\pi_\xi(\mathcal{F}\backslash \mathcal{F}^n) &\le \frac{\pi(\mathcal{F}\backslash \mathcal{F}^n)}{\pi( \Vert f\ \Vert_{\infty} \le \xi)} \\
&\le  \frac{\pi(\mathcal{F}\backslash \mathcal{F}^n)}{\pi\bigg(\|f-f_0\|_\infty \le \varepsilon \sqrt{2  \over C_{A}} \bigg)} \\
&\leq \exp(-(C_{5}+2)n)
\end{align*}
for $2^{p}F + \varepsilon \sqrt{2  \over C_{A}} \leq \xi$, the condition (\ref{eq:general_cond3}) holds for $C_{6}=C_{5}+2$ by condition (\ref{eq:ghosal_condition2}) and (\ref{eq:ghosal_condition3}).

\qed

\subsection{STEP \ref{eq:Step_N3}}

Since \ref{eq:Step_N2} holds for arbitary $\mathbf{x}^{(n)}$, we conclude that 
\begin{align*}
\lim_{n \to \infty}\mathbb{E}_{0}^{n}[\pi_{\xi}(\Vert f - f_{0} \Vert_{2,n} > \varepsilon | \mathbf{X}^{(n)},Y^{(n)})] = 0
\end{align*}
for any $\varepsilon > 0$.  

\qed

\newpage

\section{Proof of Theorem \ref{thm:posterior_rate_component-wise}}

The proof consists of the following 4 steps.

\paragraph{(STEP E.1)}
$\newline$
We first establish the rate at which the posterior concentrates under the population $\ell_{2}$ norm; specifically, we demonstrate that
\begin{align}
\label{eq:step1}
\mathbb{E}_{0}^{n}\Big[\pi_{\xi}\Big ( f \in \mathcal{F}_{\xi}^{n}: \Vert f- f_{0} \Vert_{2,\mathbb{P}_{\mathbf{X}}}  > \varepsilon | \mathbf{X}^{(n)},Y^{(n)} \Big )\Big] \xrightarrow{} 0,
\end{align}
for any $\varepsilon >0$.

\paragraph{(STEP E.2)}
$\newline$
Based on (\ref{eq:step1}), we establish that the following holds for any subset $S \subseteq [p]$.
\begin{align}
\label{eq:step2}
\mathbb{E}_{0}^{n}\Big[\pi_{\xi}\Big ( f \in \mathcal{F}_{\xi}^{n}: \Vert f_{S}- f_{0,S} \Vert_{2,\mathbb{P}_{\mathbf{X}}}  > \varepsilon | \mathbf{X}^{(n)},Y^{(n)} \Big ) \Big] \xrightarrow{} 0,
\end{align}
for any $\varepsilon > 0$.

\paragraph{(STEP E.3)}
$\newline$
We reformulate (\ref{eq:step2}) in terms of the empirical $\ell_{2}$ norm.
Specifically, we demonstrate that
\begin{align}
\label{eq:step3}
\mathbb{E}_{0}^{n}\Big[\pi_{\xi}\Big ( f \in \mathcal{F}_{\xi}^{n}: \Vert f_{S}- f_{0,S} \Vert_{2,n}  > \varepsilon | \mathbf{X}^{(n)},Y^{(n)} \Big ) \Big] \xrightarrow{} 0,
\end{align}
for any $\varepsilon > 0$.

\paragraph{(STEP E.4)}
$\newline$
The last step is to verify
\begin{align}
\mathbb{E}_{0}^{n}\Big[\pi_{\xi}( \mathcal{F}_{\xi} \backslash \mathcal{F}_{\xi}^{n} | \mathbf{X}^{(n)},Y^{(n)})\Big] \xrightarrow{} 0
\label{eq:step4}
\end{align}
as $n\rightarrow \infty.$

\subsection{Verifying (\textbf{STEP D.1})}

To verify (\textbf{STEP D.1}), we rely on the following lemma, whose proof is provided in Theorem 19.3 of \cite{gyorfi2006distribution}.

\begin{lemma}[Theorem 19.3 of \cite{gyorfi2006distribution}]\label{gyorfi2002distribution}
	Let $\mathbf{X}, \mathbf{X}_1, \dots, \mathbf{X}_n$ be independent and identically distributed random vectors with values in $\mathbb{R}^d$. Let $K_1, K_2 \geq 1$ be constants and let $\mathcal{G}$ be a class of functions $g : \mathbb{R}^d \to \mathbb{R}$ with the properties
	\begin{align}
	|g(\mathbf{x})| \leq K_1, \quad \mathbb{E}[g(\mathbf{X})^2] \leq K_2 \mathbb{E}[g(\mathbf{X})].
    \label{eq:gyo_1}
	\end{align}
	Let $0<\kappa<1$ and $\zeta>0$. Assume that
	\begin{align}
	\sqrt{n} \kappa \sqrt{1-\kappa} \sqrt{\zeta} \geq 288 \max \left\{2 K_{1}, \sqrt{2 K_{2}}\right\}
    \label{eq:gyo_2}
	\end{align}
	and that, for all $\mathbf{x}_1 , \dots , \mathbf{x}_n \in \mathbb{R}^d$ and for all $t \geq \frac{\zeta}{8}$,
    \begin{small}
	\begin{align}
	\frac{\sqrt{n} \kappa(1-\kappa) t}{96 \sqrt{2} \max \left\{K_{1}, 2 K_{2}\right\}} 
	\geq \int_{\frac{\kappa(1-\kappa)t}{16 \max \left\{K_{1}, 2 K_{2}\right\}}}^{\sqrt{t}}  
	\sqrt{\log \mathcal{N}\left(u,\left\{g \in \mathcal{G}: \frac{1}{n} \sum_{i=1}^{n} g\left(\mathbf{x}_{i}\right)^{2} \leq 16 t\right\}, ||\cdot||_{1,n}\right)} d u.
	\label{eq:gyo_3}
    \end{align}
    \end{small}
	Then,
	\begin{align*}
	\mathbb{P}^{n}_{\mathbf{X}}\Bigg(\sup _{g \in \mathcal{G}} \frac{\left|\mathbb{E}[g(\mathbf{X})]-\frac{1}{n} \sum_{i=1}^{n} g\left(\mathbf{X}_{i}\right)\right|}{\zeta +\mathbb{E}[g(\mathbf{X})]}>\kappa\Bigg) \nonumber 
	\leq 60 \exp \left(-\frac{n \zeta \kappa^{2}(1-\kappa)}{C_{g} \max \left\{K_{1}^{2}, K_{2}\right\}}\right)
	\end{align*}
for some positive constant $C_{g}$.
\end{lemma}

Since $\mathcal{F}^{n}_{\xi}$ depends on the dataset $\mathbf{X}^{(n)}$, we will apply Lemma \ref{gyorfi2002distribution} to the function class $\mathcal{G}^{n}_{\xi}$ defined as
$$
\mathcal{G}^{n}_{\xi}=\bigcup_{K=1}^{M_{n}}\mathcal{G}^{n}_{\xi}(K),
$$
where $\mathcal{G}^{n}_{\xi}(K) = \{f \in \mathcal{G}^{n}(K) : \Vert f \Vert_{\infty} \leq \xi \}$.
Here, $\mathcal{G}^{n}(K)$ is defined in (\ref{eq:g_n}).

Since 
\begin{align*}
\mathcal{N}(\epsilon, \mathcal{G}_{\xi}^{n}, \Vert \cdot \Vert_{\infty} ) &\leq \mathcal{N}(\epsilon, \mathcal{G}^{n}, \Vert \cdot \Vert_{\infty} )\\
&\leq M_{n}\bigg ( 1+ {M_{n}2^{p+3}n^{3p+1}\over \epsilon}\bigg)^{M_{n}(1+3p)},
\end{align*}
we can easily verify that conditions (\ref{eq:gyo_1}), (\ref{eq:gyo_2}), and (\ref{eq:gyo_3}) hold for $K_{1}=K_{2}=4\xi^{2}$, $\kappa={1\over 4}$, $\zeta = \varepsilon^{2}$, and $\mathcal{G} = \{ g: g=(f_{0}-f)^{2}, f \in \mathcal{G}_{\xi}^{n}\}$.
That is, we have
\begin{align*} 
&\mathbb{P}^{n}_{\mathbf{X}}\Bigg (\sup _{f \in \mathcal{F}_{\xi}^{n}} \frac{\left| ||f-f_{0}||_{2, \mathbb{P}_{\mathbf{X}}}^2  - ||f-f_{0}||_{2, n}^2 \right|}{\varepsilon^2+||f-f_{0}||_{2, \mathbb{P}_{\mathbf{X}}}^2} > \frac{1}{4}\Bigg ) \leq 60 \exp \left(-\frac{n \varepsilon^2 / 8}{C_{g}\cdot 16\xi^4}\right).
\end{align*}
We define $\mathcal{A}_{n} := \bigg\{\mathbf{X}^{(n)} : \sup _{f \in \mathcal{F}_{\xi}^{n}} \frac{\left| ||f-f_{0}||_{2, \mathbb{P}_{\mathbf{X}}}^2  - ||f-f_{0}||_{2, n}^2 \right|}{\varepsilon^2+||f-f_{0}||_{2, \mathbb{P}_{\mathbf{X}}}^2} \leq {1\over 4} \bigg\}$.
Then, we have
\begin{align*}
&\mathbb{E}_{0}^{n}\Big[ \pi_{\xi}\Big( f \in \mathcal{F}_{\xi}^{n} : \Vert f - f_{0} \Vert_{2,\mathbb{P}_{\mathbf{X}}} > \varepsilon | \mathbf{X}^{(n)},Y^{(n)}\Big) \Big] \\
&\leq \mathbb{E}_{0}^{n}\Big[ \pi_{\xi}\Big( f \in \mathcal{F}_{\xi}^{n} : \Vert f - f_{0} \Vert_{2,\mathbb{P}_{\mathbf{X}}} > \varepsilon | \mathbf{X}^{(n)},Y^{(n)}\Big)\mathbb{I}(\mathbf{X}^{(n)}\in \mathcal{A}_{n}) \Big] + \mathbb{P}_{\mathbf{X}}^{n}(\mathcal{A}_{n}^{c}) \\
&\leq \mathbb{E}_{0}^{n}\Big[ \pi_{\xi}\Big( f \in \mathcal{F}_{\xi}^{n} : \Vert f - f_{0} \Vert_{2,n} > \varepsilon /\sqrt{2} | \mathbf{X}^{(n)},Y^{(n)}\Big) \Big] +  \mathbb{P}_{\mathbf{X}}^{n}(\mathcal{A}_{n}^{c}) \\
& \xrightarrow{} 0 
\end{align*}
as $n \xrightarrow{} \infty$.

\qed

\subsection{Verifying (\textbf{STEP D.2})}

For $f \in \mathcal{F}_{\xi}^{n}$, we have
\begin{align*}
f(\mathbf{x}) = \sum_{S \subseteq [p]}f_{S}(\mathbf{x}_{S}),
\end{align*}
where $f_{S}$ satisfies the sum-to-zero condition with respect to the uniform distribution on $(0,1)$.

Consider positive constants $C_{7}$ and $C_{8}$ such that
\begin{align}
C_{7} \leq \inf_{\mathbf{x} \in \mathcal{X}}p_{\mathbf{X}}(\mathbf{x})  \leq \sup_{\mathbf{x} \in \mathcal{X}}p_{\mathbf{X}}(\mathbf{x}) \leq C_{8}. \label{eq:ratio_density}
\end{align}

Therefore, using the inequality (\ref{eq:ratio_density}), for all $S \subseteq [p]$, we have
\begin{align}
\Vert f_{0} - f \Vert_{2,\mathbb{P}_{\mathbf{X}}} &\geq \sqrt{ C_{7}\int_{\mathcal{X}} (f_{0}(\mathbf{x}) - f(\mathbf{x}))^{2}d\mathbf{x}} \nonumber\\
&= \sqrt{C_{7}\sum_{S \subseteq [p]}\int_{\mathcal{X}_{S}}(f_{0,S}(\mathbf{x}_{S})-f_{S}(\mathbf{x}_{S}))^{2} d\mathbf{x}_{S}} \label{eq:sum-to-zero_condition}\\
&\geq C_{9}\Vert f_{P,S} - f_{0,S} \Vert_{2,\mathbb{P}_{\mathbf{X}}}\nonumber,
\end{align}
where (\ref{eq:sum-to-zero_condition}) is derived from the sum-to-zero condition with respect to the uniform distribution on $(0,1)$ and $C_{9} = \sqrt{C_{7}/ C_{8}}$.

Hence, we conclude that
\begin{align*}
&\mathbb{E}_{0}^{n}\Big[ \pi_{\xi}\Big( f \in \mathcal{F}_{\xi}^{n} : \Vert f_{S} - f_{0,S} \Vert_{2,\mathbb{P}_{\mathbf{X}}} > \varepsilon | \mathbf{X}^{(n)},Y^{(n)}\Big) \Big] \\
&\leq \mathbb{E}_{0}^{n}\Big[ \pi_{\xi}\Big( f \in \mathcal{F}_{\xi}^{n} : \Vert f - f_{0} \Vert_{2,\mathbb{P}_{\mathbf{X}}} > \varepsilon C_{9}  | \mathbf{X}^{(n)},Y^{(n)}\Big) \Big] \\
& \xrightarrow{} 0,
\end{align*}
as $n \xrightarrow{} \infty$.

\qed

\subsection{Verifying (\textbf{STEP D.3})}

Following the same approach as in the proof of (\textbf{STEP D.1}), and applying Lemma \ref{gyorfi2002distribution} to the function class $\mathcal{G} = \{g : g= (f_{0,S}-f_{S})^{2} , f \in \mathcal{G}_{\xi}^{n} \}$, we have
\begin{align*}
\lim_{n\to \infty}\mathbb{E}_{0}^{n}\Big[ \pi_{\xi}\Big( f \in \mathcal{F}_{\xi}^{n} : \Vert f_{S} - f_{0,S}  \Vert_{2,n} > \varepsilon \Big| \mathbf{X}^{(n)},Y^{(n)}\Big) \Big]=0.
\end{align*}

\qed

\subsection{Verifying the (\textbf{STEP D.4})}

Since 
\begin{align*}
{\pi_{\xi}(\mathcal{F}_{\xi}\backslash \mathcal{F}_{\xi}^{n}) \over \pi_{\xi}(\mathbb{B}_{n})} \leq \exp(-2n)
\end{align*}
for given data $\mathbf{x}^{(n)}$, using Lemma 1 in \cite{ghosal2007convergence}, we conclude that
\begin{align*}
\lim_{n \to \infty}\mathbb{E}_{0}^{n}\Big[ \pi_{\xi}(\mathcal{F}_{\xi}\backslash \mathcal{F}_{\xi}^{n} | \mathbf{X}^{(n)},Y^{(n)}) \Big | \mathbf{X}^{(n)} = \mathbf{x}^{(n)} \Big] = 0.
\end{align*}
Since it holds for arbitrary $\mathbf{x}^{(n)}$, the proof is completed.

\qed

\end{document}